\documentclass[jair,twoside,11pt,theapa]{article}

\usepackage[utf8]{inputenc}
\usepackage[english]{babel}

\usepackage{tabularx,lipsum,environ,amsmath,amsthm,amssymb,float,forest,nicefrac, multirow, physics, color, courier, hyperref,comment,soul, xcolor,comment, xspace, colortbl, enumitem, arydshln}
\usepackage{soul}
\usepackage{listings}
\usepackage{fdsymbol}
\usepackage{mathtools}
\usepackage{dsfont}
\usepackage{graphicx}
\usepackage{booktabs}   
\usepackage{subcaption}
\usepackage{hhline}
\usepackage{letltxmacro, changepage}
\usepackage{array}
\usepackage{algorithm}
\usepackage[export]{adjustbox}
\usepackage[noend]{algpseudocode}
\usepackage[mathlines]{lineno}
\newcolumntype{L}{>{\centering\arraybackslash}m{1.5cm}}
\newcolumntype{M}{>{\centering\arraybackslash}m{2cm}}
\newcolumntype{N}{>{\centering\arraybackslash}m{3cm}}
\newcolumntype{O}{>{\centering\arraybackslash}m{3.05cm}}
\newcolumntype{P}{>{\centering\arraybackslash}m{3.6cm}}

\LetLtxMacro\oldttfamily\ttfamily
\DeclareRobustCommand{\ttfamily}{\oldttfamily\csname ttsize\endcsname}
\newcommand{\setttsize}[1]{\def\ttsize{#1}}%

\newcommand{\dcplus}{DC\#\xspace}

\newcommand{\dc}{DC($\mathcal{B}$)\xspace}
\setttsize{\small}
\newcounter{question}[section]

\newtheorem{theorem}{Theorem}
\newtheorem{lemma}[theorem]{Lemma}
\newtheorem{proposition}[theorem]{Proposition}
\newtheorem{definition}{Definition}
\newtheorem{example}{Example}

\usepackage{jair, theapa, rawfonts}

\jairheading{1}{1993}{1-15}{6/91}{9/91}
\ShortHeadings{First-Order Context-Specific Likelihood Weighting}
{Kumar, Ku\v zelka, \& De Raedt}
\firstpageno{1}

\begin{document}

\title{First-Order Context-Specific Likelihood Weighting in Hybrid Probabilistic Logic Programs}

\author{\name Nitesh Kumar
        \email nitesh.kr369@gmail.com \\
        \addr Department of Computer Science,\\
        KU Leuven, Belgium
        \AND
        \name Ond\v rej Ku\v zelka 
        \email ondrej.kuzelka@fel.cvut.cz \\
        \addr Faculty of Electrical Engineering,\\
        Czech Technical University in Prague, Czech Republic
        \AND
        \name Luc De Raedt 
        \email luc.deraedt@kuleuven.be \\
        \addr Department of Computer Science,\\
        KU Leuven, Belgium
        }


\maketitle
\begin{abstract}
Statistical relational AI and probabilistic logic programming have so far mostly focused on discrete probabilistic models. The reasons for this is that one needs to provide constructs to succinctly model the independencies in such models, and also provide efficient inference. 

Three types of independencies are important to represent and exploit for scalable inference in hybrid models: conditional independencies elegantly modeled in Bayesian networks, context-specific independencies naturally represented by logical rules, and independencies amongst attributes of related objects in relational models succinctly expressed by combining rules.

This paper introduces a hybrid probabilistic logic programming language, \dcplus, which integrates distributional clauses' syntax and semantics principles of Bayesian logic programs. It represents the three types of independencies qualitatively. More importantly, we also introduce the scalable inference algorithm FO-CS-LW for \dcplus. FO-CS-LW is a first-order extension of the context-specific likelihood weighting algorithm (CS-LW), a novel sampling method that exploits conditional independencies and context-specific independencies in ground models. The FO-CS-LW algorithm upgrades CS-LW with unification and combining rules to the first-order case. 

\textbf{Under consideration for publication in the journal of artificial intelligence research.} 

\end{abstract}


\section{Introduction}
Statistical relational AI (StarAI) and probabilistic logic programming (PLP) \cite{deraedt15,raedt2016statistical} have contributed many languages for declaratively modeling expressive probabilistic models and have devised numerous inference techniques.  They have been applied to
many applications in databases, knowledge graphs, social networks, robotics, chemical compounds, genomics, etc.

To enable scalable probabilistic inference, it is essential to represent three different types of independencies in the modeling language. Firstly, the traditional classical conditional independencies (CIs) represented by Bayesian networks (BNs). Secondly, the context-specific independencies (CSIs): independencies that hold only in certain contexts \cite{boutilier1996context}. These independencies arise due to structures present in conditional probability distributions (CPDs) of BNs, which BNs do not qualitatively represent, but rule-based representations do by making structures explicit in the clauses \cite{poole1997probabilistic}.
Thirdly, the combining rules such as NoisyOR to express probabilistic influences among attributes of related objects \cite{koller2007introduction,jaeger2007parameter}. Combining rules are particularly interesting since they allow one to qualitatively represent independence of causal influences \cite[ICIs]{Zhang1996ExploitingCI}, where each influence is considered independent of others. This independence is natural and commonly assumed to keep relational models succinct \cite{natarajan2008learning}. In probabilistic logic programs that deal with only discrete random variables, combining rules are the core component \cite{kersting20071,Raedt2007ProbLogAP}.
Over the past few decades, many PLP languages have been proposed; however, only a few of them are hybrid, i.e., support both discrete and continuous random variables. Nevertheless, such hybrid PLP are needed to cope with applications in areas such as activity recognition, robotics, sensing, perception, etc. Current hybrid PLP languages suffer from two problems. Firstly, they generally do not support combining rules \cite{gutmann2010extending,gutmann2011magic,islam2012inference,Michels2016ApproximatePI,alberti2017cplint}. Secondly, and more importantly, there exist, to the best of our knowledge, no inference techniques that exploit all three types of independencies for hybrid PLPs. 

To remedy this, we first introduce \dcplus (pronounced ``DC sharp''), a hybrid PLP language that supports combining rules.
\dcplus uses a special form of clauses called distributional clauses (DCs) to express probabilistic knowledge. We borrow the syntax of DCs from \cite{nitti2016probabilistic} but introduce an extended new semantics based on Bayesian Logic Programs \cite[BLPs]{kersting20071}. Thus, in terms of representation, \dcplus, a rule-based representation, differs from graphical model-based relational representations, such as BLPs, which are associated with CPDs. However, the semantics of \dcplus are based on BLPs, so \dcplus programs can be seen as BLPs qualitatively representing CSIs. 

Our second contribution is the first-order context-specific likelihood weighting (FO-CS-LW) algorithm that exploits these three types of independencies for scalable inference in \dcplus programs. Before going to the first-order case, we introduce the CS-LW algorithm for ground programs. CS-LW exploits both CIs and CSIs, which is an approximate inference algorithm that, until our earlier work, has not been well-studied yet \cite{kumar2021context}.

There exist state-of-the-art inference algorithms for exact inference \cite{friedman2018approximate} in discrete models. These algorithms are based on the knowledge compilation technique \cite{darwiche2002knowledge} that uses logical reasoning to exploit CSIs. However, for many models, exact inference quickly becomes infeasible. Stochastic sampling for approximate inference is a standard solution. Sampling algorithms are simple yet powerful tools, and they can be applied to arbitrary complex hybrid models, unlike exact inference. It is widely believed that CSI properties in distributions are difficult to exploit for approximate inference \cite{friedman2018approximate}. To solve this difficult problem, we introduce the context-specific likelihood weighting (CS-LW), a sampling algorithm that exploits both CI and CSI properties, leading to faster convergence and faster sampling speed than standard likelihood weighting (LW). 

Next, we extend CS-LW to first-order \dcplus programs specifying relational probabilistic models. Due to the use of combining rules, such models, when grounded, have many symmetrically repeated parameters. Inference algorithms designed for grounded models can not exploit these symmetries, rendering inference infeasible even for very simple relational models. It is widely believed that the inference can be feasible if one does not ground out these models, but reason at the first-order level with unification \cite{poole2003first}. There have been several attempts at doing this for various simple languages \cite{kisynski2009lifted,choi2011efficient,VandenBroeck2011LiftedPI,beame2015symmetric}, but not for hybrid and expressive languages like \dcplus. For example, well-known graphical model-based relational representation languages that do not qualitatively represent CSIs, construct ground BNs for inference \cite{getoor2007probabilistic,kersting20071}. Similarly, well-known PLP systems, which do represent CSIs qualitatively, first ground the first-order programs and then perform inference at the ground level \cite{fierens2015inference}. 
In contrast, FO-CS-LW reasons at the first-order level. Using the tools of logic, i.e., unification, substitution, and resolution, FO-CS-LW samples only relevant variables from first-order \dcplus programs determined by various forms of independencies present in the programs.
We empirically demonstrate that FO-CS-LW scales with domain size and provide an open-source implementation of our framework\footnotemark.

\footnotetext{The code is publicly available: \url{https://github.com/niteshroyal/DC-Sharp}}

This paper is a significantly extended and completed version of our previous paper \cite{kumar2021context}, where we introduced CS-LW to exploit the structures present in CPDs of BNs. The present paper first introduces a language to describe first-order probabilistic models and then extends CS-LW to the first-order case.  

\paragraph{Contribution} We summarise our contributions in this paper as follows:
\begin{itemize}
    \item We introduce a new PLP language \dcplus that supports combining rules to describe hybrid relational probabilistic models succinctly. 
    \item We present a novel sampling methodology, CS-LW, that exploits both CIs and CSIs for probabilistic inference in BNs and ground probabilistic logic programs.
    \item We present a first-order extension of CS-LW that applies directly to first-order \dcplus programs and in additional exploits the symmetries  present through combining rules.
    \item We empirically show that our inference algorithm scales with the domain size when applied to hybrid relational probabilistic models described as \dcplus programs.
\end{itemize}

\paragraph{Organization}
The paper is organized as follows. We start by motivating our discussion with some examples in Section \ref{section: motivating examples}. In Section \ref{section: background}, we review the standard likelihood weighting and basic concepts of logic programming. Section \ref{section: dc-sharp} presents the \dcplus language. Section \ref{section: cs-lw} presents the CS-LW algorithm for ground \dcplus programs describing BNs, which is then extended to first-order \dcplus programs in Section \ref{section: fo-cs-lw}. 
We then evaluate our algorithms in Section \ref{section: experiments}. Before concluding, we finally touch upon related work and directions for future work.

\section{Motivating Examples}\label{section: motivating examples}
\begin{figure}[t]
     \centering
     \begin{subfigure}[b]{0.5\linewidth}
         \centering
         \includegraphics[width=1\linewidth]{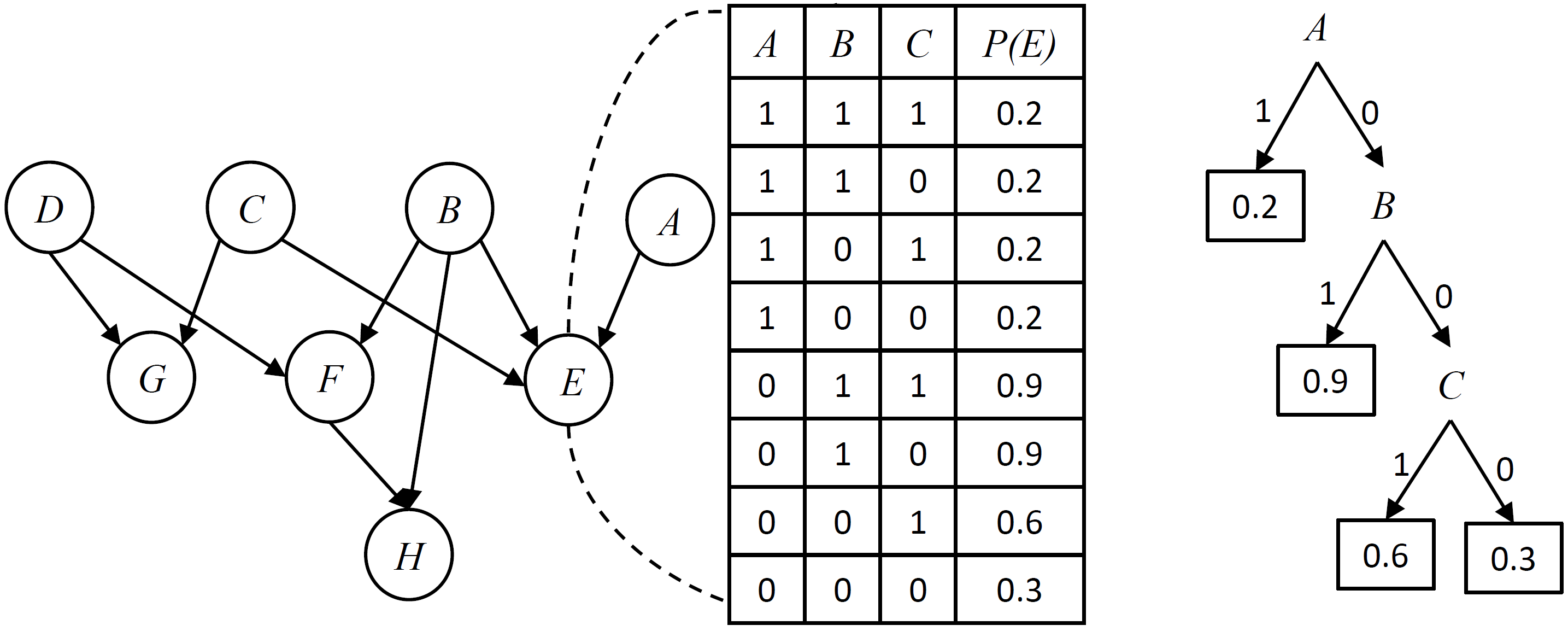}
         \caption{}
         \label{fig:context-specific independence}
     \end{subfigure}
     \hfill
     \begin{subfigure}[b]{0.45\linewidth}
         \centering
         \includegraphics[width=0.5\linewidth]{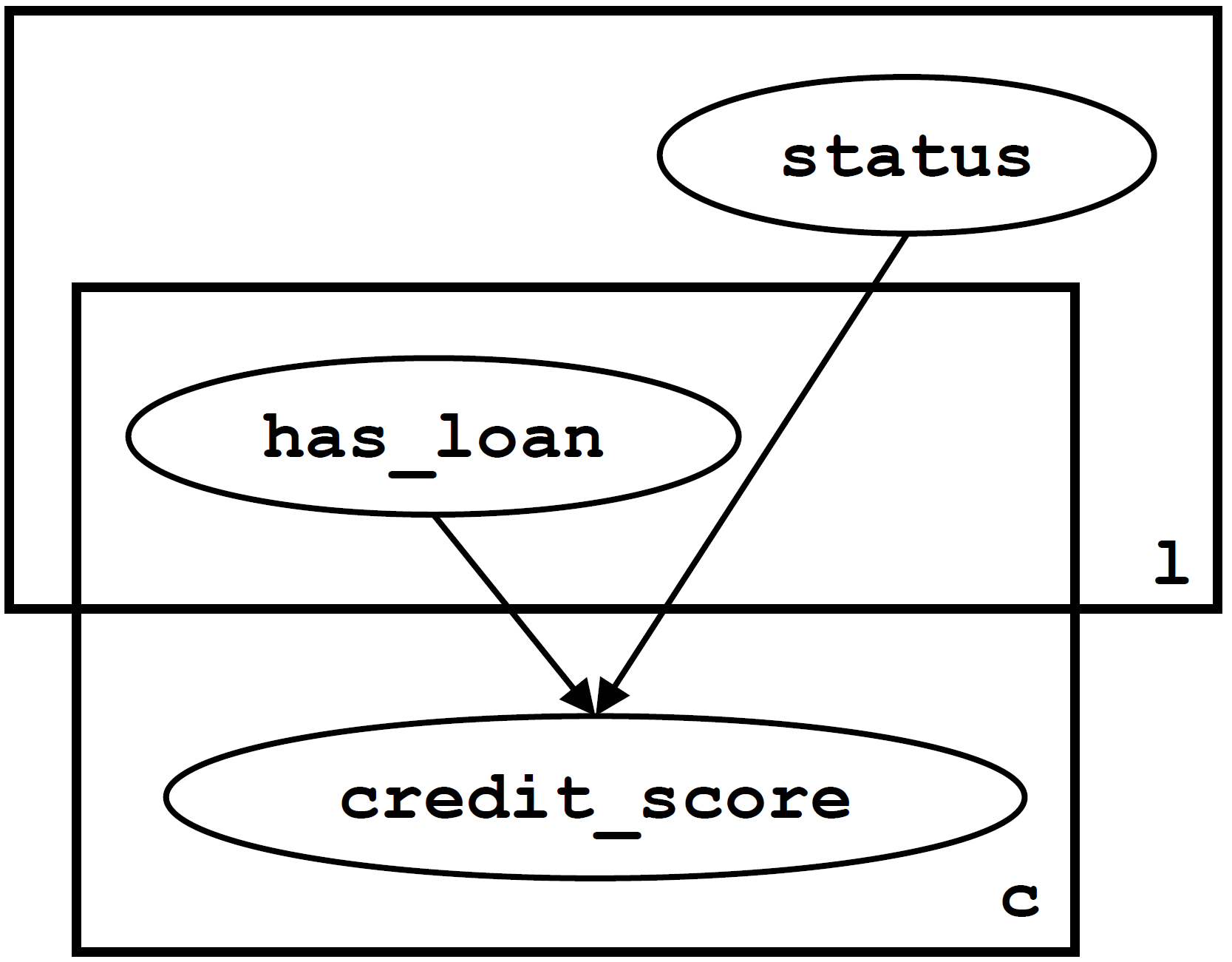}
         \caption{}
         \label{fig:plate notation}
     \end{subfigure}
     \caption{(a) Context-Specific Independence; (b) A Bayesian network with client and loan plates}
     \label{fig:three graphs}
\end{figure}

Let us illustrate, with examples, the independencies that \dcplus programs will qualitatively represent and that our algorithm will exploit. Consider a BN in Figure \ref{fig:context-specific independence}, where a tree-structure is present in the CPD of variable $E$. If one observes the CPD carefully, one can conclude that $P(E \mid A=1, B, C) = P(E \mid A=1)$, that is, $P(E \mid A=1, B, C)$ is the same for all values of $B$ and $C$. The variable $E$ is said to be independent of variables $\{B,C\}$ in the {\em context} $A=1$. This local independence statement corresponds to the influence of edges $\{B \rightarrow E, C \rightarrow E\}$ vanishing in this context; consequently, it may have global implications. For example, $E \perp B,C \mid A=1$ implies $E \perp D \mid H, A = 1$. These independencies are called CSIs. They arise naturally in various real-world situations \cite{poole2003exploiting}, including when one writes {\em if-then conditions} in probabilistic programs \cite{li2013blog}. Our CS-LW algorithm aims to exploit CSIs arising due to the structures present within CPDs of BNs when ground \dcplus programs describe such BNs.

The exploitation of CSIs in relational probabilistic models is even more crucial since a huge amount of CSIs is present in these models. As a simple example, consider the model in Figure \ref{fig:plate notation}, where the plate notation is used to represent direct influence relationships among credit scores of $\mathtt{c}$ clients, statuses of $\mathtt{l}$ loans, and $\mathtt{c} \times \mathtt{l}$ number of $\mathtt{has\_loan}$ type random variables for each client-loan pair, which can be either true or false. Given that a client has a loan, it is easy to imagine that the status of the loan may affect the client's credit score. On the other hand, if a client has only a few loans, then it is just as easy to imagine that the status of the loans that the client does not have will not affect the client's credit score. That is, the client's credit is independent of the status of all those loans that the client does not have. This is a first-order level CSI, which BNs with plates do not qualitatively represent. We will introduce a PLP language to represent it.

Furthermore, it is natural to imagine that the client's credit score depends on the number of loans that the client has with approved and rejected statuses but not on the identity of those approved and rejected loans. That is, loans are {\em exchangeable} objects. In such a case, it is common to consider that each loan's status independently has its own probabilistic influence on the client's credit score, and the final probabilistic influence is a combination of all these influences \cite{natarajan2008learning,koller2009probabilistic}. This independence assumption implies exchangeability. Relational models written as \dcplus programs qualitatively represent these independencies and the first-order level CSIs that our FO-CS-LW algorithm aims to exploit for scalability.

\section{Background}\label{section: background}
A Bayesian network $\mathcal{B}$ is a pair $(\mathcal{G}, \mathcal{D})$, where $\mathcal{G}$ is a directed acyclic graph structure specifying {\em direct influence} relationships among random variables (nodes), and $\mathcal{D}$ is a set of CPDs associated with each node. The CPD specifies the conditional probability of the variable given its parents. The graph structure $\mathcal{G}$ represents local CI statements, which states that each variable is conditionally independent of its non-descendants given its parents. The local CIs and the set of CPDs $\mathcal{D}$ together induce a joint probability distribution $P$ over all the variables. 

We denote random variables (RVs) with uppercase letters ($A$) and their assignments with lowercase letters ($a$). Bold letters denote sets of RVs ($\mathbf{A}$) and their assignments ($\mathbf{a}$). Parents of the variable $A$ are denoted with $\mathbf{Pa}(A)$ and their assignments with $\mathbf{pa}(A)$. Suppose $P$ is a probability distribution over disjoint sets of variables $\mathbf{E},\mathbf{X},\mathbf{Z}$, then $\mathbf{E}$ denotes a set of observed variables, $\mathbf{X}$ a set of unobserved query variables and $\mathbf{Z}$ a set of unobserved variable other than query variables. The expected value of $A$ relative to a distribution $Q$ is denoted by $\mathbb{E}_{Q}[A]$.

\subsection{Likelihood Weighting}\label{section: lw}
Next, we briefly review likelihood weighting (LW), one of the most popular sampling algorithms for BNs. 

A typical query to the distribution $P(\mathbf{E},\mathbf{X},\mathbf{Z})$ is to compute $P(\mathbf{x}_q \mid \mathbf{e})$, that is, the probability of $\mathbf{X}$ being assigned $\mathbf{x}_q$ given that $\mathbf{E}$ is assigned $\mathbf{e}$. Following Bayes's rule, we have: 
\begin{equation*}
    \begin{aligned}
    P(\mathbf{x}_q \mid \mathbf{e}) = \frac{P(\mathbf{x}_q, \mathbf{e})}{P(\mathbf{e})} 
    = \frac{\sum_{\mathbf{x}, \mathbf{z}} P(\mathbf{x}, \mathbf{z}, \mathbf{e}) f(\mathbf{x})}{\sum_{\mathbf{x}, \mathbf{z}} P(\mathbf{x}, \mathbf{z}, \mathbf{e})} = \mu, 
    \end{aligned}
\end{equation*}
where $f(\mathbf{x})$ is an indicator function $\mathds{1}\{\mathbf{x} = \mathbf{x}_q\}$, which takes value $1$ when $\mathbf{x} = \mathbf{x}_q$, and $0$ otherwise. We can estimate $\mu$ using LW if we specify $P$ using a Bayesian network. LW belongs to a family of importance sampling schemes that are based on the observation,
\begin{equation}\label{eq: importance sampling}
    \begin{aligned}
    \mu = \frac{ \sum_{\mathbf{x}, \mathbf{z}} Q(\mathbf{x}, \mathbf{z}, \mathbf{e}) f(\mathbf{x}) (P(\mathbf{x}, \mathbf{z}, \mathbf{e})/Q(\mathbf{x}, \mathbf{z}, \mathbf{e}))}{\sum_{\mathbf{x}, \mathbf{z}} Q(\mathbf{x}, \mathbf{z}, \mathbf{e}) (P(\mathbf{x}, \mathbf{z}, \mathbf{e})/Q(\mathbf{x}, \mathbf{z}, \mathbf{e}))},
    \end{aligned}
\end{equation}
where $Q$ is a {\em proposal distribution} such that $Q>0$ whenever $P>0$. The distribution $Q$ is different from $P$ and is used to draw independent samples. Generally, $Q$ is selected such that the samples can be drawn easily. In the case of LW, to draw a sample, variables $X_i \in \mathbf{X} \cup \mathbf{Z}$ are assigned values drawn from $P(X_i \mid \mathbf{pa}(X_i))$ and variables in $\mathbf{E}$ are assigned their observed values. These variables are assigned in a topological ordering relative to the graph structure of $\mathcal{B}$. Thus, the proposal distribution in the case of LW can be described as follows: 
\begin{equation*}
    \begin{aligned}
    Q(\mathbf{X}, \mathbf{Z}, \mathbf{E}) = \prod_{X_i \in \mathbf{X} \cup \mathbf{Z}} P(X_i \mid \mathbf{Pa}(X_i)) \mid_{\mathbf{E}=\mathbf{e}}
    \end{aligned}.
\end{equation*}
Consequently, it is easy to compute the {\em likelihood ratio}  $P(\mathbf{x}, \mathbf{z}, \mathbf{e})/Q(\mathbf{x}, \mathbf{z}, \mathbf{e})$ in Equation \ref{eq: importance sampling}. All factors in the numerator and denominator of the fraction cancel out except for $P(x_i \mid \mathbf{pa}(X_i))$ where $x_i \in \mathbf{e}$. Thus, 
\begin{equation*}
    \begin{aligned}
    \frac{P(\mathbf{X}, \mathbf{Z}, \mathbf{e})}{Q(\mathbf{X}, \mathbf{Z}, \mathbf{e})} = \prod_{x_i \in \mathbf{e}} P(x_i \mid \mathbf{Pa}(X_i)) = \prod_{x_i \in \mathbf{e}} W_{x_i} = W_{\mathbf{e}}, 
    \end{aligned}
\end{equation*}
where $W_{x_i}$, which is also a RV, is the {\em weight} of evidence $x_i$. The {\em likelihood ratio} $W_{\mathbf{e}}$ is the product of all of these weights, and thus, it is also a RV. Given $M$ independent weighted samples from $Q(\mathbf{X}, \mathbf{Z}, \mathbf{E})$, we can estimate the query: 
\begin{equation}\label{equation: Naive Estimate}
    \begin{aligned}
    \hat{\mu} = \frac{\sum_{m=1}^{M} f(\mathbf{x}[m]) w_{\mathbf{e}}[m]}{\sum_{m=1}^{M} w_{\mathbf{e}}[m] }.
    \end{aligned}
\end{equation}

\subsection{Likelihood Weighting + Bayes-Ball}\label{section: bayes-ball}

In the previous section, 
we used all random variables to estimate $\mu$. However, due to CIs encoded by the graph structure in a Bayesian network $\mathcal{B}$, observed states and CPDs of only some variables ``might'' be required for computing $\mu$. These variables are called {\em requisite variables}. To get a better estimate of $\mu$, it is recommended to use only these variables. The standard approach is to first apply the Bayes-ball algorithm \cite{shachter2013bayes} over the graph to obtain a sub-network of requisite variables, then simulate the sub-network to obtain the weighted samples. An alternative approach is to use Bayes-ball to simulate the original network $\mathcal{B}$ and focus on only requisite variables to obtain the weighted samples. This approach is trivial and might already be used by many BN inference tools. However, it is not described in the literature clearly, so we will describe it next. It will form the basis of our discussion on CS-LW, where we will also exploit structures within CPDs of BNs. 

\begin{figure}[t]
    \centering
    \includegraphics[width=0.5\linewidth]{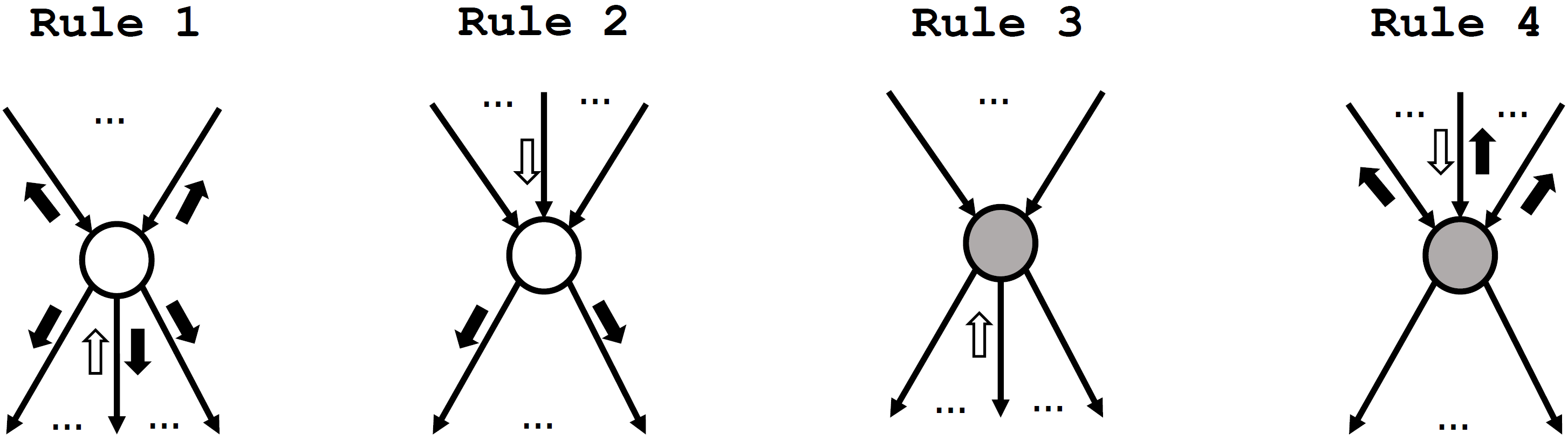}
    \caption{The four rules of Bayes-ball algorithm that decide next visits (indicated using \includegraphics[height=1.75ex]{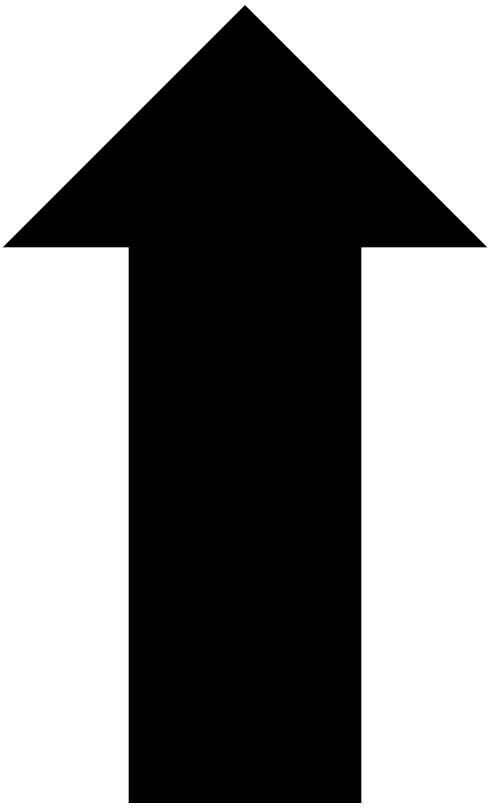}) based on the direction of the current visit (indicated using \includegraphics[height=1.75ex]{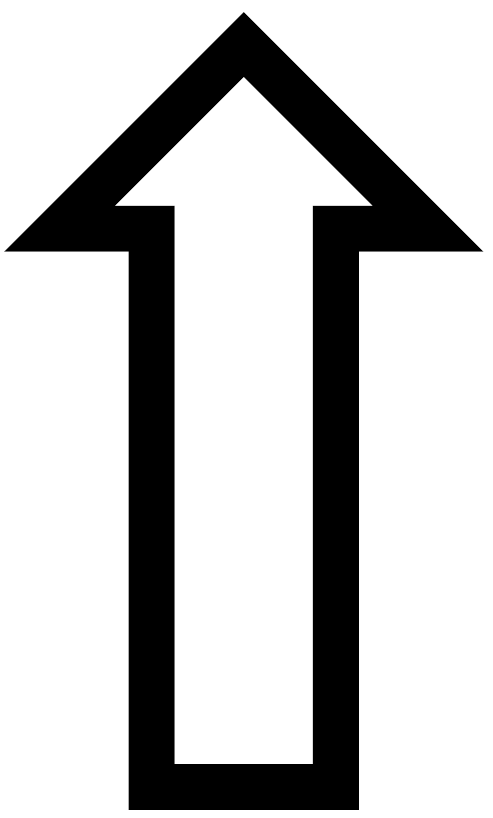}) and the type of variable. To distinguish observed variables from unobserved variables, the former type of variables are shaded.}
    \label{fig: bayes-ball rules}
\end{figure}

To obtain the samples, we need to traverse the graph in a topological ordering. The Bayes-ball algorithm, which is linear in the graph's size, can be used for it. The advantage of using Bayes-ball is that it also detects CIs; thus, it traverses only a sub-graph that depends on the query and evidence. We can also keep assigning unobserved variables, and weighting observed variables along with traversing the graph. In this way, we assign/weigh only requisite variables. The Bayes-ball algorithm uses four rules to traverse the graph (when deterministic variables are absent in $\mathcal{B}$), and marks variables to avoid repeating the same action. These rules are illustrated in Figure \ref{fig: bayes-ball rules}. Next, we discuss these rules and also indicate how to assign/weigh variables, resulting in a new algorithm that we call {\em Bayes-ball simulation of BNs}. Starting with all query variables scheduled to be visited as if from one of their children, we apply the following rules until no more variables can be visited:
\begin{enumerate}
    \item When the visit of an unobserved variable $U \in \mathbf{X} \cup \mathbf{Z}$ is from a child, and $U$ is not marked on top, then do these in the order: i) Mark $U$ on top; ii) Visit all its parents; iii) Sample a value $y$ from $P(U \mid \mathbf{pa}(U))$ and assign $y$ to $U$; iv) If $U$ is not marked on bottom, then mark $U$ on bottom and visit all its children.  
    \item When the visit of an unobserved variable is from a parent, and the variable is not marked on bottom, then mark the variable on bottom and visit all its children.
    \item When the visit of an observed variable is from a child, then do nothing. 
    \item When the visit of an observed variable $E \in \mathbf{E}$ is from a parent, and $E$ is not marked on top, then do these in the order: i) Mark $E$ on top; ii) Visit all its parents; iii) Let $e$ be an observed value of $E$ and let $w$ be the probability at $e$ according to $P(E \mid \mathbf{pa}(E))$, then the weight of $E$ is $w$.
\end{enumerate}
The above rules define an order for visiting parents and children so that variables are assigned/weighted in a topological ordering. Indeed we can define the order since the original rules for Bayes-ball do not prescribe any order. The marks record important information, and the following result holds.
\begin{lemma}\label{theorem: bayes-ball 1}
Let $\mathbf{E}_\star \subseteq \mathbf{E}$ be marked on top, $\mathbf{E}_\smwhitestar \subseteq \mathbf{E}$ be visited but not marked on top, and $\mathbf{Z}_\star \subseteq \mathbf{Z}$ be marked on top. Then the query $\mu$ can be computed as follows,
\begin{equation}\label{equation: bb}
    \begin{aligned}
    \mu = 
    \frac{\sum_{\mathbf{x}, \mathbf{z}_\star} P(\mathbf{x},  \mathbf{z}_\star, \mathbf{e}_\star \mid \mathbf{e}_\smwhitestar) f(\mathbf{x})}{\sum_{\mathbf{x}, \mathbf{z}_\star} P(\mathbf{x},  \mathbf{z}_\star, \mathbf{e}_\star \mid \mathbf{e}_\smwhitestar)}
    \end{aligned}
\end{equation}
\end{lemma}
The proof is straightforward and is present in Appendix \ref{section: missing proofs}. 
Now, since $\mathbf{X}, \mathbf{Z}_\star, \mathbf{E}_\star, \mathbf{E}_\smwhitestar$ are variables of $\mathcal{B}$ and they form a sub-network $\mathcal{B}_\star$ such that variables in $\mathbf{E}_\smwhitestar$ do not have any parent, we can write,
\begin{equation*}
\begin{aligned}
    P(\mathbf{x}, \mathbf{z}_\star, \mathbf{e}_\star \mid \mathbf{e}_\smwhitestar) = \prod_{u_i \in \mathbf{x} \cup \mathbf{z}_\star \cup \mathbf{e}_\star} P(u_i \mid \mathbf{pa}(U_i))
\end{aligned}
\end{equation*}
such that $\forall p \in \mathbf{pa}(U_i): p \in \mathbf{x} \cup \mathbf{z}_\star \cup \mathbf{e}_\star \cup \mathbf{e}_\smwhitestar$. This means the CPDs of some observed variables are not required for computing $\mu$. Now we define these variables.

\begin{definition}
The observed variables whose observed states and CPDs might be required to compute $\mu$ will be called diagnostic evidence. 
\end{definition}

\begin{definition}
The observed variables whose observed states, but not their CPDs, might be required to compute $\mu$ will be called predictive evidence.  
\end{definition}

Diagnostic evidence (denoted by $\mathbf{e}_\star$) is marked on top, while predictive evidence (denoted by $\mathbf{e}_\smwhitestar$) is visited but not marked on top. The variables $\mathbf{X}$, $\mathbf{Z}_\star$, $\mathbf{E}_\star$, $\mathbf{E}_\smwhitestar$ will be called requisite variables.
\begin{example}
Consider the network of Figure \ref{fig:context-specific independence}, and assume that our evidence is $\{D=1, F=1, G=0, H=1\}$, and our query is $\{E=0\}$. Suppose we start by visiting the query variable from its child and apply the four rules of Bayes-ball. One can easily verify that observed variables $F, G, H$ will be marked on top; hence $\{F=1, G=0, H=1\}$ is diagnostic evidence ($\mathbf{e}_\star$). The observed variable $D$ will only be visited; hence $\{D=1\}$ is predictive evidence ($\mathbf{e}_\smwhitestar$). Variables $A,B,C,E$ will be marked on top and are requisite unobserved variables ($\mathbf{X} \cup \mathbf{Z}_\star$).  
\end{example}

Now, we can sample from a factor $Q_\star$ of $Q$ such that,
\begin{equation}\label{equation: proposal distribution}
    \begin{aligned}
    Q_\star(\mathbf{X},  \mathbf{Z}_\star, \mathbf{E}_\star \mid \mathbf{E}_\smwhitestar) = \prod_{X_i \in \mathbf{X} \cup \mathbf{Z}_\star} P(X_i \mid \mathbf{Pa}(X_i)) \mid_{\mathbf{E}_\star=\mathbf{e}_\star}
    \end{aligned}
\end{equation}
When we use Bayes-ball, precisely this factor is considered for sampling. Starting by first setting $\mathbf{E}_\smwhitestar$ to their observed values, $\mathbf{X} \cup \mathbf{Z}_\star$ is assigned and $\mathbf{e}_\star$ is weighted in the topological ordering. Given $M$ weighted samples $\mathcal{D}_{\star} = \langle \mathbf{x}[1], w_{\mathbf{e}_{\star}}[1]\rangle, \dots, \langle\mathbf{x}[M],  w_{\mathbf{e}_{\star}}[M] \rangle$ from $Q_\star$, we can estimate:
\begin{equation}\label{Equation: Bayes-ball Estimate}
    \begin{aligned}
    \tilde{\mu} = \frac{ \sum_{m=1}^{M} f(\mathbf{x}[m]) w_{\mathbf{e_\star}}[m]}{\sum_{m=1}^{M} w_{\mathbf{e_\star}}[m]}.
    \end{aligned}
\end{equation}

In this way, we sample from a lower-dimensional space; thus, the new estimator $\tilde{\mu}$ has a lower variance compared to $\hat{\mu}$ due to the Rao-Blackwell theorem. Consequently, fewer samples are needed to achieve the same accuracy. Hence, for improved inference, we exploit CIs encoded by the graph structure in $\mathcal{B}$.

\subsection{Context-Specific Independence}
Next, we formally define the independencies that arise due to the structures present within CPDs, which were informally discussed in Section \ref{section: motivating examples}. 
\begin{definition}
Let $P$ be a probability distribution over variables $\mathbf{U}$, and
let $\mathbf{A}, \mathbf{B}, \mathbf{C}, \mathbf{D}$ be disjoint subsets of $\mathbf{U}$. The variables $\mathbf{A}$ and $\mathbf{B}$ are independent given $\mathbf{D}$ and context $\mathbf{c}$ if $P(\mathbf{A} \mid \mathbf{B}, \mathbf{D}, \mathbf{c}) = P(\mathbf{A} \mid \mathbf{D}, \mathbf{c})$  whenever $P(\mathbf{B}, \mathbf{D}, \mathbf{c}) > 0$. This is denoted by $\mathbf{A} \perp \mathbf{B} \mid \mathbf{D}, \mathbf{c}$. If $\mathbf{D}$ is empty then $\mathbf{A}$ and $\mathbf{B}$ are independent given context $\mathbf{c}$, denoted by $\mathbf{A} \perp \mathbf{B} \mid \mathbf{c}$. 
\end{definition}

Independence statements of the above form are called {\em context-specific independencies} (CSIs). When $\mathbf{A}$ is independent of $\mathbf{B}$ given all possible assignments to $\mathbf{C}$ then we have: $\mathbf{A} \perp \mathbf{B} \mid \mathbf{C}$. The independence statements of this form are generally referred to as {\em conditional independencies} (CIs). Thus, CSI is a more fine-grained notion than CI. The graphical structure in $\mathcal{B}$ can only represent CIs. Any CI can be verified in linear time in the size of the graph. However, verifying any arbitrary CSI has been recently shown to be coNP-hard \cite{corander2019logical}. 

\subsection{A Bit of Logic Programming}
Probabilistic logic programming is a probabilistic characterization of logic programming. So, before describing our system, in this section, we review relevant syntactic and semantic notions related to logic programming. More details can be found in \cite{nilsson1990logic}.

An {\em atom} $\mathtt{p(t_1, \dots, t_n)}$ consists of a predicate $\mathtt{p/n}$ of arity $\mathtt{n}$ and terms $\mathtt{t_1, \dots, t_n}$. A {\em term} is either a constant (written in lowercase), a variable (in uppercase), or a structured term of the form $\mathtt{f(u_1, \dots, u_k)}$ where $\mathtt{f}$ is a functor and the $\mathtt{u_i}$ are terms. For example, $\mathtt{has\_account(ann,L)}$, $\mathtt{has\_account(ann,a\_1)}$ and $\mathtt{has\_account(ann,func(A))}$ are atoms and $\mathtt{ann}$, $\mathtt{L}$, $\mathtt{a\_1}$ and $\mathtt{func(A)}$ are terms. A {\em literal} is an atom or the negation of an atom. A {\em positive literal} is an atom. A {\em negative literal} is the negation of an atom. A {\em clause} is a universally quantified disjunction of literals. A {\em definite clause} is a clause which contains exactly one positive literal and zero or more negative literals. For example, $\mathtt{\forall(A_0 \vee \neg A_1 \vee \dots \vee \neg A_n )}$ is a definite clause, where $\mathtt{A_0, A_1, \dots, A_n}$ are atoms. In logic programming, one usually writes definite clauses in the implication form $\mathtt{A_0 \leftarrow A_1, \dots, A_n}$ (where we omit the universal quantifiers for ease of writing). Here, the atom $\mathtt{A_0}$ is called {\em head} of the clause; and the set of atoms $\mathtt{\{A_1, \dots, A_n\}}$ is called {\em body} of the clause. A clause with an empty body is called a {\em fact}.
A {\em definite program} consists of a finite set of definite clauses.

\begin{example}\label{ex:ex1}
A clause  $\mathcal{C} \equiv \mathtt{has\_loan(C,L) \leftarrow has\_account(C,A), account\_loan(A,L)}$ is a definite clause. Intuitively, it states that a client $\mathtt{C}$ has a loan $\mathtt{L}$ if $\mathtt{C}$ has an account $\mathtt{A}$ and $\mathtt{A}$ is associated to the loan $\mathtt{L}$. 
\end{example}

An {\em expression}, which can be either a term, an atom or a clause, is \textit{ground} if it does not contain any variable. A {\em substitution} $\mathtt{\theta = \{V_1/t_1, ..., V_m/t_m\}}$ assigns  terms $\mathtt{t_i}$ to variables $\mathtt{V_i}$. The element $\mathtt{V_i/t_i}$ is a {\em binding} for variable $\mathtt{V_i}$. Applying $\mathtt{\theta}$ to an expression $\mathcal{E}$ yields $\mathcal{E}\theta$, the {\em instance} of $\mathcal{E}$, where all occurrences of $\mathtt{V_i}$ in $\mathcal{E}$ are replaced by the corresponding terms $\mathtt{t_i}$. A substitution $\theta$ is a {\em grounding} for $\mathcal{E}$ if $\mathcal{E\theta}$ is ground, i.e., contains no variables (when there is no risk of confusion we drop ``for $\mathcal{E}$'').

\begin{example}
Applying a substitution $\mathtt{\theta = \{C/ann\}}$ to the clause $\mathcal{C}$ from Example \ref{ex:ex1} yields $\mathcal{C\theta}$ which is  $\mathtt{has\_loan(ann,L) \leftarrow has\_account(ann,A), account\_loan(A,L)}$.  
\end{example}

A substitution $\theta$ {\em  unifies} two expressions $\mathcal{E}_1$ and $\mathcal{E}_2$ if $\mathcal{E}_1\theta$ and $\mathcal{E}_2\theta$ are identical (denoted $\mathcal{E}_1\theta = \mathcal{E}_2\theta$). Such a substitution is called a {\em unifier}. Unifiers may not always exist. If there exists a unifier for two expressions $\mathcal{E}_1$ and $\mathcal{E}_2$, we call such atoms {\em unifiable} and we say that $\mathcal{E}_1$ and $\mathcal{E}_2$ {\em unify}.
\begin{example}
A substitution $\mathtt{\theta = \{ C/ann, M/l\_1, L/l\_1 \}}$ unify $\mathtt{has\_loan(ann,L)}$ and $\mathtt{has\_loan(C,M)}$.
\end{example}
A substitution $\theta$ is said to be more general than a substitution $\sigma$ iff  there exists a substitution $\sigma^{\prime}$ such that $\sigma = \theta\sigma^{\prime}$. A unifier $\theta$ is said to be a {\em most general unifier (mgu)} of two expressions iff $\theta$ is more general than any other unifier of the expressions. Expression $\mathcal{E}_1$ is a {\em renaming} of $\mathcal{E}_2$ if they differ only in the names of variables. 
\begin{example}
Unifiers $\mathtt{\{ C/ann, M/L\}}$ and $\mathtt{\{ C/ann, L/M\}}$ are both most general unifiers of $\mathtt{has\_loan(ann,L)}$ and $\mathtt{has\_loan(C,M)}$. The resulting applications,
\begin{align*}
    & \mathtt{has\_loan(C,M)\{ C/ann, M/L\} = has\_loan(ann,L)} \\
    & \mathtt{has\_loan(ann,L)\{ C/ann, L/M\} = has\_loan(ann,M)}
\end{align*}
are renamings of each other. 
\end{example}

The {\em Herbrand universe} of a definite program $\mathcal{P}$, denoted $U_\mathcal{P}$, is the set of all ground terms constructed from functors and constants appearing in $\mathcal{P}$. The {\em Herbrand base} $B_\mathcal{P}$ is the set of all ground atoms that can be constructed by using predicates from $\mathcal{P}$ with ground terms from $U_\mathcal{P}$ as arguments. Subsets of the Herbrand base are called {\em Herbrand interpretations}. A Herbrand interpretation $\mathcal{I}$ is a model of a clause $\mathtt{A_0 \leftarrow A_1, \dots A_n}$ iff for all grounding substitutions $\theta$, such that $\{\mathtt{A_1\theta, \dots, A_n\theta}\} \subseteq \mathcal{I}$, it also holds that $\mathtt{A_0\theta} \in \mathcal{I}$. A {\em Herbrand model} of a set of clauses is a Herbrand interpretation which is a model of every clause in the set.

The {\em least Herbrand model} of a definite program $\mathcal{P}$, denoted $M_\mathcal{P}$, is the intersection of all Herbrand models of $\mathcal{P}$, i.e., $M_\mathcal{P}$ is the set of all ground atoms that are logical consequences of the program. $M_\mathcal{P}$ is unique for definite programs and can be constructed by repeatedly applying the so-called $T_\mathcal{P}$ operator, which is defined as a function on Herbrand interpretations of $\mathcal{P}$ as follows:
\begin{equation*}
    T_\mathcal{P}(\mathcal{I}) := \{\mathtt{A_0\theta \mid A_0 \leftarrow A_1, \dots, A_n} \in \mathcal{P} \wedge \mathtt{\{A_1\theta, \dots, A_n\theta\}} \subseteq \mathcal{I} \},
\end{equation*}
where, $\theta$ are grounding substitutions. Let $\mathcal{I}_1$ be the set of all ground facts in the program. Now, applying the operator on $\mathcal{I}_1$, it is possible to use every ground instance of each clause to construct new ground atoms from $\mathcal{I}_1$. In this way, a new set $\mathcal{I}_2 := T_\mathcal{P}(\mathcal{I}_1)$ is obtained, which can be used again to construct more ground atoms. The new atoms added to $\mathcal{I}_{i+1}$ are those which must follow immediately from $\mathcal{I}_{i}$. It is possible to construct $M_\mathcal{P}$ by recursively applying the operator until a {\em fixpoint} is reached ($T_\mathcal{P}(M_\mathcal{P}) = M_\mathcal{P}$), i.e., until no more ground atoms can be constructed. 

A {\em query } $\mathtt{Q}$ is of the form $\mathtt{B_1, \dots, B_m}$ where the $\mathtt{B_j}$ are atoms and all variables are understood to be existentially quantified. Given a definite program $\mathcal{P}$, a correct answer to the query $\mathtt{Q}$ is a substitution $\theta$ such that $\mathtt{Q\theta}$ is entailed by $\mathcal{P}$, denoted by $\mathcal{P} \models \mathtt{Q\theta}$. That is, $\mathtt{Q\theta}$ belongs to $M_\mathcal{P}$. The answer substitution $\theta$ is often computed using {\em SLD-resolution}. Finally, the answer set of $\mathtt{Q}$ is the set of all correct answer substitutions to $\mathtt{Q}$.

\section{\dcplus: A Representation Language for Hybrid Relational Models}\label{section: dc-sharp}
This section presents a PLP language \dcplus for describing hybrid relational probabilistic models. The syntax of our language is based on the elegant syntax of distributional clauses (DCs) used by \cite{nitti2016probabilistic}. However, we extend its semantics to support combining rules.
The new semantics, however, do not allow for describing open-universe probabilistic models (OUPMs) \cite{milch2010extending}, which was possible in the previous system. The semantics supporting both OUPMs and combining rules is another topic of research. We do not study it in this paper. The new semantics have been developed along the lines of Bayesian Logic Programs (BLPs) \cite{kersting20071}. 

\subsection{Syntax}\label{section: syntax}
DC is a natural extension of definite clauses for representing conditional probability distributions.

\begin{definition}
A DC is a formula of the form $\mathtt{A_0 \sim D \leftarrow A_1, \dots, A_n}$, where $\mathtt{\sim}$ is a special binary predicate used in infix notation, and $\mathtt{A_i}(i>0)$ are atoms. Term $\mathtt{A_0}$ is called a random variable term and $\mathtt{D}$ is called a distributional term.
\end{definition}

Intuitively, the clause states that RV $\mathtt{A_0\theta}$ is distributed as $\mathtt{D\theta}$ whenever all $\mathtt{A_i\theta}$ are true for a grounding substitution $\theta$.
The ground terms $\mathtt{A_0\theta}$ and $\mathtt{D\theta}$ belong to the Herbrand universe. 

Ground RV terms are interpreted as RVs. To refer to the values of RV terms, we use a binary predicate $\cong$, which is used in infix notation for convenience. A ground atom $\mathtt{A\theta \cong x}$ is defined to be true if $\mathtt{x}$ is the value of RV (or ground RV term) $\mathtt{A\theta}$.

\begin{example}
Consider the following clause,
\begin{equation*}
\mathtt{credit\_score(C) \sim gaussian(755.5,0.1) \leftarrow has\_loan(C,L)\cong true, status(L) \cong appr.}
\end{equation*}

\noindent
Applying a grounding substitution $\mathtt{\{C/ann,L/l\_1\}}$ to the clause results in defining a RV $\mathtt{credit\_score(ann)}$ distributed as $\mathtt{gaussian(755.5, 0.1)}$ when RVs $\mathtt{has\_loan(ann,l\_1)}$ and $\mathtt{status(l\_1)}$ take values $\mathtt{true}$ and $\mathtt{appr}$ (``approved'') respectively; that is, when atoms $\mathtt{has\_loan(ann,l\_1)\cong true}$  and $\mathtt{status(l\_1) \cong appr}$ are true.
\end{example}
A  DC without body is called a {\em probabilistic fact}, for example: 
\begin{equation*}
    \mathtt{age(bob) \sim gaussian(40,0.2).}
\end{equation*}
This clause states that the age of $\mathtt{bob}$ is normally distributed with mean $40$ and variance $0.2$. It is also possible to define deterministic variables that take only one value with probability 1, e.g., to express our absolute certainty that the age of $\mathtt{bob}$ takes value $40$, we can write,
\begin{equation*}
    \mathtt{\mathtt{age(bob) \sim val(40).}}
\end{equation*}
Hence, it is also possible to write the definite clause $\mathcal{C}$ of Example \ref{ex:ex1} as a DC like this:
\begin{equation*}
    \mathtt{\mathtt{has\_loan(C,L) \sim val(true) \leftarrow has\_account(C,A)\cong true, account\_loan(A,L)\cong true}}
\end{equation*}

DC also supports comparing the values of RVs with constants or with values of other RVs in the ground instance of clauses. This can be done using the binary infix predicates $\mathtt{==, <, >, \geq, and \leq}$, which are especially useful while conditioning continuous RVs as illustrated by the following clause:
\begin{equation*}
\mathtt{credit\_score(C) \sim gaussian(645.5,0.1) \leftarrow age(C) \cong X, X<40.}
\end{equation*}

\noindent
This clause specifies the distribution of a client's credit score when the client's age is less than $40$.

A distributional program $\mathbb{P}$ consists of a set of distributional clauses.

\begin{example}\label{example: DC example with CR}
The following program describes probabilistic influences among attributes of clients and related loans. Here, we have used $\mathtt{t}$ as a shorthand for $\mathtt{true}$, $\mathtt{f}$ for $\mathtt{false}$, $\mathtt{a}$ for approved, and $\mathtt{d}$ for declined.
{\small
\begin{align*}
    & \mathtt{client(ann) \sim val(t).}\\
    & \mathtt{loan(l\_1) \sim val(t).} \\
    & \mathtt{loan(l\_2) \sim val(t).} \\
    & \mathtt{has\_loan(C,L) \sim bernoulli(0.2) \leftarrow client(C)\cong t, loan(L)\cong t.} \\
    & \mathtt{status(L) \sim discrete([0.3:a, 0.7:d]) \leftarrow loan(L)\cong t.} \\
    & \mathtt{credit\_score(C) \sim gaussian(650, 15.4) \leftarrow has\_loan(C,L)\cong Y, Y==f.} \\
    & \mathtt{credit\_score(C) \sim gaussian(700, 10.9) \leftarrow has\_loan(C,L)\cong  t, status(L)\cong X, X== a.}\\
    & \mathtt{credit\_score(C) \sim gaussian(600, 20.5) \leftarrow has\_loan(C,L)\cong  t, status(L)\cong X, X==d.}
\end{align*}
}%
\end{example}

Now, we provide the semantics of such a program. 

\subsection{Semantics}\label{Section: sematics}
First, we specify the form of DCs allowed in our programs.  

\begin{definition}\label{definition: dc sharp}
A \dcplus program is a finite set of DCs $\mathtt{A_0 \sim D \leftarrow A_1, \dots, A_n}$ whose atoms $\mathtt{A_i (i>0)}$ are either of the following form\footnotemark: 
\begin{enumerate}
    \item\label{enum item: conditional on variables} $\mathtt{q(t_1, \dots, t_k) \cong V}$, where $\mathtt{q(t_1, \dots, t_k)}$ is a RV term and $\mathtt{V}$ can either be a variable or a constant belonging to the domain of the RV. If $\mathtt{V}$ is a variable then it must not appear in RV terms of the DC.
    \item $\mathtt{V_1 \diamond V_2}$, where $\mathtt{V_1, V_2}$ can be logical variables or constants belonging to the corresponding domains of RVs, and $\diamond$ is a comparison infix predicate that can be either of these:  $==, <, >, \geq, \leq$. The predicates $\diamond$ have the same meaning as they have in 
    Prolog.
\end{enumerate}
\end{definition}
\footnotetext{Even though only two forms of atoms are allowed, one can still write a deterministic atom $\mathtt{u(t_1, \dots, t_n)}$ as in definite clauses like this: $\mathtt{u(t_1, \dots, t_n)\cong true}$. So, by restricting the form of allowed atoms, we are not losing expressivity. Recall that a definite clause can be expressed as a DC.}

For example, all clauses shown in Section \ref{section: syntax} are already of this form, and the program in Example \ref{example: DC example with CR} is a \dcplus program. The reason why logical variables $\mathtt{V_i}$, in the above-stated form (Condition \ref{enum item: conditional on variables}), are not allowed to appear in RV terms is that the existence of RVs is not uncertain in \dcplus programs. This is not the case in the following program:
\begin{align*}
    & \mathtt{loan\_id \sim poisson(10).}\\
    & \mathtt{status(L) \sim discrete([0.3:a, 0.7:d]) \leftarrow loan\_id\cong L.}
\end{align*}
Here, the first clause models the identity of loans as a Poisson distribution with a mean of $10$. Thus, $\mathtt{loan\_id}$ can be any natural number starting with $0$, and the existence of the statuses of loans, say $\mathtt{status(12)}$, is uncertain. We disallow writing such open-universe models in the \dcplus framework because identifying RVs in such models may require analyzing the programs dynamically. It is not clear how to exploit CSIs in such a case.
For closed-universes, they can be identified using simple static analysis, which we describe next. 

\begin{definition}\label{definition: RV set}
Let $\mathbb{P}$ be a \dcplus program. An RV set of the program, denoted $rv(\mathbb{P})$, is the set of definite clauses obtained by transforming each clause $\mathtt{A_0 \sim D \leftarrow A_1, \dots, A_n} \in \mathbb{P}$ as follows:
\begin{enumerate}
    \item Let $\mathtt{Body}$ be the empty set. 
    \item For each atom $\mathtt{A_i (i>0)}$ of the form $\mathtt{q(t_1, \dots, t_k)\cong V}$, an atom $\mathtt{rv(q(t_1, \dots, t_k))}$ is added to $\mathtt{Body}$.
    \item A clause $\mathtt{rv(A_0) \leftarrow Body}$ is added to $rv(\mathbb{P})$. 
\end{enumerate}
\end{definition}
Notice that we ignore comparison atoms while constructing the RV set since they do not contain RV terms. They only deal with the values of RVs.
\begin{example}\label{example: rv set}
The RV set for the program in Example \ref{example: DC example with CR} is:
{\small \begin{align*}
    & \mathtt{rv(client(ann)).}\\
    & \mathtt{rv(loan(l\_1)).} \\
    & \mathtt{rv(loan(l\_2)).} \\
    & \mathtt{rv(has\_loan(C,L)) \leftarrow rv(client(C)), rv(loan(L)).} \\
    & \mathtt{rv(status(L)) \leftarrow rv(loan(L)).} \\
    & \mathtt{rv(credit\_score(C)) \leftarrow rv(has\_loan(C,L)).} \\
    & \mathtt{rv(credit\_score(C)) \leftarrow rv(has\_loan(C,L)), rv(status(L)).} 
\end{align*}}
\end{example}

Atom $\mathtt{rv(has\_loan(ann,l\_1))}$ is in the least Herbrand model of the above RV set but $\mathtt{rv(has\_loan(l\_1,l\_2))}$ is not. So, term $\mathtt{has\_loan(ann,l\_1)}$ is identified as a RV but the meaningless term $\mathtt{has\_loan(l\_1,l\_2)}$ is not, even though it belongs to the Herbrand universe. Recall that a ground instance of a DC defines a RV only when its body is true. No ground DC defines $\mathtt{has\_loan(l\_1,l\_2)}$ as a RV since term $\mathtt{l\_1}$ is a loan and not a client.
We will show that the RV set of a program identifies all RVs defined by the program. This is similar to Bayesian clauses in Bayesian logic programs, where atoms in the least Herbrand model of Bayesian clauses are RVs over which a probability distribution is defined \cite{kersting20071}.

Now, it is possible to ground a \dcplus program $\mathbb{P}$ given an assignment $\mathbf{u}$ of RVs. We denote such ground programs by $ground(\mathbb{P})_\mathbf{u}$.
\begin{example}\label{example: ground program}
Given the following assignment of RVs identified from the RV set (Example \ref{example: rv set}), 
{\small
\begin{align*}
    & \mathtt{\mathbf{u} = \mathtt{\{client(ann)\cong t, loan(l\_1)\cong t, loan(l\_2) \cong t, has\_loan(ann,l\_1) \cong t,}}\\
    & \mathtt{has\_loan(ann,l\_2) \cong t, status(l\_1) \cong a, status(l\_2) \cong d, credit\_score(ann) \cong 601.2\}},
\end{align*}
}%
the program of Example \ref{example: DC example with CR} grounds with respect to the assignments like this:
{\small
\begin{align*}
    & \mathtt{client(ann) \sim val(t).}\\
    & \mathtt{loan(l\_1) \sim val(t).} \\
    & \mathtt{loan(l\_2) \sim val(t).} \\
    & \mathtt{has\_loan(ann,l\_1) \sim bernoulli(0.2) \leftarrow client(ann)\cong t, loan(l\_1)\cong t.} \\
    & \mathtt{has\_loan(ann,l\_2) \sim bernoulli(0.2) \leftarrow client(ann)\cong t, loan(l\_2)\cong t.} \\
    & \mathtt{status(l\_1) \sim discrete([0.3:a, 0.7:d]) \leftarrow loan(l\_1)\cong t.} \\
    & \mathtt{status(l\_2) \sim discrete([0.3:a, 0.7:d]) \leftarrow loan(l\_2)\cong t.} \\
    & \mathtt{credit\_score(ann) \sim gaussian(650, 15.4) \leftarrow has\_loan(ann,l\_1)\cong t, t==f.} \\
    & \mathtt{credit\_score(ann) \sim gaussian(650, 15.4) \leftarrow has\_loan(ann,l\_2)\cong t, t==f.} \\
    & \mathtt{credit\_score(ann) \sim gaussian(700, 10.9) \leftarrow has\_loan(ann,l\_1)\cong  t, status(l\_1)\cong a, a== a.}\\
    & \mathtt{credit\_score(ann) \sim gaussian(700, 10.9) \leftarrow has\_loan(ann,l\_2)\cong  t, status(l\_2)\cong d, d== a.}\\
    & \mathtt{credit\_score(ann) \sim gaussian(600, 20.5) \leftarrow has\_loan(ann,l\_1)\cong  t, status(l\_1)\cong a, a==d.}\\
    & \mathtt{credit\_score(ann) \sim gaussian(600, 20.5) \leftarrow has\_loan(ann,l\_2)\cong  t, status(l\_2)\cong d, d==d.}
\end{align*}
}%
Notice that neither meaningless terms like $\mathtt{has\_loan(l\_1,l\_2)}$ appear in this ground program nor atoms like $\mathtt{has\_loan(ann,l\_1)\cong f}$ appear.
\end{example}

One might wonder why we require RVs and their assignment to be {\em given} before grounding \dcplus programs. There are two main reasons. First, the programs may define continuous RVs that can take infinitely many values, so we would be constructing infinitely large ground programs without knowing the values of RVs. Second, we do not want meaningless terms like $\mathtt{has\_loan(l\_1, l\_2)}$ to appear in the head of clauses in the ground programs. These clauses define RVs, and it does not make sense to treat these meaningless terms as RVs. So, we should know RVs before grounding the programs.

Furthermore, the assignment $\mathbf{u}$ can be thought of as asserted facts in the ground programs, which decide the truth values of atoms of the form $\mathtt{X \cong Y}$ in the body of clauses. Since the truth values of bodies of clauses depend on the assignment $\mathbf{u}$ of RV terms, when the body of a clause is true, we say that the body is true {\em with respect to} $\mathbf{u}$.

Clearly, the next step is to identify direct influence relationships among RVs defined by programs.
\begin{definition}\label{definition: direct influence}
Let $\mathbb{P}$ be a \dcplus program, $\mathbf{u}$ be an assignment of RVs, 
and $ground(\mathbb{P})_\mathbf{u}$ be a ground program constructed given $\mathbf{u}$. A ground RV term $\mathtt{A}$ directly influences $\mathtt{B}$ if there is clause $\mathtt{B \sim D \leftarrow B_1, \dots, B_n} \in ground(\mathbb{P})_\mathbf{u}$ for some $\mathbf{u}$ such that $\mathtt{n > 0}$, $\mathtt{A}$ is in the body of the clause, and the body is true with respect to $\mathbf{u}$.
\end{definition}


For example, we observe from the ground program of Example \ref{example: ground program} that ground RV terms $\mathtt{has\_loan(ann,l\_1)}$, $\mathtt{has\_loan(ann,l\_2)}$, $\mathtt{status(l\_1)}$, and $\mathtt{status(l\_2)}$ directly influence $\mathtt{credit\_score(ann)}$. However, these relationships are defined with respect to assignments $\mathbf{u}$, and it is hard to construct ground programs with respect to all $\mathbf{u}$ for identifying these relationships. For this purpose, we need a simple approach that can be implemented and executed efficiently. Next, we present such an approach. It performs a simple transformation of the RV sets of \dcplus programs.

\begin{definition}\label{definition: dependency set}
Let $rv(\mathbb{P})$ be the RV set of a \dcplus program $\mathbb{P}$. The dependency set $dep(\mathbb{P})$ is the union of $rv(\mathbb{P})$ and the set of definite clauses $pa(\mathbb{P})$ obtained by transforming each clause $\mathtt{rv(A_0) \leftarrow rv(A_1), \dots, rv(A_n)} \in rv(\mathbb{P})$ with non empty body as follows: 
\begin{itemize}
    \item For each $\mathtt{rv(A_i)}$, such that $i > 0$, a clause $\mathtt{pa(A_0, A_i) \leftarrow rv(A_0), rv(A_1), \dots, rv(A_n)}$ is added to $pa(\mathbb{P})$.
\end{itemize}
\end{definition}

\begin{example}\label{example: dependency set}
By transforming the RV set constructed in Example $\ref{example: rv set}$, we obtain the dependency set of Example \ref{example: DC example with CR} that consists of the RV set and the following extra clauses:
{\small \begin{align*}
    & \mathtt{pa(has\_loan(C,L), client(C)) \leftarrow rv(has\_loan(C,L)), rv(client(C)), rv(loan(L)).} \\
    & \mathtt{pa(has\_loan(C,L), loan(L)) \leftarrow rv(has\_loan(C,L)), rv(client(C)), rv(loan(L)).} \\
    & \mathtt{pa(status(L), loan(L)) \leftarrow rv(status(L)), rv(loan(L)).} \\
    & \mathtt{pa(credit\_score(C), has\_loan(C,L)) \leftarrow rv(credit\_score(C)), rv(has\_loan(C,L)).} \\
    & \mathtt{pa(credit\_score(C), has\_loan(C,L)) \leftarrow rv(credit\_score(C)), rv(has\_loan(C,L)), rv(status(L)).}\\
    & \mathtt{pa(credit\_score(C), status(L)) \leftarrow rv(credit\_score(C)), rv(has\_loan(C,L)), rv(status(L)).}
\end{align*}}
\end{example}
One can easily infer from the above definite program that $\mathtt{status(l\_1)}$ directly influences $\mathtt{credit\_score(ann)}$ since $\mathtt{pa(credit\_score(ann), status(l\_1))}$ is entailed by the above program. Basically, $\mathtt{pa(A,B)}$ states that the parent of $\mathtt{A}$ is $\mathtt{B}$. 
Note that Definition \ref{definition: direct influence} defines the direct influences and Definition \ref{definition: dependency set} presents an approach to identify them efficiently. The dependency set is similar to the dependency graph in BLPs \cite{kersting20071}. The only difference is that we use definite clauses instead of graphs to represent direct influences. These clauses can be automatically constructed from \dcplus programs using the transformations.

However, it is still unclear how to interpret the clauses, in the ground programs, which may specify multiple distributions for RVs. 

\begin{definition}
Let $ground(\mathbb{P})_{\mathbf{u}}$ be a ground \dcplus program constructed given an assignment $\mathbf{u}$, and let 
\begin{align*}
    & \mathtt{A_0 \sim D_{A_{10}} \leftarrow A_{11}, \dots, A_{1n_1}}\\
    & \qquad \qquad \quad \vdots \\
    & \mathtt{A_0 \sim D_{A_{k0}} \leftarrow A_{k1}, \dots, A_{kn_k}}
\end{align*}
be $k$ clauses for $\mathtt{A_0}$ in $ground(\mathbb{P})_\mathbf{u}$, whose bodies are true with respect to $\mathbf{u}$. Then, there is a multiset of distributions $\mathtt{[D_{A_{10}}, \dots, D_{A_{k0}}]}$ specified for $\mathtt{A_0}$ in $ground(\mathbb{P})_\mathbf{u}$. If $\mathtt{k>1}$ for some $\mathtt{A}_0$ then we say the clauses in $ground(\mathbb{P})_\mathbf{u}$ are mutually inclusive; otherwise, they are mutually exclusive. 
\end{definition}

For example, there are two distributions specified for $\mathtt{credit\_score(ann)}$ in the above example since client $\mathtt{ann}$ has two loans and these loans make the bodies of two clauses for $\mathtt{credit\_score(ann)}$ true according to the used assignment of RVs. This, however, raises several questions: 
According to which distribution is the RV distributed? How to describe that multiple loans probabilistically influence the credit score using DCs?

The standard answer to these questions is to combine multiple distributions into a single distribution using so-called {\em combining rules} \cite{ngo1995probabilistic,kersting2001adaptive,jaeger2007parameter,natarajan2008learning}. Combining rules are based on the assumption of {\em independence of causal influence} (ICI), where it is assumed that multiple causes on a target variable can be decomposed into several independent causes whose effects are combined to yield a final value \cite{Zhang1996ExploitingCI}. In terms of the above example, this means that we can let each loan independently define a distribution for the credit score and then somehow combine all the defined distributions into a single distribution using a combining rule $\mathcal{CR}$. The most commonly used combining rules in relational systems are Mean, and NoisyOR \cite{kersting20071,natarajan2008learning,fierens2015inference}.

Let $[\mathcal{D}_1, \dots, \mathcal{D}_n]$ be a non-empty multiset of probability distributions. Then, the mean combining rule is defined as follows:
\begin{equation*}
    \mathrm{Mean}([\mathcal{D}_1, \dots, \mathcal{D}_n]) = \dfrac{1}{m}\sum_{i=1}^{n} \mathcal{D}_n
\end{equation*}
where the right-hand side of the equation is a mixture of distributions. When all distributions in the multiset are Bernoulli distributions, denoted $\mathtt{bernoulli(p_i)}$, with parameters $\mathtt{p_i}$, then generally noisy OR combining rule is used, which is defined as follows:
\begin{equation*}
    \mathrm{NoisyOR}(\mathtt{[bernoulli(p_1), \dots, bernoulli(p_n)]}) = \mathtt{bernoulli}(1- \prod_{i=1}^{n} (1-\mathtt{p_i}))
\end{equation*}

To specify a proper probability distribution, a \dcplus program has to be {\em well-defined}.
\begin{definition}
Let $\mathbb{P}$ be a \dcplus program, $\mathbf{V}_\mathbb{P}$ be the set of all RVs identified from $rv(\mathbb{P})$, $\omega$ be an assignment of $\mathbf{V}_\mathbb{P}$, and $\Omega$ be the set of all such $\omega$. The program $\mathbb{P}$ is well-defined if it satisfies the following conditions:
\begin{enumerate}
    \item {\em Exhaustiveness:} For all $\mathtt{A_0} \in \mathbf{V}_\mathbb{P}$ and for all $\omega \in \Omega$, there is at least one clause of the form $\mathtt{A_0 \sim D \leftarrow A_1, \dots, A_n}$ in $ground(\mathbb{P})_\omega$ such that the clause's body is true with respect to $\omega$. 
    \item {\em Acyclicity:} There exists a function $rank(.)$ that maps $\mathtt{A} \in \mathbf{V}_\mathbb{P}$ to natural numbers $\mathbb{N}$. Let $\mathtt{A_0 \sim D \leftarrow A_1, \dots, A_n}$ be a clause in $ground(\mathbb{P})_\omega$, and let $\{\mathtt{rv_1, \dots rv_m}\}$ be the set of RV terms in $\{\mathtt{A_1, \dots, A_n}\}$. For all $\omega$, for all clauses in $ground(\mathbb{P})_\omega$, and for all $i$, $rank(\mathtt{A_0}) > rank(\mathtt{rv_i})$.
    \item {\em Finite Support:} The set $\mathbf{V}_\mathbb{P}$ is non-empty and each $\mathtt{A_0} \in \mathbf{V}_\mathbb{P}$ is directly influenced by a finite set of RVs. 
\end{enumerate}
\end{definition}

These conditions are similar to the conditions imposed on BLPs for them to be well-defined \cite{kersting20071}. Since BLPs consist of Bayesian clauses along with CPDs, the exhaustiveness condition is implicitly imposed. Recall that CPDs define a conditional probability distribution for an RV given any possible assignment of the RV's parents, which means CPDs are exhaustive. 


Let us investigate some {\em ill-defined} programs.
\begin{example}\label{example: ill-defined}
The program
\begin{equation*}
    \mathtt{a(X) \sim bernoulli(0.2) \leftarrow b(X)\cong t .}
\end{equation*}
is not well-defined since the least Herbrand model of the RV set of the program is empty. The program
\begin{align*}
    & \mathtt{s(a,b) \sim val(t).}\\
    & \mathtt{s(X,f(Y)) \sim val(t) \leftarrow s(X,Y)\cong t.}\\
    & \mathtt{r(X) \sim val(t) \leftarrow s(X,f(Y))\cong t.}
\end{align*}
is ill-defined because $\mathtt{r(a)}$ is influenced by infinite number of RVs: $\mathtt{s(a,b)}$, $\mathtt{s(a,f(b))}$, $\mathtt{s(a,f(f(b)))}$, and so on. However, the following program 
\begin{align*}
    & \mathtt{s(a,b) \sim val(t).}\\
    & \mathtt{s(X,f(Y)) \sim val(t) \leftarrow s(X,Y)\cong t.}
\end{align*}
is well-defined since each RV is influenced by a finite number of RVs. Such programs allow for writing models that can have infinite RVs such as hidden Markov models.
The following program violates the exhaustiveness condition.
\begin{align*}
    & \mathtt{a(1) \sim discrete([0.2:t, 0.8:f]).}\\
    & \mathtt{b(1) \sim bernoulli(0.6) \leftarrow a(1)\cong t.}
\end{align*}
This is because the distribution of $\mathtt{b(1)}$ when $\mathtt{a(1)}$ is $\mathtt{f}$ (``false'') is undefined.
The program
\begin{align*}
    & \mathtt{a(1) \sim discrete([0.2:t, 0.8:f]).}\\
    & \mathtt{a(1) \sim discrete([0.1:t, 0.9:f]) \leftarrow a(1)\cong t.}\\
    & \mathtt{a(1) \sim discrete([0.7:t, 0.3:f]) \leftarrow a(1)\cong f.}
\end{align*}
is ill-defined as well since it has a cyclic dependency.  
\end{example}

We can show that the RV set of a well-defined \dcplus program identifies all RVs defined by the program, and the dependency set all direct influences among RVs. 
\begin{proposition}\label{theorem: RV set}
Let $rv(\mathbb{P})$ be a RV set of a well-defined \dcplus program $\mathbb{P}$. Then, $\mathbb{P}$ defines $\mathtt{A}$ as a RV iff
$\mathtt{rv(A)}$ is in the least Herbrand model of $rv(\mathbb{P})$.
\end{proposition}
The proofs of all results presented in this paper are in Appendix \ref{section: missing proofs}.
\begin{proposition}\label{theorem: dep set}
Let $dep(\mathbb{P})$ be a dependency set of a well-defined \dcplus program $\mathbb{P}$. Then, $\mathtt{A}$ directly influences $\mathtt{B}$ iff $\mathtt{pa(B,A)}$ is in the least Herbrand model of $dep(\mathbb{P})$. 
\end{proposition}

A {\em possible world} is an assignment of all RVs $\mathbf{V}_\mathbb{P}$ defined by a program. Since $\mathbf{V}_\mathbb{P}$ can be an infinite set, as discussed in Example \ref{example: ill-defined}, defining how probabilities are assigned to possible worlds is non-trivial \footnotemark. However, we can still define how probabilities are assigned to assignments of a certain subset of RVs. 

\footnotetext{To define a probability distribution over infinite RVs, one uses the Kolmogorov extension theorem.}


\begin{definition}
Let $\mathbf{V}_\mathbb{P}$ be the set of all RVs defined by a well-defined \dcplus program $\mathbb{P}$, $\mathbf{u}$ be an assignment of a finite subset $\mathbf{U} \subseteq \mathbf{V}_\mathbb{P}$. Denote the set of RVs that directly influence $X$ by $\mathbf{Pa}(X)$. The assignment $\mathbf{u}$ is said to be closed under direct influence relationships if $X \in \mathbf{U}$ implies $\mathbf{Pa}(X) \subseteq \mathbf{U}$.
\end{definition}

\begin{example}
Reconsider the \dcplus program of Example \ref{example: DC example with CR}. Partial assignment $\mathbf{u}_1 =$ $\{ \mathtt{client(ann)\cong t,}$ $\mathtt{loan(l\_1)\cong t,}$ $\mathtt{has\_loan(ann,l\_1)\cong f}$ $ \}$ is closed under direct influence relationships, but partial assignment $\mathbf{u}_2 =$ $\{ \mathtt{credit\_score(ann)\cong 651.2, }$ $\mathtt{client(ann)\cong t, }$  $\mathtt{loan(l\_1)\cong t, }$ $\mathtt{has\_loan(ann,l\_1)\cong f} \}$  is not closed under such relationships. This is because random variable term $\mathtt{has\_loan(ann,l\_2)}$ directly influences $\mathtt{credit\_score(ann)}$, but is not assigned in $\mathbf{u}_2$.
\end{example}

\begin{definition}
Let $\mathbb{P}$ be a well-defined \dcplus program, $\mathbf{u}$ be an assignment that is closed under direct influence relationships, and $ground(\mathbb{P})_{\mathbf{u}}$ be the ground program constructed given $\mathbf{u}$. Then the probability (or density) that $ground(\mathbb{P})_{\mathbf{u}}$ assigns to $\mathbf{u}$ is given by
\begin{equation*}
    P(\mathbf{u}) = \prod_{\mathtt{A_i\cong x} \in \mathbf{u}} {\mathcal{CR}([\mathtt{D}_{\mathtt{A_{1i}}}, \dots, \mathtt{D}_{\mathtt{A_{ki}}}])}(\mathtt{A_i\cong x})
\end{equation*}
where $[\mathtt{D}_{\mathtt{A_{1i}}}, \dots, \mathtt{D}_{\mathtt{A_{ki}}}]$ is a multiset of distributions specified for RV term $\mathtt{A_i}$ in $ground(\mathbb{P})_{\mathbf{u}}$ and ${\mathcal{CR}([\mathtt{D}_{\mathtt{A_{1i}}}, \dots, \mathtt{D}_{\mathtt{A_{ki}}}])}(\mathtt{A_i\cong x})$ is the conditional probability density/mass of $\mathtt{A_i}$ at $\mathtt{x}$ according to the combined distribution ${\mathcal{CR}([\mathtt{D}_{\mathtt{A_{1i}}}, \dots, \mathtt{D}_{\mathtt{A_{ki}}}])}$ obtained after applying $\mathcal{CR}$ on the multiset.
\end{definition}

\begin{example}
Suppose we use Mean as a combining rule for the program of Example \ref{example: ground program}. Then, to compute the probability density given to the assignment in the example, we will use the following conditional probability densities/masses:
\begin{table}[H]
\centering
 \begin{tabular}{|c|c|} 
 \hline
 $\mathtt{client(ann)\cong t}$ & $1.00$ \\
 $\mathtt{loan(l\_1)\cong t}$ & $1.00$ \\
 $\mathtt{loan(l\_2)\cong t}$ & $1.00$ \\
 $\mathtt{has\_loan(ann,l\_1)\cong t}$ & $0.20$ \\
 $\mathtt{has\_loan(ann,l\_2)\cong t}$ & $0.20$ \\
 $\mathtt{status(l\_1)\cong a}$ & $0.30$ \\
 $\mathtt{status(l\_2)\cong d}$ & $0.70$ \\
 $\mathtt{credit\_score(ann)\cong 601.2}$ & $0.04$ \\
 \hline
\end{tabular}
\label{table: STp example2}
\end{table}

\noindent
where, the last entry is the probability density at $\mathtt{601.2}$ according to the mixture of Gaussians $\{\mathcal{N}(x \mid 700, 10.9) + \mathcal{N}(x \mid 600, 20.5)\}/2$. Thus, the density assigned to the assignment is $0.20 \times 0.20 \times 0.30 \times 0.70 \times 0.04$.
\end{example}

We can show that these probability (or density) assignments define a unique probability distribution.
\begin{proposition}
Let $\mathbb{P}$ be a well-defined \dcplus program, and $\mathbf{V}_{\mathbb{P}}$ be a set of all RVs defined by it. Then $\mathbb{P}$ specifies a unique probability distribution over $\mathbf{V}_{\mathbb{P}}$.
\end{proposition}
Since well-defined \dcplus programs satisfy all conditions of well-defined BLPs, this proposition follows from Proposition 1 of \cite{kersting2001adaptive}

\section{Inference in Ground \dcplus Programs}\label{section: cs-lw}
Before presenting the sampling algorithm for first-order \dcplus programs, we will first design an efficient sampling algorithm for ground \dcplus programs describing BNs with structured CPDs. Thus, this section will focus only on programs having mutually exclusive clauses, which are sufficient to describe such BNs. The algorithm presented in this section aims to exploit structures within CPDs or the structure of clauses in ground programs. However, in this section, we will assume that continuous RVs are absent for simplicity. Once this algorithm is clear, the extended algorithm for first-order \dcplus programs presented in the next section will be easy to comprehend.

A natural representation of the structures in CPDs is via {\em tree-CPDs} \cite{koller2009probabilistic}, as illustrated in Figure \ref{fig:context-specific independence}. For all assignments to the parents of a variable $ A $, a unique leaf in the tree specifies a (conditional) distribution over $ A $. The path to each leaf dictates the contexts, i.e., {\em partially assigned parents}, given which this distribution is used. We can easily represent tree-CPDs using DCs, where each path from the root to a leaf in each tree-CPD maps to a rule. 

\begin{example}\label{example: distributional clauses}
\normalfont The set of clauses for the tree-CPD in Figure \ref{fig:context-specific independence}:

$\mathtt{e \sim bernoulli(0.2) \leftarrow a\cong1.}$

$\mathtt{e \sim bernoulli(0.9) \leftarrow a\cong0, b\cong1.}$

$\mathtt{e \sim bernoulli(0.6) \leftarrow a\cong0, b\cong0, c\cong1.}$

$\mathtt{e \sim bernoulli(0.3) \leftarrow a\cong0, b\cong0, c\cong0.}$
\end{example}

Conversely, we can view a ground program with mutually exclusive clauses as a representation of tree-CPDs of all variables of a BN.
Thus, we can alternatively define the probability distribution specified by such ground programs as follows,

\begin{definition}
Let $\mathcal{B}$ be a Bayesian network with tree-CPDs specifying a distribution $P$. Let $\mathbb{P}$ be a set of distributional clauses such that each path from the root to a leaf of each tree-CPD corresponds to a clause in $\mathbb{P}$. Then $\mathbb{P}$ specifies the same distribution $P$.
\end{definition}

To avoid confusion, such ground programs will be called DC($\mathcal{B}$) programs.

While an efficient sampling algorithm for $\mathcal{B}$ only exploits the graph structure (CIs properties) in $\mathcal{B}$, the key to designing an efficient sampling algorithm for DC($\mathcal{B}$) programs is to exploit both the underlying graph and the clause structure (CSIs properties). To this end, we start with our discussion on the estimation of unconditional probability queries, which is necessary to support our further discussion on conditional probability queries, where we present the full algorithm for DC($\mathcal{B}$) programs. 



\subsection{Top-Down Proof Procedure for DC($\mathcal{B}$) programs}\label{Section: proof procedure}

Estimating unconditional probability queries in BNs is easy. We just need to generate some random samples of query variables and to find the fraction of times the query is true, which is the estimated probability of the query. In this sampling process, all ancestors of query variables are also sampled. However, due to the clause structure in DC($\mathcal{B}$) programs, it is possible to generate random samples of query variables by sampling only some (not all) ancestors, which makes the sampling process more efficient. A simplified version of an approach due to \cite{nitti2016probabilistic} is discussed in Algorithm \ref{algorithm: proof0}. This approach resembles SLD resolution \cite{kowalski1974predicate} for definite programs. However, there are some differences due to the stochastic nature of sampling. Unlike SLD resolution, this approach maintains global variable $\mathtt{Asg}$ to record sampled values of RVs.


\begin{algorithm}[t]
\caption{\dc Proof Procedure}
\label{algorithm: proof0}
\begin{algorithmic}
\Procedure{prove-ground}{$\mathtt{G}$}
\begin{itemize}
    \item Proves a conjunction of ground atoms $\mathtt{G}$.
    \item Returns $\mathtt{yes}$ if there is a choice that makes $\mathtt{G}$ empty; otherwise the procedure fails.
\end{itemize}
\begin{enumerate}
    \item While $\mathtt{G}$ is not empty:
    \begin{enumerate}
        \item Select the first atom $\mathtt{A}$ from $\mathtt{G}$.
        \item If $\mathtt{A}$ is of the form $\mathtt{t\cong v}$:
        \begin{enumerate}
            \item If a value of $\mathtt{t}$ is recorded in $\mathtt{Asg}$:
            \begin{enumerate}
                \item If $\mathtt{Asg[t]==v}$: remove $\mathtt{A}$ from $\mathtt{G}$. \label{condition1: equality0}
            \end{enumerate}
            \item Else:
            \begin{enumerate}
                \item \textbf{Choose} $\mathtt{t \sim D \leftarrow B_1, \dots, B_n} \in \mathbb{P}$ with $\mathtt{t}$ in the head.
                \item Set $\mathtt{G :=  B_1, \dots, B_n, sample(t, D), G}$.
            \end{enumerate}
        \end{enumerate}
            \item Else If $\mathtt{A}$ is of the form $\mathtt{sample(t, D)}$:
            \begin{enumerate}
                \item Sample a value $\mathtt{v}$ from $\mathtt{D}$, record $\mathtt{Asg[t] := v}$, and remove $\mathtt{A}$ from $\mathtt{G}$. 
            \end{enumerate}
    \end{enumerate}
    \item Return $\mathtt{yes}$.
\end{enumerate}
\EndProcedure
\end{algorithmic}
\end{algorithm}

Given an initial goal $\mathtt{G_0}$ and global variable $\mathtt{Asg}$, the algorithm recursively produces new goals $\mathtt{G_1}$,$\mathtt{G_1}$, etc., and updates $\mathtt{Asg}$. There are two cases when it is not possible to obtain $\mathtt{G_{i+1}}$ from $\mathtt{G_i}$: 
\begin{itemize}
    \item the first is when the selected subgoal of the form $\mathtt{t \cong v}$ cannot be resolved because the sampled value of the RV term already recorded in $\mathtt{Asg}$ is different from $\mathtt{v}$. 
    \item the other case appears when $\mathtt{G_i = \square}$ (i.e. the empty goal). 
\end{itemize}
The procedure\footnotemark \ results in a {\em derivation} of $\mathtt{G_0}$, a finite sequence of goals starting with the initial goal
\footnotetext{Since cyclic dependency among RVs is not allowed in well-defined programs, and the number of clauses in the program is finite, the procedure is guaranteed to terminate.} 
\begin{example}\label{example: proof process}
Consider the initial goal ($\mathtt{G_0}$) $\mathtt{\leftarrow e\cong 1}$ 
and the following DC($\mathcal{B}$) program: 

$\mathtt{a \sim bernoulli(0.1).}$

$\mathtt{d \sim bernoulli(0.3).}$

$\mathtt{b \sim bernoulli(0.2) \leftarrow a\cong0.}$

$\mathtt{b \sim bernoulli(0.6) \leftarrow a\cong1.}$

$\mathtt{c \sim bernoulli(0.2) \leftarrow a\cong1.}$

$\mathtt{c \sim bernoulli(0.7) \leftarrow a\cong0, b\cong1.}$

$\mathtt{c \sim bernoulli(0.8) \leftarrow a\cong0, b\cong0.}$

$\mathtt{e \sim bernoulli(0.9) \leftarrow c\cong1.}$

$\mathtt{e \sim bernoulli(0.4) \leftarrow c\cong0, d\cong1.}$

$\mathtt{e \sim bernoulli(0.3) \leftarrow c\cong0, d\cong0.}$

\noindent
Use $\mathtt{sample(k, l)}$ as a shorthand notation for $\mathtt{sample(k, bernoulli(l))}$. A derivation of $\mathtt{G_0}$, the state of $\mathtt{Asg}$ and the program clause used in each step is shown in Figure \ref{fig: derivation}. 
\begin{figure}[t]
    \centering
    \includegraphics[width=0.9\linewidth]{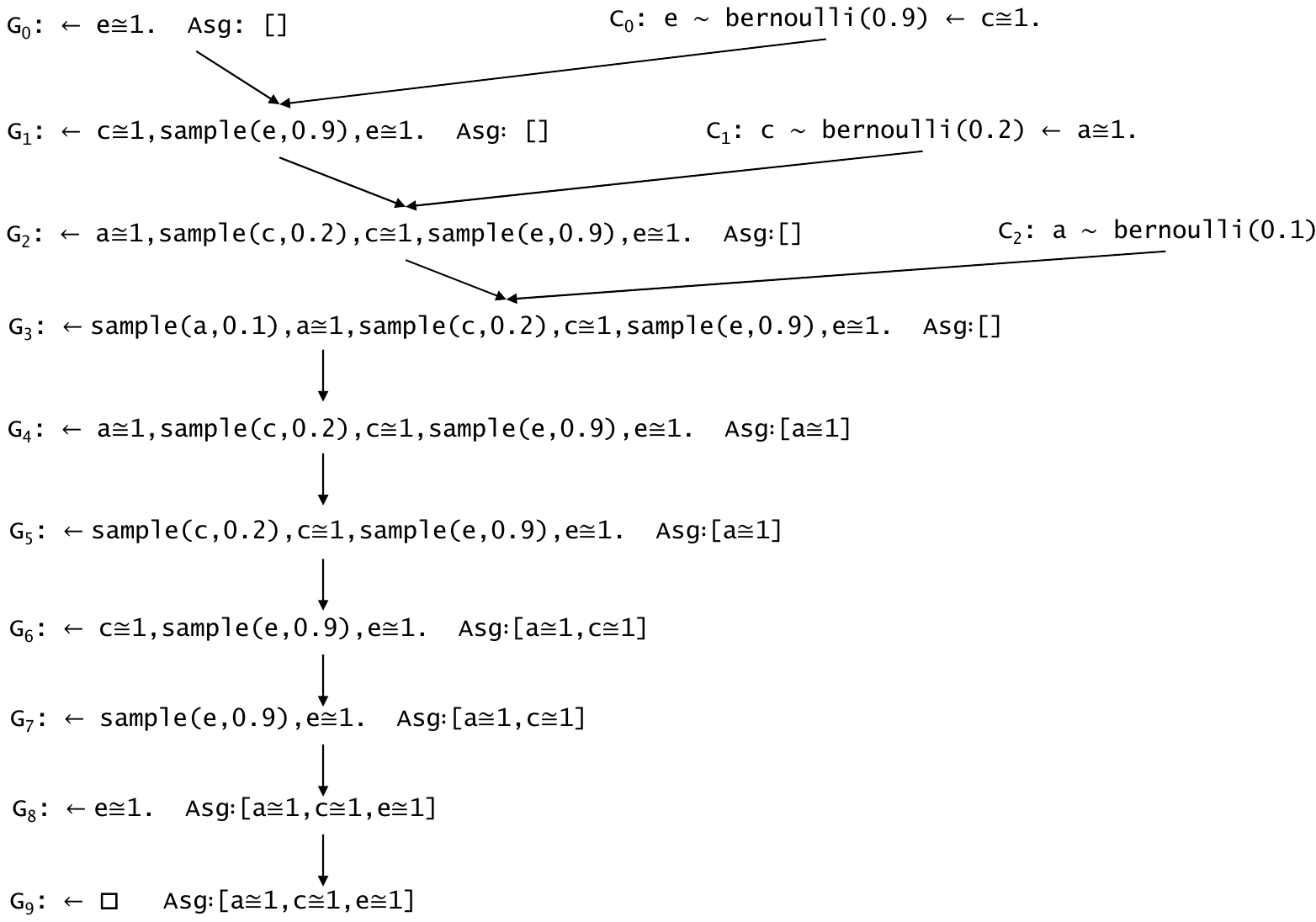}
    \caption{A search graph for a top-down derivation}
    \label{fig: derivation}
\end{figure}













\end{example}

As usual, the derivation of goal $\mathtt{G_0}$ that ends in the empty goal corresponds to a {\em refutation} of the goal. Not all derivations lead to refutations. As already pointed out, if the selected subgoal cannot be resolved, the derivation {\em fails}. A refutation or a failed derivation is a {\em complete derivation}. If the selected subgoal of some goal can be resolved with more than one program clause, there can be many complete derivations. The procedure searches a refutation by generating multiple complete derivations in a depth-first search fashion. Notice that, unlike SLD-resolution, here derivations might depend on previous complete derivations. This is because the previous derivations might have updated the $\mathtt{Asg}$ table, which in turn might influence the current derivation. Additionally, it should be clear that no or only one derivation can be the refutation when clauses in a program are mutually exclusive. So, for efficiency, one should stop generating more derivations in such programs once a refutation is obtained.

To estimate the probability of $\mathtt{e \cong 1}$, we call PROVE-GROUND($\mathtt{e \cong 1}$) repeatedly. The fraction of times we get refutation (i.e., the algorithm returns $\mathtt{yes}$) is the estimated probability.
Notice that some ancestors of the query variables may not be sampled on some occasions, e.g., in Example \ref{example: proof process}, variables $\mathtt{d}$ and $\mathtt{b}$  were not sampled.
Thus, we speed up the sampling process and, in this way, exploit the structure of clauses while estimating marginal probabilities. 


\subsection{Exploiting Context-Specific Independencies}
We now turn our attention to the problem of estimating conditional probabilities, which is more interesting but is significantly more complicated. Exploiting CSIs in this problem is unexplored in
in \cite{nitti2016probabilistic}.
Here we introduce a sampling algorithm that combines the proof procedure, discussed in the previous section, with the Bayes-ball simulation, discussed in Section \ref{section: bayes-ball}. The combined algorithm then exploits both the clause structures and the underlying graph structure of \dc programs. As we will see, it samples variables given the states of only some of their requisite ancestors. This contrasts with the Bayes-ball simulation of BNs,
where knowledge of all such ancestors' states is required.
This section is divided into two parts. The first part presents a novel notion of contextual assignment that allows for exploiting CSIs. It provides insight into the computation of $\mu$ using partial assignments of requisite variables. We will show that CSIs allow for breaking the main problem of computing $\mu$ into several sub-problems that can be solved independently. 
The second part presents the sampling algorithm and justifies it using the notion introduced in the first part.

\subsubsection{Notion of Contextual Assignments}
Recall from Section \ref{section: bayes-ball} that variables $\mathbf{X}$, $\mathbf{Z}_\star$, $\mathbf{E}_\star$, $\mathbf{E}_\smwhitestar$ are requisite for computing the query $\mu$ and the requisite network $\mathcal{B}_\star$ is formed by these variables. We will consider partial assignments of these variables, which will be used to compute $\mu$.
Let us start by defining these assignments. 
\begin{definition} 
Let $\mathbf{Z}_\dagger \subseteq \mathbf{Z}_\star$ and $\mathbf{e}_\dagger \subseteq \mathbf{e}_\star$. Denote $\mathbf{Z}_\star \setminus \mathbf{Z}_\dagger$ by $\mathbf{Z}_\ddagger$, and $\mathbf{e}_\star \setminus \mathbf{e}_\dagger$ by $\mathbf{e}_\ddagger$.
A partial assignment $\mathbf{x}$, $\mathbf{z}_\dagger$, $\mathbf{Z}_\ddagger$,  $\mathbf{e}_\dagger$, $\mathbf{e}_\ddagger$ will be called contextual assignment if due to CSIs in $P$,
\begin{equation*}
    \prod_{u_i \in \mathbf{x} \cup \mathbf{z}_\dagger \cup \mathbf{e}_\dagger} P(u_i \mid \mathbf{pa}(U_i)) = \prod_{u_i \in \mathbf{x} \cup \mathbf{z}_\dagger \cup \mathbf{e}_\dagger} P(u_i \mid \mathbf{ppa}(U_i))
\end{equation*}
where $\mathbf{ppa}(U_i)$ is a set of partially assigned parents of $U_i$, that is, $\mathbf{Ppa}(U_i) = \mathbf{Pa}(U_i) \setminus \mathbf{Z}_\ddagger$.
\end{definition}
Here, we summarise some important RVs or their assignments:
\begin{table}[H]
\centering
 \begin{tabular}{|c|c|} 
 \hline
 $\mathbf{X}$ & Query variables\\ \hline
 $\mathbf{Z}_\star$ & Unobserved requisite variables apart from query variables\\ \hline
 $\mathbf{e}_\star$ & Diagnostic evidence\\ \hline
 $\mathbf{e}_\smwhitestar$ & Predictive evidence\\ \hline
 $\mathbf{z}_\dagger$ & Assigned subset of $\mathbf{Z}_\star$ in a contextual assignment\\ \hline
 $\mathbf{Z}_\ddagger$ & Subset of $\mathbf{Z}_\star$ not assigned in a contextual assignment\\ \hline
 $\mathbf{e}_\dagger$ & Subset of $\mathbf{e}_\star$ in a contextual assignment\\ \hline
 $\mathbf{e}_\ddagger$ & Subset of $\mathbf{e}_\star$ not in a contextual assignment\\ \hline
\end{tabular}
\end{table}

\begin{example}\label{example: contexual assignment}
Consider the network of Figure \ref{fig:context-specific independence}, and assume that our diagnostic evidence is $\{F=1, G=0, H=1\}$, predictive evidence is $\{D=1\}$, and query is $\{E=0\}$. From the CPD's structure, we have: $P(E=0 \mid A=1, B, C) = P(E=0 \mid A=1)$; consequently, a contextual assignment is $\mathbf{x} = \{E=0\}, \mathbf{z}_\dagger = \{A=1\}, \mathbf{e}_\dagger= \{\}, \mathbf{Z}_\ddagger = \{B,C\}, \mathbf{e}_\ddagger=\{F=1,G=0,H=1\}$. We also have: $P(E=0 \mid A=0, B=1, C) = P(E=0 \mid A=0, B=1)$; consequently, another such assignment is $\mathbf{x} = \{E=0\}, \mathbf{z}_\dagger = \{A=0,B=1\}, \mathbf{e}_\dagger= \{H=1\}, \mathbf{Z}_\ddagger = \{C\}, \mathbf{e}_\ddagger=\{F=1,G=0\}$.
\end{example}

We aim to treat the evidence $\mathbf{e}_\ddagger$ independently, thus, we define it first.
\begin{definition}\label{Definition: residual diagnostic evidence} 
The diagnostic evidence $\mathbf{e}_\ddagger$ in a contextual assignment  $\mathbf{x}$, $\mathbf{z}_\dagger$, $\mathbf{Z}_\ddagger$,  $\mathbf{e}_\dagger$, $\mathbf{e}_\ddagger$ will be called residual evidence.
\end{definition}

\begin{figure}[t]
    \centering
    \includegraphics[width=0.5\linewidth]{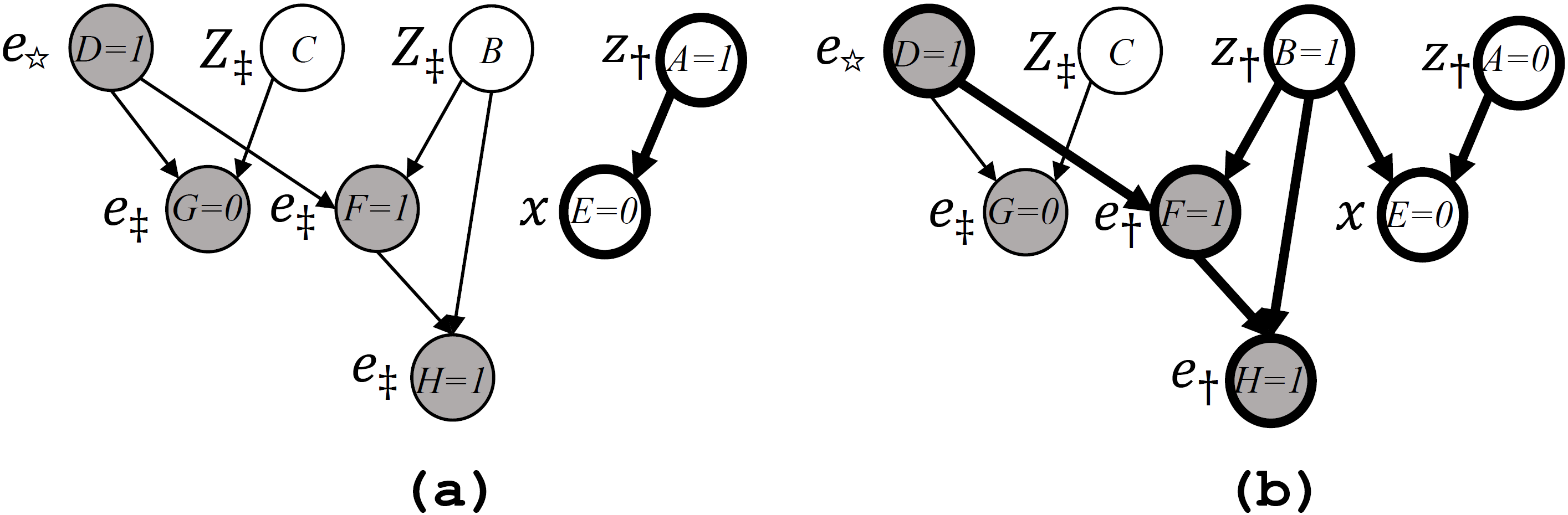}
    \caption{Two safe contextual assignments to RVs of BN of Figure \ref{fig:context-specific independence}: (a) in the context ${A=1}$, where edges $C \rightarrow E$ and $B \rightarrow E$ are redundant since $E \perp B, C \mid A=1$; (b) in the context ${A=0, B=1}$, where the edge $C \rightarrow E$ is redundant since $E \perp C \mid A=0, B=1$. To identify such assignments, intuitively, we should apply the Bayes-ball algorithm after removing these edges. Portions of graphs that the algorithm visits, starting with visiting the variable $E$ from its child, are highlighted. Notice that variables $\mathbf{X}, \mathbf{Z}_\dagger, \mathbf{E}_\dagger$ lie in the highlighted portion.}
    \label{fig: sub-graphs}
\end{figure}

However, contextual assignments do not immediately allow us to treat the residual evidence independently.  We need the assignments to be safe. 
\begin{definition}\label{definition: Basis} 
Let $e \in \mathbf{e}_\star$ be a diagnostic evidence, and let $S$ be an unobserved ancestor of $E$ in the graph structure in $\mathcal{B}_\star$, where $\mathcal{B}_\star$ is the sub-network formed by the requisite variables. Let $S \rightarrow \cdots\ B_i\ \cdots \rightarrow E$ be a causal trail such that either no $B_i$ is observed or there is no $B_i$. Let $\mathbf{S}$ be the set of all such $S$. Then the variables $\mathbf{S}$ will be called basis of $e$. Let $\mathbf{\dot{e}}_\star \subseteq \mathbf{e}_\star$, and let $\mathbf{\dot{S}}_{\star}$ be the set of all such $S$ for all $e \in \mathbf{\dot{e}}_\star$. Then $\mathbf{\dot{S}}_{\star}$ will be called basis of $\mathbf{\dot{e}}_\star$.
\end{definition}
Reconsider Example \ref{example: contexual assignment}; the basis of $\{F=1\}$ is $\{B\}$. 
\begin{definition}\label{Definition: safe contexual assignments} 
Let $\mathbf{x}$, $\mathbf{z}_\dagger$, $\mathbf{Z}_\ddagger$,  $\mathbf{e}_\dagger$, $\mathbf{e}_\ddagger$ be a contextual assignment, and let $\mathbf{S}_{\ddagger}$ be the basis of the residual evidence $\mathbf{e}_\ddagger$. If $\mathbf{S}_\ddagger \subseteq \mathbf{Z}_\ddagger$ then the contextual assignment will be called safe.
\end{definition}
\begin{example}\label{example: safe contexual assignment}
Reconsider Example \ref{example: contexual assignment}; the first example of a contextual assignment is safe, but the second is not since the basis $B$ of $\mathbf{e}_\ddagger$ has a non-empty intersection with $\mathbf{Z}_\dagger$. We can make the second safe like this: $\mathbf{x} = \{E=0\}, \mathbf{z}_\dagger = \{A=0,B=1\}, \mathbf{e}_\dagger= \{F=1,H=1\}, \mathbf{Z}_\ddagger = \{C\}, \mathbf{e}_\ddagger=\{G=0\}$. See Figure \ref{fig: sub-graphs}.
\end{example}
Before showing that the residual evidence can now be treated independently, we first define a random variable called {\em weight}. 
\begin{definition} 
Let $e \in \mathbf{e}_\star$ be a diagnostic evidence, and let $W_e$ be a random variable defined as follows:
\begin{equation*}
    W_e = P(e \mid \mathbf{Pa}(E)).
\end{equation*}
The variable $W_e$ will be called weight of $e$. The weight of a subset $\mathbf{\dot{e}}_\star \subseteq \mathbf{e}_\star$ is defined as follows:
\begin{equation*}
    W_{\mathbf{\dot{e}}_\star} = \prod_{u_i \in \mathbf{\dot{e}}_\star} P(u_i \mid \mathbf{Pa}(U_i)).
\end{equation*}
\end{definition}

Now we can show the following result:
\begin{theorem}\label{Theorem: the expecation}
Let $\mathbf{\dot{e}}_{\star} \subseteq \mathbf{e}_\star$, and let $\mathbf{\dot{S}}_{{\star}}$ be the basis of $\mathbf{\dot{e}}_{\star}$. Then the expectation of weight $W_{\mathbf{\dot{e}}_{\star}}$ relative to the distribution $Q_\star$ as defined in Equation \ref{equation: proposal distribution} can be written as:
\begin{equation*}
\begin{aligned}
    \mathbb{E}_{Q_\star}[W_{\mathbf{\dot{e}}_{\star}}] = \sum_{\mathbf{\dot{s}}_{\star}} \prod_{u_i \in \mathbf{\dot{e}}_{\star} \cup \mathbf{\dot{s}}_{\star}} P(u_i \mid \mathbf{pa}(U_i)).
\end{aligned}
\end{equation*}
\end{theorem}
Hence, apart from unobserved variables $\mathbf{\dot{S}}_{{\star}}$, the computation of $\mathbb{E}_{Q_\star}[W_{\mathbf{\dot{e}}_{\star}}]$ does not depend on other unobserved variables.

Let $\psi$ denotes a contextual assignment $\mathbf{x}$, $\mathbf{z}_\dagger$, $\mathbf{Z}_\ddagger$,  $\mathbf{e}_\dagger$, $\mathbf{e}_\ddagger$. The range of $\psi$, denoted $range(\psi)$, is the set of all full assignments constructed by assigning the unobserved variables in $\mathbf{Z}_\ddagger$. Reconsider the contextual assignment of Example \ref{example: safe contexual assignment}. Since $C$ is a Boolean random variable, the range of the contextual assignment is $\{\{E=0, A=0, B=1, F=1, H=1, C=0, G=0\}, \{E=0, A=0, B=1, F=1, H=1, C=1, G=0\}\}$.

It is worth noting that a full assignment $\psi$ is a safe contextual assignment, where the set of unobserved variables and the residual evidence set are empty. In such a case, $range(\psi) = \{\psi\}$.

The next theorem requires contextual assignments to be mutually exclusive. Two contextual assignments $\psi$ and $\psi^{\prime}$ are mutually exclusive if $range(\psi) \cap range(\psi^{\prime}) = \emptyset$.


\begin{theorem}\label{theorem: cslw_main}
Let $\Psi$ be a set of mutually exclusive contextual assignments such that each full assignment $\mathbf{x}$, $\mathbf{z}_\star$, $\mathbf{e}_\star$, $\mathbf{e}_\smwhitestar$ of the requisite network $\mathcal{B}_\star$ is under the range of a safe contextual assignment $\psi \in \Psi$. Let $\mathbf{x}[\psi]$, $\mathbf{z}_\dagger[\psi]$, $\mathbf{Z}_\ddagger[\psi]$,  $\mathbf{e}_\dagger[\psi]$, $\mathbf{e}_\ddagger[\psi]$ denote assigned variables, unobserved variables and evidence in $\psi \in \Psi$. Then the query $\mu$ to $P$ can be computed as follows:
\begin{equation}
    \begin{aligned}
    \frac{\sum\limits_{\psi \in \Psi} \bigl(\prod\limits_{u_i \in \mathbf{x}[\psi] \cup \mathbf{z}_\dagger[\psi] \cup \mathbf{e}_\dagger[\psi]} P(u_i \mid \mathbf{ppa}(U_i)) f(\mathbf{x}[\psi]) R[\psi]\bigr)}{\sum\limits_{\psi \in \Psi} \bigl(\prod\limits_{u_i \in \mathbf{x}[\psi] \cup \mathbf{z}_\dagger[\psi] \cup \mathbf{e}_\dagger[\psi]} P(u_i \mid \mathbf{ppa}(U_i)) R[\psi])\bigr)}
    \end{aligned}
\end{equation}
where $R[\psi]$ denotes $\mathbb{E}_{Q_\star}[W_{\mathbf{e}_\ddagger[\psi]}]$.
\end{theorem}

We draw some important conclusions: i) $\mu$ can be exactly computed by performing the summation over safe contextual assignments; notably, variables in $\mathbf{Z}_\dagger$ vary, and so do variables in $\mathbf{E}_\dagger$; ii) For all $\psi \in \Psi$, the computation of $\mathbb{E}_{Q_\star}[W_{\mathbf{e}_\ddagger[\psi]}]$ does not depend on the context $\mathbf{x[\psi]}, \mathbf{z}_\dagger[\psi]$ since no basis of $\mathbf{e}_\ddagger[\psi]$ is assigned in the context (by Theorem \ref{Theorem: the expecation}). Hence, $\mathbb{E}_{Q_\star}[W_{\mathbf{e}_\ddagger[\psi]}]$ can be computed independently. However, the context decides which evidence should be in the subset $\mathbf{e}_\ddagger[\psi]$. That is why we can not cancel $\mathbb{E}_{Q_\star}[W_{\mathbf{e}_\ddagger[\psi]}]$ from the numerator and denominator.

\subsubsection{Context-Specific Likelihood Weighting}
First, we present an algorithm that simulates a \dc program $\mathbb{P}$ and generates safe contextual assignments. Then we discuss how to estimate the expectations independently before estimating $\mu$.

\paragraph{\textbf{Simulation of \dc Programs}}
We start by asking a question. Suppose we modify the first and the fourth rule of Bayes-ball simulation, discussed in Section \ref{section: bayes-ball}, as follows:
\begin{itemize}
    \item In the first rule, when the visit of an unobserved variable is from its child, everything remains the same except that only \underline{some parents} are visited, not all.
    \item Similarly, in the fourth rule, when the visit of an observed variable is from its parent, everything remains the same except that only \underline{some parents} are visited.
\end{itemize}
Which variables will be assigned, and which will be weighted using the modified simulation rules? Intuitively, only a subset of variables in $\mathbf{Z}_\star$ should be assigned, and only a subset of variables in $\mathbf{E}_\star$ should be weighted. But then how to assign/weigh a variable knowing the state of only some of its parent. We can do that when structures are present within CPDs of $\mathcal{B}$, and these structures are explicitly represented using clauses in a \dc program $\mathbb{P}$.  Recall that the proof procedure, discussed in Section \ref{Section: proof procedure}, can sample random variables without sampling some of their ancestors. Hence, the key idea is to visit only some parents (if possible due to structures); consequently, those unobserved parents that are not visited might not be needed to be sampled. 

To realize that, we need to adapt the Bayes-ball simulation such that it works on \dc programs. However, there is a problem: $\mathbb{P}$ is a set of clauses, and no explicit graph is associated with it on which the Bayes-ball can be applied. Fortunately, we can infer direct influence relationships using the dependency set of $\mathbb{P}$, which is automatically constructed, as discussed in Section \ref{Section: sematics}. 
The adapted simulation for \dc programs is defined procedurally in Algorithm \ref{algorithm: forward-backward}. The algorithm visits variables from their parents and calls the top-down proof procedure (Algorithm \ref{algorithm: proof}) to visit variables from their children. Like Bayes-ball, these algorithms also mark variables on top and bottom to avoid repeating the same action. 

\begin{algorithm}[t]
\caption{Simulation of \dc Programs}
\label{algorithm: forward-backward}
\begin{algorithmic}
\Procedure{Simulate-ground-DC}{$\mathbf{x}, \mathbf{e}$}
\begin{itemize}
    \item Simulates a \dc program $\mathbb{P}$ based on inputs: i) $\mathbf{x}$: a query; ii) $\mathbf{e}$: evidence. 
    \item Let $dep(\mathbb{P})$ be the dependency set of $\mathbb{P}$. 
    \item The procedure maintains global data structures: i) $\mathtt{Asg}$, a table that records assignments of variables ($\mathbf{x} \cup \mathbf{z}_\dagger$);  ii) $\mathtt{Forward}$, a set of variables whose children to be visited from parent; iii) $\mathtt{Top}$, a set of variables marked on top; iv) $\mathtt{Bottom}$, a set of variables marked on bottom.
    \item Output: i) $f(\mathbf{x})$ that can be either $0$ or $1$; ii) $\mathtt{W}$: a table of weights of diagnostic evidence ($\mathbf{e}_\dagger$).
\end{itemize}
\begin{enumerate}
    \item Empty $\mathtt{Asg, W, Top, Bottom, Forward}$.
    \item If \Call{prove-marked-ground}{$\mathbf{x}$} is $\mathtt{yes}$ then  $f(\mathbf{x}):=1$ else $f(\mathbf{x}):=0$.
    \item While $\mathtt{Forward}$ is not empty:
    \begin{enumerate}
        \item Remove $\mathtt{A}$ from $\mathtt{Forward}$ and add $\mathtt{A}$ to $\mathtt{Bottom}$.
        \item For all $\mathtt{A_0}$ such that $dep(\mathbb{P}) \models \mathtt{pa(A_0,A)}$:
        \begin{enumerate}
            \item If $\mathtt{A_0}$ is observed with value $\mathtt{v}$ in $\mathbf{e}$ and $\mathtt{A_0 \notin Top}$:
            \begin{enumerate}
                \item \textbf{Choose} $\mathtt{A_0 \sim D \leftarrow A_1, \dots, A_n} \in \mathbb{P}$ such that \Call{prove-marked-ground}{$\mathtt{A_1, \dots, A_n}$} is $\mathtt{yes}$. Add $\mathtt{A_0}$ to $\mathtt{Top}$.
                \item Let $\mathtt{p}$ be the likelihood at $\mathtt{v}$ according to distribution $\mathtt{D}$. Record $\mathtt{W[A_0]:=p}$.
            \end{enumerate}
            \item If $\mathtt{A_0}$ is not observed in $\mathbf{e}$ and $\mathtt{A_0 \notin Bottom}$: add  $\mathtt{A_0}$ to  $\mathtt{Forward}$. 
        \end{enumerate}
    \end{enumerate}
    \item Return $[f(\mathbf{x}), \mathtt{W}]$.
\end{enumerate}
\EndProcedure
\end{algorithmic}
\end{algorithm}

\begin{algorithm}[t]
\caption{\dc Proof Procedure Marked}
\label{algorithm: proof}
\begin{algorithmic}
\Procedure{prove-marked-ground}{$\mathtt{G}$}
\begin{itemize}
    \item Proves a conjunction of ground atoms $\mathtt{G}$, consequently, visits variables from child.
    \item Accesses $\mathbb{P}$, $\mathtt{Top, Bottom, Forward, Asg}$, and $\mathbf{e}$ as defined in Algorithm \ref{algorithm: forward-backward}. 
    \item Returns $\mathtt{yes}$ if there is a choice that makes $\mathtt{G}$ empty; otherwise the procedure fails.
\end{itemize}
\begin{enumerate}
    \item While $\mathtt{G}$ is not empty:
    \begin{enumerate}
        \item Select the first atom $\mathtt{A}$ from $\mathtt{G}$.
        \item If $\mathtt{A}$ is of the form $\mathtt{t\cong v}$:
        \begin{enumerate}
            \item If $\mathtt{t}$ is observed in $\mathbf{e}$ and its observed value is $\mathtt{v}$: remove $\mathtt{A}$ from $\mathtt{G}$.
            \item Else if $\mathtt{t\in Top }$: 
            \begin{enumerate}
                \item If $\mathtt{Asg[t]==v}$: remove $\mathtt{A}$ from $\mathtt{G}$. \label{condition1: equality}
            \end{enumerate}
            \item Else:
            \begin{enumerate}
                \item Add $\mathtt{t}$ to $\mathtt{Top}$.
                \item \textbf{Choose} $\mathtt{t \sim D \leftarrow B_1, \dots, B_n} \in \mathbb{P}$ with $\mathtt{t}$ in the head. 
                \item Set $\mathtt{G :=  B_1, \dots, B_n, sample(t, D), G}$.
            \end{enumerate}
        \end{enumerate}
            \item Else If $\mathtt{A}$ is of the form $\mathtt{sample(t, D)}$:
            \begin{enumerate}
                \item Sample a value $\mathtt{v}$ from $\mathtt{D}$, record $\mathtt{Asg[t] := v}$, and remove $\mathtt{A}$ from $\mathtt{G}$. 
                \item If $\mathtt{t \notin Bottom}$: add $\mathtt{t}$ to $\mathtt{Forward}$
            \end{enumerate}
    \end{enumerate}
    \item Return $\mathtt{yes}$.
\end{enumerate}
\EndProcedure
\end{algorithmic}
\end{algorithm}

\begin{algorithm}[t]
\caption{Generation of residual evidence's weights for \dc Programs}
\label{algorithm: generate residual weights ground}
\begin{algorithmic}
\Procedure{Weight-Res-Ground}{$\mathtt{Res}$}
\begin{itemize}
    \item Generates residual evidence's weights based on input: i) $\mathtt{Res}$, a list of residual evidence. This procedure accesses $\mathbb{P}$ of Algorithm \ref{algorithm: forward-backward}.
    \item Output: i) $\mathtt{W_1}$, a table of residual evidence's weights.
\end{itemize}
\begin{enumerate}
    \item For all random variables $\mathtt{A}_0$ in $\mathtt{Res}$.
    \begin{enumerate}
        \item Let $\mathtt{v}$ be the value observed for $\mathtt{A}_0$ in $\mathtt{Res}$.
        \begin{enumerate}
            \item \textbf{Choose} $\mathtt{A_0 \sim D \leftarrow A_1, \dots, A_n} \in \mathbb{P}$ such that \Call{prove-marked-ground}{$\mathtt{A_1, \dots, A_n}$} is $\mathtt{yes}$.
            \item Let $\mathtt{p}$ be the likelihood of $\mathtt{v}$ according to distribution $\mathtt{D}$. Record $\mathtt{W_1[A_0]:=p}$.
        \end{enumerate}
    \end{enumerate}
    \item Return $\mathtt{W_1}$.
\end{enumerate}
\EndProcedure
\end{algorithmic}
\end{algorithm}

Since the simulation of $\mathbb{P}$ follows the same four rules of Bayes-ball simulation except that only some parents are visited in the first and fourth rule, we show that 
\begin{lemma}\label{theorem: partial assignments}
Let $\mathbf{E_{\dagger}}$ be a set of observed variables weighed and let $\mathbf{Z_{\dagger}}$ be a set of unobserved variables, apart from query variables, assigned in a simulation of $\mathbb{P}$, then,
\begin{equation*}
    \mathbf{Z_{\dagger}} \subseteq \mathbf{Z_{\star}} \text{ and } \mathbf{E_{\dagger}} \subseteq \mathbf{E_{\star}}.
\end{equation*}
\end{lemma}
The query variables $\mathbf{X}$ are always assigned since the simulation starts with visiting these variables as if visits are from one of their children. To simplify notation, from now on we use $\mathbf{Z}_\dagger$ to denote the subset of variables in $\mathbf{Z}_\star$ that are assigned, $\mathbf{E}_\dagger$ to denote the subset of variables in $\mathbf{E}_\star$ that are weighted in the simulation of $\mathbb{P}$. $\mathbf{Z}_\ddagger$ to denote $\mathbf{Z}_\star \setminus \mathbf{Z}_\dagger$,
and $\mathbf{E}_\ddagger$ to denote $\mathbf{E}_\star \setminus \mathbf{E}_\dagger$
We show that the simulation performs safe contextual assignments to requisite variables. 
\begin{theorem}\label{theorem: dc-partial-justification}
Partial assignments $\mathbf{x}$, $\mathbf{z}_\dagger$, $\mathbf{Z}_\ddagger$,  $\mathbf{e}_\dagger$, $\mathbf{e}_\ddagger$ generated in simulations of \dc programs are safe contextual assignments. 
\end{theorem}
The proof of Theorem \ref{theorem: dc-partial-justification} relies on the following Lemma. 

\begin{lemma}\label{theorem: DC CSI}
Let $\mathbb{P}$ be a \dc program specifying a distribution $P$. Let $\mathbf{B}, \mathbf{C}$ be disjoint sets of parents of a variable $A$. In a simulation of $\mathbb{P}$, if $A$ is sampled/weighted, given an assignment $\mathbf{c}$, and without assigning $\mathbf{B}$, then,
\begin{equation*}
P(A \mid \mathbf{c}, \mathbf{B}) = P(A \mid \mathbf{c}).
\end{equation*}
\end{lemma}


Furthermore, two contextual assignments generated in two simulations of a \dc program are either identical or mutually exclusive. This is because the algorithm behind the two simulations is the same, so both will generate identical contextual assignments unless a different value is sampled for at least one unobserved variable. However, the two contextual assignments will clearly become mutually exclusive when a different value is sampled for an unobserved variable.


Hence, just like the standard LW, we sample from a factor $Q_\dagger$ of the proposal distribution $Q_\star$, which is given by, 
\begin{equation*}
    \begin{aligned}
    Q_\dagger = \prod_{u_i \in \mathbf{x} \cup \mathbf{z}_\dagger \cup \mathbf{e}_\dagger} P(u_i \mid \mathbf{ppa}(U_i))
    \end{aligned}
\end{equation*}
where $P(u_i \mid \mathbf{ppa}(U_i)) = 1$ if $u_i \in \mathbf{e}_\dagger$.
It is precisely this factor that Algorithm \ref{algorithm: forward-backward} considers for the simulation of $\mathbb{P}$. Starting by first setting $\mathbf{E}_\smwhitestar$, $\mathbf{E}_\ddagger$ their observed values, it assigns $\mathbf{X} \cup \mathbf{Z}_\dagger$ and weighs $\mathbf{e}_\dagger$ in the topological ordering. In this process, 
it records {\em partial weights} $\mathbf{w}_{\mathbf{e}_\dagger}$, such that: $\prod_{x_i \in \mathbf{e}_\dagger} w_{x_i} = w_{\mathbf{e}_\dagger}$ and $w_{x_i} \in \mathbf{w}_{\mathbf{e}_\dagger}$. Given $M$ partially weighted samples $\mathcal{D}_{\dagger} = \langle \mathbf{x}[1], \mathbf{w}_{\mathbf{e}_{\dagger}[1]}\rangle, \dots, \langle\mathbf{x}[M],  \mathbf{w}_{\mathbf{e}_{\dagger}[M]} \rangle$ from $Q_\dagger$, we could estimate $\mu$ using Theorem \ref{theorem: cslw_main} as follows:
\begin{equation}\label{equation: forward-backward sampling}
    \begin{aligned}
    \overline{\mu} = \frac{ \sum_{m=1}^{M} f(\mathbf{x}[m])\times w_{\mathbf{e_\dagger}[m]}\times\mathbb{E}_{Q_\star}[W_{\mathbf{e}_\ddagger[m]}] }{\sum_{m=1}^{M} w_{\mathbf{e_\dagger}[m]}\times \mathbb{E}_{Q_\star}[W_{\mathbf{e}_\ddagger[m]}]}
    \end{aligned}
\end{equation}

However, we still can not estimate it since we still do not have expectations $\mathbb{E}_{Q_\star}[W_{\mathbf{e}_\ddagger[m]}]$. 
Fortunately, there are ways to estimate them from partial weights in $\mathcal{D}_\dagger$. We discuss one such way next. 

\paragraph{\textbf{Estimating the Expected Weight of Residuals}}
We start with the notion of sample mean. Let $\mathcal{W}_\star = \langle w_{e_1}[1], \dots, w_{e_m}[1]\rangle, \dots, \langle w_{e_1}[n], \dots, w_{e_m}[n]\rangle$ be a data set of $n$ observations of weights of $m$ diagnostic evidence drawn using the standard LW. How can we estimate the expectation $\mathbb{E}_{Q_\star}[W_{e_i}]$ from $\mathcal{W}_\star$? The standard approach is to use the sample mean: $\overline{W}_{e_i} = \frac{1}{n}\sum_{r=1}^{n}w_{e_i}[r]$. In general, $\mathbb{E}_{Q_\star}[W_{e_i}\dots W_{e_j}]$ can be estimated using the estimator: $\overline{W_{e_i}\dots W}_{e_j} = \frac{1}{n}\sum_{r=1}^{n}w_{e_i}[r]\dots w_{e_j}[r]$. Since LW draws are independent and identical distributed (i.i.d.), it is easy to show that the estimator is unbiased. 

However, some entries, i.e., weights of residual evidence, are missing in the data set $\mathcal{W}_\dagger$ obtained using CS-LW. The trick is to fill the missing entries by drawing samples of the missing weights once we obtain $\mathcal{W}_\dagger$. More precisely, missing weights $\langle W_{e_i}, \dots, W_{e_j}\rangle$ in $r^{\text{th}}$ row of $\mathcal{W}_\dagger$ are filled in with a joint state $\langle w_{e_i}[r], \dots, w_{e_j}[r] \rangle$ of the weights. To draw the joint state, we use Algorithm \ref{algorithm: generate residual weights ground} and call \Call{weight-res-ground}{$[e_i, \dots, e_j]$} to visit observed variables $\langle E_i, \dots, E_j\rangle$ from parent. 
Once all missing entries are filled in, we can estimate $\mathbb{E}_{Q_\star}[W_{e_i}\dots W_{e_j}]$ using the estimator $\overline{W_{e_i}\dots W}_{e_j}$ as just discussed. Once we estimate all required expectations, it is straightforward to estimate $\mu$ using Equation \ref{equation: forward-backward sampling}. 

\paragraph{\textbf{Interpretation}}
At this point, we can gain some insight into the role of CSIs in sampling. They allow us to estimate the expectation $\mathbb{E}_{Q_\star}[W_{\mathbf{e}_\ddagger}]$ separately. We estimate it from all samples obtained at the end of the sampling process, thereby reducing the contribution $W_{\mathbf{e}_\ddagger}$ makes to the variance of our main estimator $\overline{\mu}$. The residual evidence $\mathbf{e}_\ddagger$ would be large if much CSIs are present in the distribution; consequently, we would obtain a much better estimate of $\mu$ using significantly fewer samples. Moreover, drawing a single sample would be faster since only a subset of requisite variables is visited. Hence, in addition to CIs, we exploit CSIs and improve LW further. We observe all these speculated improvements in our experiments. 

\section{Inference in First-Order \dcplus Programs}\label{section: fo-cs-lw}
\begin{algorithm}[t]
\caption{Simulation of \dcplus Programs}
\label{algorithm: forward-backward first-order}
\begin{algorithmic}
\Procedure{Simulate-\dcplus}{$\mathbf{x}, \mathbf{e}$}
\begin{itemize}
    \item Simulates a DC program $\mathbb{P}$ based on inputs: i) $\mathbf{x}$, a ground query; ii) $\mathbf{e}$: evidence. 
    \item Let $dep(\mathbb{P})$ be the dependency set of $\mathbb{P}$. 
    \item The procedure maintains global data structures: i) $\mathtt{Asg}$, a table that records partial assignments of unobserved variables;  ii) $\mathtt{Forward}$, a set of variables whose children to be visited from parent; iii) $\mathtt{Top}$, a set of variables marked on top; iv) $\mathtt{Bottom}$, a set of variables marked on bottom; v) $\mathtt{Dst}$, a table that records variables' distributions.
    \item Output: i) $f(\mathbf{x})$ that can be either $0$ or $1$; ii) $\mathtt{W}$: a table of weights of diagnostic evidence ($\mathbf{e}_\dagger$).
\end{itemize}
\begin{enumerate}
    \item Empty $\mathtt{Asg, W, Top, Bottom, Dst, Forward}$. Let $\mathcal{CR}$ be combining rules used for $\mathbb{P}$. 
    \item If \Call{prove-marked}{$\mathbf{x}$} fails then $f(\mathbf{x}):=0$ else $f(\mathbf{x}):=1$
    \item While $\mathtt{Forward}$ is not empty:
    \begin{enumerate}
        \item Remove $\mathtt{A}$ from $\mathtt{Forward}$ and add $\mathtt{A}$ to $\mathtt{Bottom}$
        \item For all $\mathtt{A_0}$ such that $dep(\mathbb{P}) \models \mathtt{pa(A_0,A)}$:
        \begin{enumerate}
            \item If $\mathtt{A_0}$ is observed to be $\mathtt{v}$ in $\mathbf{e}$ and $\mathtt{A_0 \notin Top}$:
            \begin{enumerate}
                \item Add $\mathtt{A_0}$ to $\mathtt{Top}$.
                \item For all (renamed) clause $\mathtt{B_0 \sim D \leftarrow B_1, \dots, B_m} \in \mathbb{P}$ such that mgu $\sigma$ unifies $\mathtt{B_0}$ and $\mathtt{A_0}$: call \Call{prove-marked}{$(\mathtt{B_1, \dots, B_m,add\_dst(B_0,D))\sigma}$}
                \item Let $[\mathtt{D_1,\dots, D_l}]$ be distributions for $\mathtt{A_0}$ recorded in table $\mathtt{Dst}$, and let $\mathtt{p}$ be the likelihood at $\mathtt{v}$ according to $\mathcal{CR}([\mathtt{D_1,\dots, D_l}])$.  Record $\mathtt{W[A_0]:=p}$.
            \end{enumerate}
            \item If $\mathtt{A_0}$ is not observed in $\mathbf{e}$ and $\mathtt{A_0 \notin Bottom}$: add  $\mathtt{A_0}$ to  $\mathtt{Forward}$ 
        \end{enumerate}
    \end{enumerate}
    \item Return $[f(\mathbf{x}), \mathtt{W}]$
\end{enumerate}
\EndProcedure
\end{algorithmic}
\end{algorithm}

This section extends CS-LW to first-order \dcplus programs where clauses need not be mutually exclusive. The extended algorithm is called {\em first-order context-specific likelihood weighting} (FO-CS-LW). 

Using tools of logic such as unification and substitution, it is easy to simulate first-order programs without completely grounding them first. The simulation process defined in Algorithm \ref{algorithm: forward-backward first-order} does that. It uses dependency sets of programs to visit RVs from parents and calls Algorithm \ref{algorithm: generate residual weights} to visit RVs from children. The RV sets of programs are used to identify RVs defined by programs. There are two important features of the top-down proof procedure with logical variables defined in Algorithm \ref{algorithm: generate residual weights}, which are worth highlighting. 

Firstly, it computes answer substitutions of a  query $\mathtt{\leftarrow A_1, \dots, A_m}$, that is, the substitutions of the refutations of the query restricted to the variables in the query. To realize that, instead of the query, the initial goal $\mathtt{G_0}$ is of the form: 
\begin{equation*}
    \mathtt{yes(V_1, \dots, V_k) \leftarrow A_1, \dots, A_m}
\end{equation*}
where $\mathtt{V_1, \dots, V_k}$ are the logical variables that appear in the query. This allows the procedure to return the answer $\mathtt{\{V_1/t_1, \dots, V_k/t_k\}}$ when the body of goal $\mathtt{G_i}$ is empty after applying the resolution rules. This is just like the proof procedure for definite programs with logical variables \cite{poole2010artificial}. 

Secondly and more importantly, the procedure searches and collects all distributions defined for an RV in a global variable $\mathtt{Dst}$ before sampling values of the RV from the combined distribution obtained using the combining rules. This is as per the semantics of \dcplus programs, and in this way, the algorithm also exploits ICIs. 

The generation of residual evidence's weights is defined in Algorithm \ref{algorithm: generate residual weights}.

\begin{algorithm}[t]
\small
\caption{\dcplus Proof Procedure}
\label{algorithm: proof first-order}
\begin{algorithmic}
\Procedure{prove-marked}{$\mathtt{Q}$}
\begin{itemize}
    \item Proves a conjunction of atoms $\mathtt{Q}$ with variables $\mathtt{V_1, \dots, V_k}$.
    \item Accesses $\mathbb{P}$, $\mathtt{Asg, Top, Bottom, Dst, Forward}$, $\mathbf{e}$, and $\mathcal{CR}$ as defined in Algorithm \ref{algorithm: forward-backward first-order}. 
    \item Returns substitutions of variables $\mathtt{V_1, \dots, V_k}$ if refutation exists; otherwise fails.
    \item Let $rv(\mathbb{P})$ be an RV set of $\mathbb{P}$.
\end{itemize}
\begin{enumerate}
    \item Set $\mathtt{G}$ to a clause $\mathtt{yes(V_1, \dots, V_k) \leftarrow Q}$
    \item While the body of $\mathtt{G}$ is not empty:
    \begin{enumerate}
        \item Suppose $\mathtt{G}$ is $\mathtt{yes(t_1, \dots, t_k) \leftarrow A_1, \dots, A_n}$. Select $\mathtt{A_1}$
        \item If $\mathtt{A_1}$ is an atom of the form $\mathtt{X\cong V}$:
        \begin{enumerate}
            \item \textbf{Choose} a grounding substitution $\sigma_1$ such that $rv(\mathbb{P}) \models \mathtt{X\sigma_1}$
            \item If $\mathtt{X\sigma_1}$ is observed in $\mathbf{e}$: let $\mathtt{v}$ be the observed value.
            \item Else if $\mathtt{X\sigma_1 \in Top }$: let $\mathtt{v}$ be $\mathtt{Asg[X\sigma_1]}$.
            \item Else:
            \begin{enumerate}
                \item For all (renamed) clause $\mathtt{B_0 \sim D \leftarrow B_1, \dots, B_m} \in \mathbb{P}$ such that mgu $\sigma_2$ unifies $\mathtt{B_0}$ and $\mathtt{X\sigma_1}$: call \Call{prove-marked}{$(\mathtt{B_1, \dots, B_m,add\_dst(B_0,D))\sigma_2}$}
                \item Let $\mathtt{M} = [\mathtt{D_1,\dots, D_l}]$ be distributions for $\mathtt{X\sigma_1}$ recorded in table $\mathtt{Dst}$. Let $\mathtt{v}$ be value sampled from $\mathcal{CR}(\mathtt{M})$. Record $\mathtt{Asg[X\sigma_1] := v}$.
                \item Add $\mathtt{X\sigma_1}$ to $\mathtt{Top}$. If $\mathtt{X\sigma_1 \notin Bottom}$ then add $\mathtt{X\sigma_1}$ to $\mathtt{Forward}$
            \end{enumerate}
            \item If mgu $\sigma_3$ unify $\mathtt{X\sigma_1 \cong v}$ and $\mathtt{X \cong V}$: set $\mathtt{G := ((yes(t_1, \dots, t_k) \leftarrow A_2, \dots, A_n)\sigma_1)\sigma_3}$
        \end{enumerate}
        \item Else if $\mathtt{A_1}$ is of the form $\mathtt{add\_dst(X, D)}$: // $\mathtt{A_1}$ will always be ground here
        \begin{enumerate}
            \item Record $\mathtt{Dst[X] := D}$ and set $\mathtt{G := yes(t_1, \dots, t_k) \leftarrow A_2, \dots, A_n}$
        \end{enumerate}
        \item\label{line: resolve comparison} Else if $\mathtt{A_1}$ is of the form $\mathtt{V_1 \diamond V_2}$ and evaluates to true: $\mathtt{G := yes(t_1, \dots, t_k) \leftarrow A_2, \dots, A_n}$
        
    \end{enumerate}
    \item Return $\mathtt{\{V_1/t_1, \dots, V_k/t_k\}}$ when $\mathtt{G}$ is $\mathtt{yes(t_1, \dots, t_k) \leftarrow}$
\end{enumerate}
\EndProcedure
\end{algorithmic}
\end{algorithm}

\begin{algorithm}[t]
\caption{Generation of residual evidence's weights for \dcplus Programs}
\label{algorithm: generate residual weights}
\begin{algorithmic}
\Procedure{Weight-Residuals}{$\mathtt{Res}$}
\begin{itemize}
    \item Generates residual evidence's weights based on input: i) $\mathtt{Res}$, a list of residual evidence. This procedure accesses $\mathbb{P}$ and $\mathcal{CR}$ of  Algorithm \ref{algorithm: forward-backward first-order}.
    \item Output: i) $\mathtt{W_1}$, a table of residual evidence's weights.
\end{itemize}
\begin{enumerate}
    \item For all random variable $\mathtt{A}_0$ in $\mathtt{Res}$.
    \begin{enumerate}
        \item Let $\mathtt{v}$ be the value observed for $\mathtt{A}_0$ in $\mathtt{Res}$.
        \begin{enumerate}
            \item For all (renamed) clause $\mathtt{B_0 \sim D \leftarrow B_1, \dots, B_m} \in \mathbb{P}$ such that mgu $\sigma$ unifies $\mathtt{B_0}$ and $\mathtt{A_0}$: call \Call{prove-marked}{$(\mathtt{B_1, \dots, B_m,add\_dst(B_0,D))\sigma}$}
            \item Let $[\mathtt{D_1,\dots, D_l}]$ be distributions for $\mathtt{A_0}$ recorded in table $\mathtt{Dst}$, and let $\mathtt{p}$ be the likelihood at $\mathtt{v}$ according to $\mathcal{CR}([\mathtt{D_1,\dots, D_l}])$.  Record $\mathtt{W_1[A_0]:=p}$.
        \end{enumerate}
    \end{enumerate}
    \item Return $\mathtt{W_1}$.
\end{enumerate}
\EndProcedure
\end{algorithmic}
\end{algorithm}



\paragraph{\textbf{Dealing with Continuous RVs}}
Till now, we have not talked about inference in programs with continuous RVs. Nevertheless, the notions discussed so far are sufficient to explain inference in such programs. Likelihood weighting naturally extends to continuous domains \cite{koller2009probabilistic}. We just need rules to resolve comparison atoms of the form $\mathtt{V_1 \diamond V_2}$ that appear in the body of clauses. These rules are already present in the proof procedure.

To ensure that FO-CS-LW correctly estimates the probabilities, we need to ensure that the simulation of \dcplus programs also generates safe contextual assignments that are identical or mutually exclusive. For this, we just need to ensure that Lemma \ref{theorem: DC CSI} also holds for the \dcplus programs' simulation.
\begin{theorem}\label{theorem: DC Plus CSI}
Lemma \ref{theorem: DC CSI} is also true for simulation of \dcplus programs.
\end{theorem}

\section{Empirical Evaluation}\label{section: experiments}
In this section, we empirically evaluate our inference algorithms and answer several research questions.
\subsection{How do the sampling speed and the accuracy of estimates obtained using CS-LW compare with the standard LW in the presence of CSIs?}

To answer it, we need BNs with structures present within CPDs. Such BNs, however, are not readily available since the structure while designing inference algorithms is generally overlooked. We identified two BNs from the Bayesian network repository \cite{bnrepository}, which have many structures within CPDs: i) \textit{Alarm}, a monitoring system for patients with 37 variables; ii) \textit{Andes}, an intelligent tutoring system with 223 variables.

\begin{table}[t]
\centering
\begin{adjustbox}{width=0.75\columnwidth,center}
\begin{tabular}{|c|c|cc|cc|}
\toprule
         &   &   \multicolumn{2}{|c|}{\textbf{LW}} &   \multicolumn{2}{|c|}{\textbf{CS-LW}} \\
\midrule
 \textbf{BN} & \textbf{N} & \textbf{MAE $\pm$ Std.} & \textbf{Time} & \textbf{MAE $\pm$ Std.} & \textbf{Time} \\
\midrule
\midrule
\multirow{4}{*}{\textit{Alarm}} & 100 & 0.2105 $\pm$ 0.1372	& 0.09 & 	\textbf{0.0721} $\pm$	\textbf{0.0983} & \textbf{0.06}\\ \cline{2-6}
 & 1000 & 0.0766 $\pm$ 0.0608 &	0.86 & \textbf{0.0240}	$\pm$ \textbf{0.0182}	& \textbf{0.53}
 \\
 \cline{2-6}
 & 10000 & 0.0282 $\pm$	0.0181 & 8.64 & \textbf{0.0091} $\pm$ \textbf{0.0069}	& \textbf{5.53} \\ \cline{2-6}
 & 100000 & 0.0086 $\pm$ 0.0067 & 89.93	& \textbf{0.0034} $\pm$ \textbf{0.0027} & \textbf{57.64}\\
\midrule
\midrule
\multirow{4}{*}{\textit{Andes}} & 100 & 0.0821 $\pm$ 0.0477 & 1.07 &	\textbf{0.0619} $\pm$ \textbf{0.0453} & \textbf{0.22} \\ \cline{2-6}
 & 1000 & 0.0257 $\pm$ 0.0184 & 10.62 & \textbf{0.0163} $\pm$ \textbf{0.0139} & \textbf{2.20} \\ \cline{2-6}
 & 10000 & 0.0087 $\pm$	0.0069 & 106.55 & \textbf{0.0058} $\pm$ \textbf{0.0042} &	\textbf{22.62}\\ \cline{2-6} 
 & 100000 & 0.0025 $\pm$ \textbf{0.0015} & 1074.93 & \textbf{0.0020} $\pm$ 0.0016 & \textbf{233.72}\\
\bottomrule
\end{tabular}
\end{adjustbox}
\caption{The mean absolute error (MAE), the standard deviation of the error (Std.), and the average execution time (in seconds) versus the number of samples (N). For each case, LW and CS-LW were executed 30 times.}
\label{Table: Result1}
\end{table}

We used the standard decision tree learning algorithm to detect structures and overfitted it on tabular-CPDs to get tree-CPDs, which was then converted into clauses. Let us denote the program with these clauses by $\mathbb{P}_{tree}$. CS-LW is implemented in the Prolog programming language, thus to compare the sampling speed of LW with CS-LW, we need a similar implementation of LW. Fortunately, we can use the same implementation of CS-LW for obtaining LW estimates. Recall that if we do not make structures explicit in clauses and represent each entry in tabular-CPDs with clauses, then CS-LW boils down to LW. Let $\mathbb{P}_{table}$ denotes the program where each rule in it corresponds to an entry in tabular-CPDs. Table \ref{Table: Result1} shows the comparison of estimates obtained using $\mathbb{P}_{tree}$ (CS-LW) and $\mathbb{P}_{table}$ (LW). Note that CS-LW automatically discards non-requisite variables for sampling. So, we chose the query and evidence such that almost all variables in BNs were requisite for the conditional query. 

As expected, we observe that less time is required by CS-LW to generate the same number of samples. This is because it visits only the subset of requisite variables in each simulation. \textit{Andes} has more structures compared to \textit{Alarm}. Thus, the sampling speed of CS-LW is much faster compared to LW in \textit{Andes}. Additionally, we observe that the estimate, with the same number of samples, obtained by CS-LW is much better than LW. This is significant. It is worth mentioning that approaches based on collapsed sampling obtain better estimates than LW with the same number of samples, but then the speed of drawing samples significantly decreases \cite{koller2009probabilistic}. In CS-LW, the speed increases when structures are present. This is possible because CS-LW exploits CSIs. 

Hence, we get the answer to our first question: When many structures are present, and when they are made explicit in clauses, then CS-LW will draw samples faster compared to LW. Additionally, estimates will be better with the same number of samples.

\subsection{How does FO-CS-LW perform as the domain size increases?}
\label{section: experiment 2}
The exact inference algorithms that PLP systems generally use for inference do not scale with domain sizes of logical variables in the program. Large domain sizes result in a huge ground program on which exact inference becomes intractable even for PLPs supporting only Boolean RVs. Thus, it is interesting to investigate how FO-CS-LW, an approximate inference algorithm, performs on such PLPs. 

For this purpose, we compared FO-CS-LW with the inference algorithm used in ProbLog, one of the most popular PLP systems. This algorithm first grounds first-order programs and then performs exact inference to exploit structures of clauses in the programs. We used a ProbLog program shown in Figure \ref{fig: experiment 2 example} for this experiment. 
Note that in ProbLog, when multiple distributions are specified for an RV, they are combined using NoisyOR, and when no distribution is specified, the RV is set to false. So, using this combining rule, the ProbLog program of Figure \ref{fig: experiment 2 example} can be expressed as a \dcplus program.
The domain size of each logical variable in the program is two since there are two clients, two accounts, and two loans. In such cases, instead of specifying the domain size of each logical variable separately, we will simply say that the domain size is two. Notice that relationships among clients, accounts, and loans are also probabilistic, so the number of RVs explicated by the program becomes huge as the domain size increases. More precisely, there are $3n^2 + 6n$ RVs when domain size is $n$. 

\begin{figure}[t]
    \centering
    \includegraphics[width=1.0\linewidth]{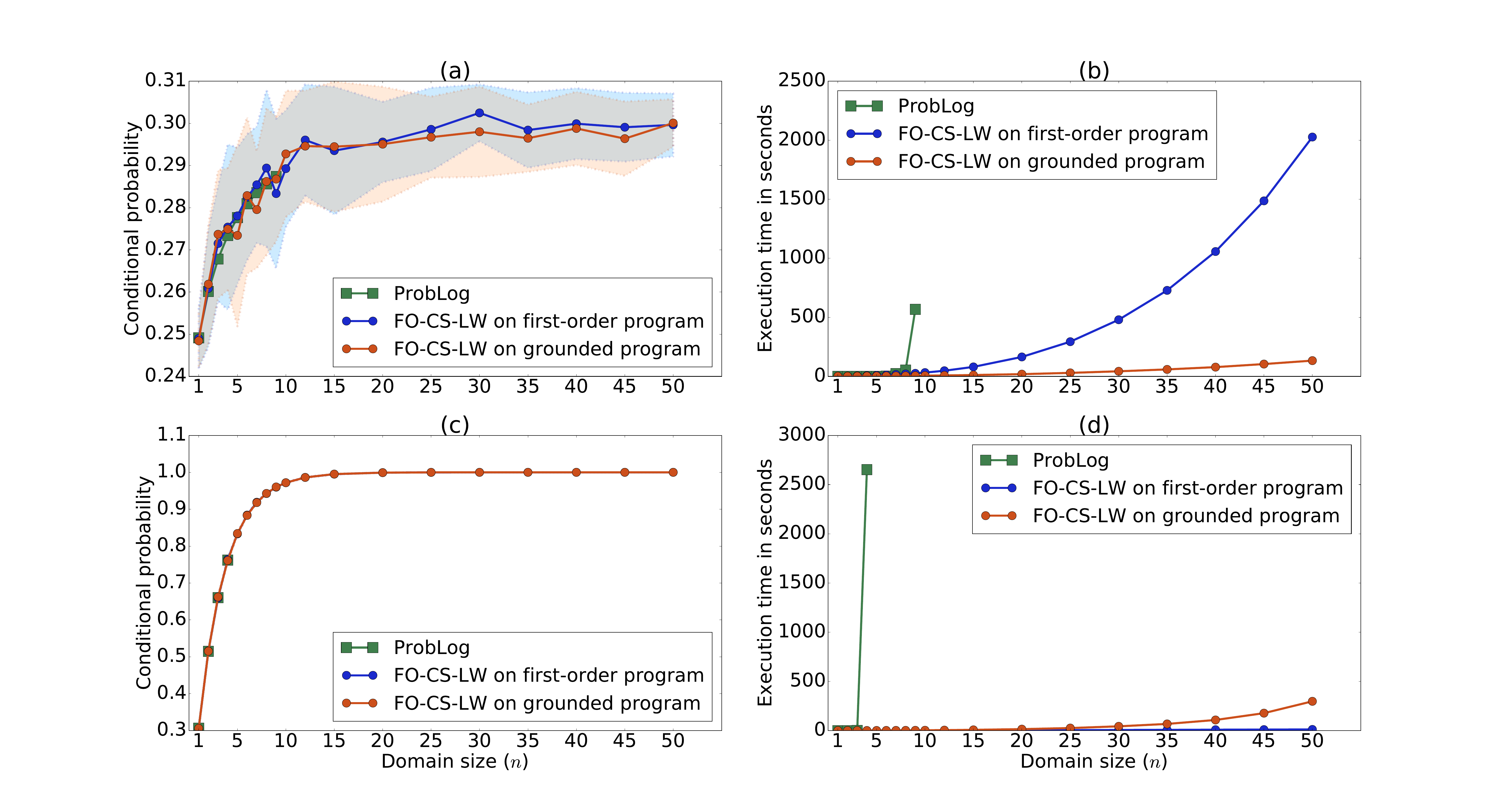}
     \caption{The first two graphs are for query $Q_1$, and the last two graphs are for query $Q_2$, mentioned in Section \ref{section: experiment 2}. (a) Comparison of exact probabilities of $Q_1$ computed by ProbLog, and probabilities estimated by FO-CS-LW when applied to the first-order and the grounded program. (b) The average time for processing $Q_1$ versus the domain size. (c) Comparison of probabilities of query $Q_2$ for the three cases. (d) The average time for processing $Q_2$ versus the domain size. To estimate the probabilities, FO-CS-LW used $10,000$ samples. The size of $n$ indicates that $n$ clients, $n$ accounts, and $n$ loans are present in the program. The shaded region denotes the standard deviation from the mean probability estimated by FO-CS-LW when executed $30$ times for each $n$. The execution time includes the grounding time when FO-CS-LW is applied to grounded programs.
     }
     \label{fig: problog comparision}
\end{figure}

We also compared the performance of FO-CS-LW when applied directly to equivalent first-order programs versus when applied to the grounded programs. Indeed, when the full ground network is huge, and the requisite network is also huge, it is better first to ground the programs and then apply FO-CS-LW. This is because unifications used to reason in first-order programs are somewhat costly operations, which are performed once if programs are grounded first. This case is illustrated in Figure \ref{fig: problog comparision}, where
the probability of query $Q_1 = P$($\mathtt{high\_savings(a1)\cong t}$ $\mid$ $\mathtt{home\_loan(l1)\cong f}$, $\mathtt{debt(c1)\cong t}$, $\mathtt{has\_loan(c1,l1)\cong f}$) is estimated and FO-CS-LW performs well if programs are grounded first. However, if the full ground network is huge, but the requisite network is very small, it is better to reason on the first-order level. This is because the cost of searching a few relevant clauses in a huge set of ground clauses exceeds the unification cost.
This is the case when the query ($Q_2$) is to compute $P(\mathtt{debt(c1)\cong t})$ given all other RVs (all RVs of type $\mathtt{has\_loan(C,L)}$ were set to $\mathtt{f}$ (``false''), all RVs of type $\mathtt{high\_savings(A)}$ were set to $\mathtt{f}$, and rest RVs were set to $\mathtt{t}$ (``true''). Furthermore, ProbLog could not compute probabilities beyond the domain size of $9$ on a machine with $132$ GB main memory, whereas FO-CS-LW quickly scaled to a domain size of 50.

\begin{figure}[t]
 \begin{adjustwidth}{-0.0cm}{}
\begin{tabular}{l | l}
\parbox{.4\textwidth}{
{\tiny
\begin{align*}
    & \mathtt{client(c1). \quad client(c2). \quad \dots}\\
    & \mathtt{account(a1). \quad account(a2). \quad \dots}\\
    & \mathtt{loan(l1). \quad loan(l2). \quad \dots}\\
    & \mathtt{0.7::home\_loan(L) \mathop{:\!\!-} loan(L).}\\
    & \mathtt{0.3::high\_savings(A) \mathop{:\!\!-} account(A).} \\
    & \mathtt{0.01::has\_account(C,A) \mathop{:\!\!-} client(C), account(A).} \\
    & \mathtt{0.02::account\_loan(A,L) \mathop{:\!\!-} account(A), loan(L).} \\
    & \mathtt{0.9::has\_loan(C,L) \mathop{:\!\!-} has\_account(C,A), account\_loan(A,L).} \\
    & \mathtt{0.001::has\_loan(C,L) \mathop{:\!\!-} client(C), loan(L).} \\
    & \mathtt{0.9::debt(C) \mathop{:\!\!-}  has\_loan(C,L), home\_loan(L).} \\
    & \mathtt{0.6::debt(C) \mathop{:\!\!-} has\_loan(C,L), \backslash \! + \! home\_loan(L).} \\
    & \mathtt{0.3::debt(C) \mathop{:\!\!-} has\_account(C,A), \backslash \! + \! high\_savings(A).} \\
    & \mathtt{0.01::debt(C) \mathop{:\!\!-} client(C).}
\end{align*}
}%
}
&
\parbox{.55\textwidth}{
{\tiny
\begin{align*}
    & \mathtt{client(c1) \sim val(t). \quad client(c2) \sim val(t). \quad \dots}\\
    & \mathtt{account(a1) \sim val(t). \quad account(a2) \sim val(t). \quad \dots}\\
    & \mathtt{loan(l1) \sim val(t). \quad loan(l2) \sim val(t). \quad \dots}\\
    & \mathtt{home\_loan(L) \sim bernoulli(0.7) \leftarrow loan(L)\cong t .}\\
    & \mathtt{high\_savings(A)\sim bernoulli(0.3) \leftarrow account(A)\cong t .} \\
    & \mathtt{has\_account(C,A)\sim bernoulli(0.01) \leftarrow client(C) \cong t, account(A)\cong t .} \\
    & \mathtt{account\_loan(A,L) \sim bernoulli(0.02) \leftarrow account(A)\cong t , loan(L)\cong t.} \\
    & \mathtt{has\_loan(C,L) \sim bernoulli(0.9) \leftarrow has\_account(C,A)\cong t , account\_loan(A,L)\cong t .} \\
    & \mathtt{has\_loan(C,L) \sim bernoulli(0.001) \leftarrow client(C)\cong t, loan(L)\cong t .} \\
    & \mathtt{debt(C) \sim bernoulli(0.9) \leftarrow has\_loan(C,L)\cong t , home\_loan(L)\cong t.} \\
    & \mathtt{debt(C) \sim bernoulli(0.6) \leftarrow has\_loan(C,L)\cong t , home\_loan(L)\cong f .} \\
    & \mathtt{debt(C) \sim bernoulli(0.3) \leftarrow has\_account(C,A)\cong t , high\_savings(A)\cong f .} \\
    & \mathtt{debt(C) \sim bernoulli(0.01) \leftarrow client(C)\cong t .}
\end{align*}
}%
}
\end{tabular}
    \caption{(Left) A ProbLog program specifying a probability distribution over relationships and attributes of clients, accounts, and loans. (Right) The equivalent \dcplus program.}
    \label{fig: experiment 2 example}
 \end{adjustwidth}
\end{figure}

One might expect that FO-CS-LW would perform poorly as domain size increases, and more samples would be required to get a reasonable estimate of probabilities. Figure \ref{fig: problog comparision} suggests the opposite. Using the same number of samples, the standard deviation from the mean does not increase as domain size increases. This is because FO-CS-LW exploits symmetries that arise due to noisy OR. Notice that exact probabilities do not change much as domain size increases because unique parameters do not increase even though the domain size increases and the number of parameters increases.



We conclude that FO-CS-LW scales with the domain size and can be useful on problems where the ProbLog inference algorithm fails.

\subsection{How does FO-CS-LW compare to the state-of-the-art inference algorithms for hybrid relational probabilistic models?}
\label{section: experiment 3}

\begin{figure}[t]
    \centering
    \includegraphics[width=1.0\linewidth]{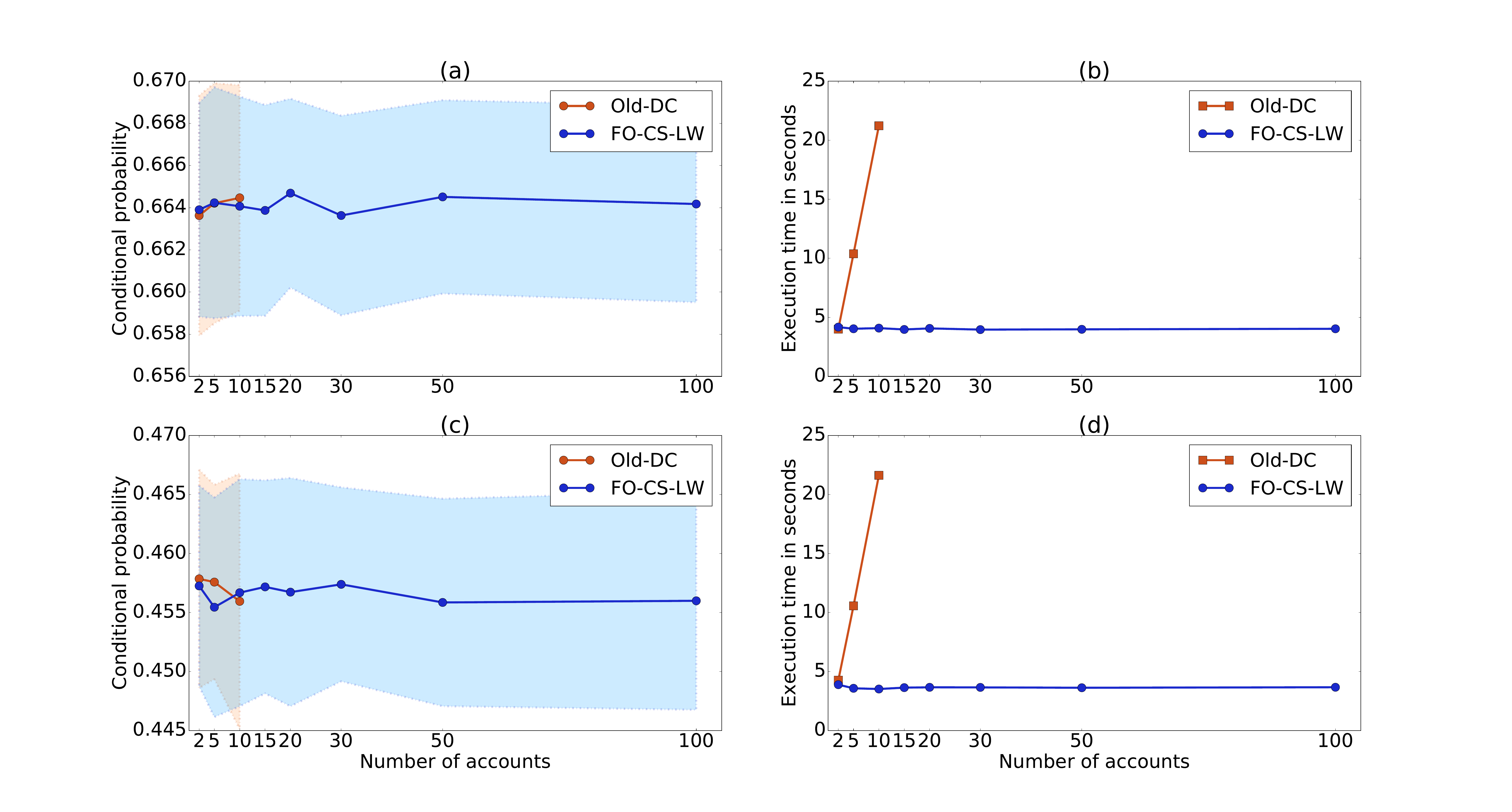}
    \caption{The first two graphs are for query $Q_3$, and the last two graphs are for query $Q_4$, mentioned in Section \ref{section: experiment 3}. (a) Comparison of estimates obtained using FO-CS-LW and Old-DC for query $Q_3$. (b) Average time taken by FO-CS-LW and Old-DC for $Q_3$. (c) Comparison of estimates for query $Q_4$. (d) Average time taken for $Q_4$. $10,000$ samples were used for each of the two algorithms. The shaded region denotes the standard deviation from the mean probability estimated by algorithms when executed $100$ times for each case. Old-DC suffered arithmetic underflow as observed data increased.}
    \label{fig: dc comparision}
\end{figure}

\begin{figure}
    \centering
    \includegraphics[width=1.0\linewidth]{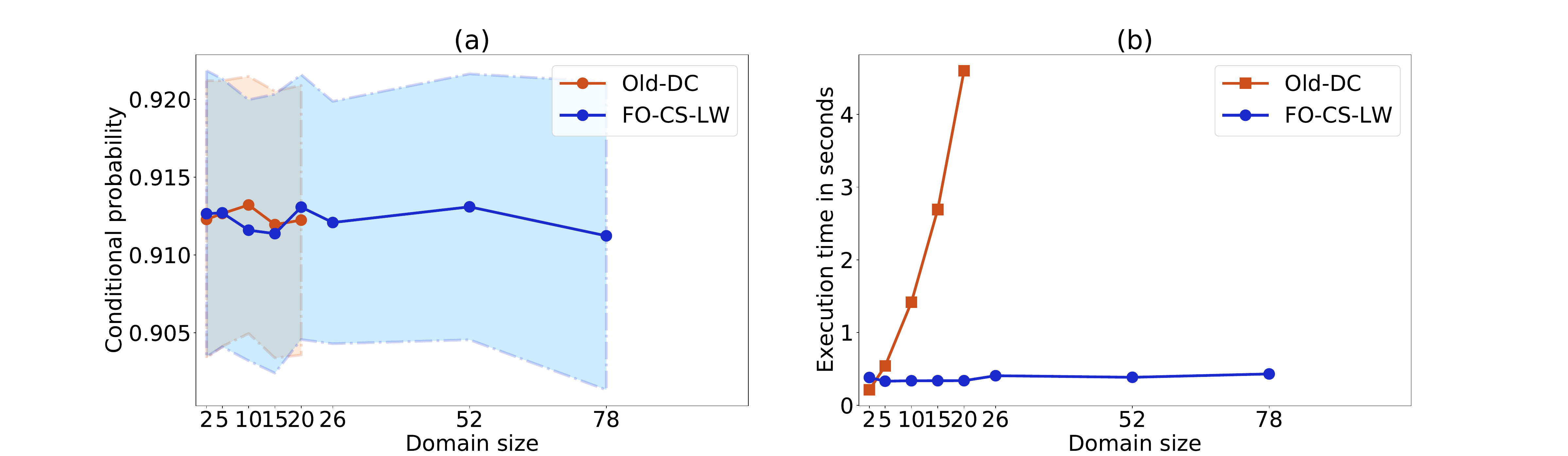}
    \caption{(a) Comparison of estimates obtained using FO-CS-LW and Old-DC for query $Q_5$. (b) Average time taken by FO-CS-LW and Old-DC for $Q_5$.}
    \label{fig: dc comparison nba}
\end{figure}

Those exact inference algorithms that exploit CSIs are not readily applicable to hybrid relational probabilistic models. There are not many open-source implementations of inference algorithms for such models\footnotemark. The exploitation of first-order CSIs and symmetries arising due to aggregations in such models is challenging, and to the best of our knowledge, no algorithm can exploit them in such models. To some extent, the likelihood weighting-based inference algorithm developed for the old version of DC can exploit them \cite{nitti2016probabilistic}. However, it does not filter out irrelevant evidence before inference, which is crucial in the case of relational databases. So, in this experiment, we aim to investigate how much FO-CS-LW improves upon old-DC's inference algorithm. \footnotetext{The publicly available code for BLPs is based on SICStus Prolog 3, the version that SICStus no longer supports. So, it is difficult to run the code.}

For this purpose, we used a real-world relational data generated by processing the financial database from the PKDD'99 Discovery Challenge. This data set is about services that a bank offers to its clients, such as loans, accounts, and credit cards. It contains information of four types of entities: $5,358$ clients, $4,490$ accounts, $680$ loans and $77$ districts. Ten attributes are of the continuous type, and three are of the discrete type. The data set contains four relations: $\mathtt{hasAccount/2}$ that links clients to accounts; $\mathtt{hasLoan/2}$ that links accounts to loans; $\mathtt{clientDistrict/2}$ that links clients to districts; and finally $\mathtt{clientLoan/2}$ that links clients to loans.

We learned a model in the form of distributional clauses as described in \cite{kumar2020learning} from the financial data set. The learned model, which was a program, specified a probability distribution over all attributes of all instances of the entities in the data set. Since the program was learned just as described in \cite{kumar2020learning}, relationships among entities could not be probabilistic. Furthermore, clauses in the program were mutually exclusive, and {\em aggregation atoms} and {\em statistical model atoms} were used in the bodies of the clauses. {\em Negations} were allowed in the bodies to deal with missing values or missing relationships. Details about these advanced constructs are present in Appendix \ref{Section: Advanced Constructs}. A snippet of the learned program is shown in Figure \ref{figure: snippet}.
\begin{figure}[t]
\begin{tabular}{l}
\parbox{1\textwidth}{
{\tiny
\begin{align*}
    & \% Entities\\
    & \mathtt{account(a\_10001) \sim val(true).}\\
    & \mathtt{loan(l\_7034) \sim val(true).}\\
    & \mathtt{client(c\_12303) \sim val(true).}\\
    & \mathtt{district(d\_24) \sim val(true).}\\
    & \mathtt{\dots} \\
    & \% Relations\\
    & \mathtt{clientLoan(c\_12303,l\_7034) \sim val(true).}\\
    & \mathtt{hasLoan(a\_10001,l\_7034) \sim val(true).}\\
    & \mathtt{hasAccount(c\_12303,a\_10001) \sim val(true).}\\
    & \mathtt{clientDistrict(c\_12303,d\_24) \sim val(true).}\\
    & \mathtt{\dots}\\
    & \% Model\\
    & \mathtt{loanAmount(L) \sim gaussian(151080.0,12853887266.5) \leftarrow loan(L)\cong true.}\\
    & \mathtt{\dots}\\
    & \mathtt{stdMonthInc(A) \sim gaussian(M,36225870.62) \leftarrow account(A)\cong true, avgSumOfW(A)\cong X, avgNrWith(A)\cong Y, avg(Z1,(hasLoan(A,L)\cong true, }\\
    & \qquad\mathtt{loanAmount(L)\cong Z1),Z), freq(A)\cong m, linear([X,Y,Z], [0.34,-516.12,0.004,12797.18], M).}
\end{align*}
}%
}
\end{tabular}
\caption{A snippet of the program learned from the financial data set. The aggregation atom $\mathtt{avg(\_,\_,\_)}$ in the body of the last clause collects all loans' amounts that a client has into a list and unifies $\mathtt{Z}$ with the average of the list. The statistical model atom $\mathtt{linear(\_,\_,\_)}$ implements a linear function $\mathtt{M\ is\  0.34\times X - 516.12\times Y + 0.004\times Z + 12797.18}$. Predicates like $\mathtt{loanAmount/1, stdMonthInc/1, \dots,}$ represent the attributes of entities.} 
\label{figure: snippet}
\end{figure}

Next, we created multiple subsets of the financial data set with varying numbers of accounts. These subsets were created by considering $\mathtt{account}$ to be the central entity. All information about clients, loans, and districts related to an account appeared in the same subset. All subsets had an account $\mathtt{a\_10001}$ that was linked with a loan $\mathtt{l\_7034}$. Two queries to the learned program were considered: i) $Q_3 = P$($\mathtt{stdMonthInc(a\_10001)\cong X , X > 22000.0}$) given the rest subsets of data, ii) $Q_4 = P$($\mathtt{loanAmount(l\_7034)\cong X, X > 20000.0}$) given the rest subsets of data. Figure \ref{fig: dc comparision} shows the comparison of estimates obtained for queries $Q_3$ and $Q_4$. The old-DC suffered arithmetic underflow as observed data increased, whereas FO-CS-LW scaled quite well. An important observation, in this case, is that the execution time of FO-CS-LW does not increase by adding more data. This is because real-world data is often not highly relational, e.g., clients generally have one or two accounts, not ten accounts. This is the case here. By adding more data, we are just adding data irrelevant to account $\mathtt{a\_10}$. FO-CS-LW that exploits CIs can detect that, but the old-DC inference engine can not.

Similar observations were made when we used the program learned from the real-world NBA data set as described in \cite{kumar2020learning}. This data set is about basketball matches from the National Basketball Association \cite{schulte2014aggregating}. It records information about matches played between two teams and actions performed by each player of those two teams in the matches. This data set also contains relations and discrete-continuous attributes. We created multiple subsets of the data set with varying numbers of actions (the domain size). These subsets were created by considering actions to be the central entity. All subsets had actions of a player with id $\mathtt{41}$ in game $\mathtt{27}$. The query that we considered was: $Q_5 = P$($\mathtt{points(27,41) \cong X, X > 10}$) given the rest subsets of data. As shown in Figure \ref{fig: dc comparison nba}, the observations are similar to those made for the financial dataset.

Thus, we conclude that when a significant amount of independencies are present in hybrid relational probabilistic models, FO-CS-LW outperforms the state-of-the-art inference algorithm of Old-DC.

\section{Related Work}\label{section: related work}
We describe relationships of the \dcplus framework introduced here with previous works in terms of representation and inference. 

\subsection{Relation to Representation Languages}
Several relational representation languages based on aggregation functions and combining rules have been introduced in the past.
Examples include probabilistic relational models \cite[PRMs]{friedman1999learning}, directed acyclic probabilistic entity-relationship models \cite[DAPER]{heckerman2004probabilistic}, probabilistic relational language \cite[PRL]{getoor2006prl}, Bayesian logic programs \cite[BLPs]{kersting20071}, and first-order conditional influence language \cite[FOCIL]{natarajan2008learning}. These languages consist of two components: a qualitative component representing the relational structure of the domain (either using graphs or definite clauses) and a quantitative component specifying conditional probability distributions (CPDs) in the tabular form. So, they do not qualitatively represent the structures present within the CPDs.

A tree or a rule-based representation lets us represent the structures qualitatively \cite{boutilier1996context,poole1997probabilistic,ngo1997answering}. That is why, probabilistic logic programming (PLP) has received much attention. Over the past three decades, many PLP languages have been proposed:  PHA \cite{poole1993probabilistic}, Prism \cite{sato1997prism}, LPADs \cite{vennekens2004logic}, ProbLog \cite{Raedt2007ProbLogAP}, ICL \cite{poole2008independent}, CP-Logic \cite{vennekens2009cp}. However, only a few of them support both discrete and continuous RVs: HProbLog \cite{gutmann2010extending}, DC \cite{gutmann2011magic,nitti2016probabilistic}, Extended-Prism \cite{islam2012inference},  Hybrid-cplint \cite{alberti2017cplint}, \cite{Michels2016ApproximatePI}. Nonetheless, the hybrid PLPs that have been studied in the past do not support combining rules, which is a core component of PLPs. BLPs do support combining rules, but then they do not qualitatively represent the structures. The syntax of \dcplus is the same as the syntax of DC introduced by \cite{nitti2016probabilistic}, but the semantics is different as \dcplus supports combining rules. The semantics of \dcplus is based on BLPs, so one can view \dcplus programs as BLPs qualitatively representing structures. Furthermore, a huge fragment of Problog programs, one of the most popular PLPs, is expressible as \dcplus programs. Annotated disjunctions and directed cycles are not supported in \dcplus currently.

\dcplus is a probabilistic programming language (PPL); thus, we should also describe how \dcplus relates to various other PPLs that generally extends imperative programming with probability distributions such as Infer.NET \cite{minka2010infer}, BLOG \cite{li2013blog}, Stan \cite{stan2015stan}, etc. \dcplus relates to these languages in the manner similar to how logic programming relates to imperative programming. One must use {\em if-then conditions} in these languages to express the structures and {\em for loops} to express aggregations/combining rules. Using statistical model atoms in the body of DCs, one can describe complex probabilistic models written in these languages.
However, \dcplus currently does not support writing open-universe models that some PPLs, such as BLOG, support. On the other hand, it is unclear how to write DC programs with negations in these PPLs, which is important when dealing with missing data.

\subsection{Relation to Inference Algorithms}
Resolving representation issues does not quickly imply that crucial issues inherent to inference are also resolved. The prime focus of past works on relational representation languages, for example, those discussed at the beginning of the previous section, has been the development of syntax and semantics so that these languages succinctly represent first-order probabilistic models. For inference, most of them construct ground BNs and rely on inference algorithms for BNs. Thus, they do not exploit symmetrical parameters that arise due to grounding the first-order models. Indeed they do not exploit structures present within CPDs of ground BNs since they do not even qualitatively represent them. 

The case of the related PPLs is quite surprising. A remarkable feature of PPLs is that the structures can be represented using if-then conditions in programs. However, equivalent BNs are constructed during inference, and algorithms for BNs are used; thus, CSIs implied by the structures are ignored. For example, Stan, the most popular PPLs that is also commercially used, does that. Interestingly, BLOG \cite{Milch2005ApproximateIF} does exploit CSIs, but a different class of CSIs, CSIs that arise in open-universe models due to the addition and subtraction of some RVs from the model when some other RVs take certain values. It does not exploit those that are implied by if-then conditions even in closed-universe models/programs. The exploitation of such CSIs and symmetrical parameters that appear after unrolling (similar to grounding in relational models) for loops even in closed-universe PPLs are not well studied.

Leaving expressive languages aside, the exploitation of CSIs implied by structures within CPDs of ground BNs is itself a difficult problem that has puzzled researchers for decades. Research in this direction has mainly been focused on exact inference \cite{boutilier1996context,poole2003exploiting}. Nowadays, it is common to use knowledge compilation-based exact inference for this purpose \cite{chavira2008probabilistic,fierens2015inference,shen2016tractable}; however, this approach does not apply to hybrid models. The problem of exploiting CSIs in hybrid models is non-trivial and is poorly studied. Recently, it has attracted some attention \cite{zeng2019efficient}. However, proposed approaches are also exact and rely on complicated weighted model integration \cite{belle2015probabilistic}, which do not scale. CS-LW that we propose for BNs is simple, scalable, and applies to hybrid BNs.

FO-CS-LW that extends CS-LW to first-order \dcplus programs also exploits symmetries in such programs. So, it makes sense to relate FO-CS-LW with the literature of lifted inference algorithms that aim to exploit symmetries in first-order models. Broadly these algorithms can be divided into two categories. Firstly, those that exploit symmetries but do not exploit CSIs. Belonging to this category, there are exact \cite{poole2003first,10.5555/1597348.1597367,Getoor2007LiftedFP,kisynski2009lifted,Taghipour2013LiftedVE} and approximate algorithms \cite{Niepert2012LiftedPI,Ahmadi2013ExploitingSF,Chen2020LiftedHV}. Belonging to the second category are those that also exploit CSIs. These algorithms are generally exact and do not easily apply to hybrid models \cite{VandenBroeck2011LiftedPI,Gogate2011ProbabilisticTP}. Recently, an exact inference algorithm belonging to the second category was applied to hybrid models \cite{feldstein2021lifted}. However, this algorithm inherits the problem of weighted model integration, i.e., scalability. It is well known that exact lifted inference has two major limitations: i) only a small class of models is liftable \cite{VandenBroeck2011OnTC,Kazemi2016NewLC}; ii) the inference is hard when binary relations are observed \cite{van2013complexity}. FO-CS-LW belongs to the second category of algorithms, which is approximate thus does not have these limitations. It applies to  \dcplus language in which a larger class of models can be written, which is remarkable.

\section{Future Work}\label{section: future work}
Our paper intends to introduce an interesting probabilistic programming framework that can be further developed in the future. Several features can be added to this framework, which are discussed next. 

The semantics of \dcplus can be extended to support open-universe models. The syntax already supports them, which was shown at the beginning of Section \ref{Section: sematics}. One can borrow ideas from BLOG for this purpose \cite{milch2010extending}. However, extending FO-CS-LW to the language supporting open-universe models will require solving a complex problem: how to detect requisite networks in such models while sampling? To detect them in closed-universe models, RV and dependency sets are automatically constructed by static analysis of programs as discussed in this paper; however, it is difficult to construct such sets for open-universe models as it might require dynamic analysis. 

Our algorithm does not support inference in \dcplus programs specifying mixtures of discrete and continuous distributions. This is the case when some clauses state that an RV is distributed according to continuous distributions, while others state that the same RV is distributed according to discrete distributions. More precisely, suppose a program contains the following two clauses:
\begin{align*}
    & \mathtt{credit\_score(C) \sim gaussian(700,10.9) \leftarrow has\_loan(C,L)\cong true.} \\
    & \mathtt{credit\_score(C) \sim discrete([0.9:600, 0.1:650]) \leftarrow has\_loan(C,L)\cong false.}
\end{align*}
It is not clear whether the client's credit score is of discrete type or continuous type. Likelihood weighting (LW) based algorithms do not estimate correct probabilities in such a case. To resolve this issue, \cite{wu2018discrete} proposed lexicographic LW. A context-specific variant of the lexicographic LW is needed for such programs. This has been left for the future.

As shown in Example \ref{example: DC example with CR}, the $\mathtt{val(.)}$ functor in \dcplus provides an elegant way of describing deterministic dependencies (constraints) among RVs in probabilistic models \cite{Mateescu2009MixedDA}. It is, however, well known that sampling algorithms, in general, perform poorly when a significant amount of determinism is present in models. So, an important and interesting avenue for future work is to combine constraint propagation with FO-CS-LW. 

Finally, we aim to open up a new direction towards improved sampling algorithms that exploit CSIs. Like LW, we believe MCMC algorithms can also be extended along the same line.

\section{Conclusion}\label{section: conclusion}

We have introduced a hybrid PLP framework called \dcplus.
It supports combining rules 
required to describe relational models succinctly. 

We have emphasized the exploitation of both CIs and CSIs for efficient inference. After realizing that a sampling algorithm that can properly do that is not well studied, we studied the role of CSIs in sampling. Subsequently, we introduced a notion of contextual assignment to show that CSIs allow for breaking the main problem of estimating conditional probability queries into several small problems that can be estimated independently. Based on this notion, we presented CS-LW that exploits CSIs implied by structures present within CPDs of BNs. We empirically showed that when a significant amount of structures are present, CS-LW generates samples faster than the standard LW and provides a better estimate of the query with much fewer samples. 

Since CS-LW is based on theorem proving, we easily extended it with unification and substitution of logical variables for inference in first-order \dcplus programs. The resulting first-order inference algorithm, called FO-CS-LW, naturally exploits symmetries or ICIs that arise due to combining rules in programs. 

Empirical results showed that FO-CS-LW scales with the domain size and outperforms the inference algorithm of the state-of-the-art hybrid PLP.

\acks{This work has received funding from the European Research Council (ERC) under the European Union’s Horizon 2020 research and innovation programme (grant agreement No [694980] SYNTH: Synthesising Inductive Data Models). OK was supported by Czech Science Foundation project ``Generative Relational Models’' (20-19104Y) and partially by the OP VVV project {\it CZ.02.1.01/0.0/0.0/16\_019/0000765} ``Research Center for Informatics’’. Part of this work was done while NK was visiting CTU in Prague, supported by Research Center for Informatics.}

\vskip 0.2in
\bibliography{mybibfile}
\bibliographystyle{newtheapa}

\appendix
\section{Missing Proofs}\label{section: missing proofs}

\subsection{Proof of Lemma \ref{theorem: bayes-ball 1}}
In this section, we present the detailed proof of Lemma  \ref{theorem: bayes-ball 1}. 
\begin{proof}
Let us denote the variables in $\mathbf{Z}$ that are marked on the top (requisite) by $\mathbf{Z}_\star$ and that are not marked on the top (not requisite) by $\mathbf{Z}_{\bar{\star}}$. The required probability $\mu$ is then given by,
\begin{equation*}
    \begin{aligned}
    \mu = P(\mathbf{x}_q \mid \mathbf{e}) = 
    \frac{\sum_{\mathbf{x}, \mathbf{z}_{\star}, \mathbf{z}_{\bar{\star}}} P(\mathbf{x},  \mathbf{z}_{\star}, \mathbf{z}_{\bar{\star}}, \mathbf{e}) f(\mathbf{x})}{\sum_{\mathbf{x}, \mathbf{z}_{\star}, \mathbf{z}_{\bar{\star}}} P(\mathbf{x},  \mathbf{z}_{\star}, \mathbf{z}_{\bar{\star}}, \mathbf{e})} = \frac{\sum_{\mathbf{x}, \mathbf{z}_\star} P(\mathbf{x},  \mathbf{z}_\star, \mathbf{e}) f(\mathbf{x}) \sum_{\mathbf{z}_{\bar{\star}}} P(\mathbf{z}_{\bar{\star}} \mid \mathbf{x},  \mathbf{z}_\star, \mathbf{e}) }{\sum_{\mathbf{x}, \mathbf{z}_\star} P(\mathbf{x},  \mathbf{z}_\star, \mathbf{e}) \sum_{\mathbf{z}_{\bar{\star}}} P(\mathbf{z}_{\bar{\star}} \mid \mathbf{x},  \mathbf{z}_\star, \mathbf{e})}
    \end{aligned}
\end{equation*}
Since $\sum_{\mathbf{z}_{\bar{\star}}} P(\mathbf{z}_{\bar{\star}} \mid \mathbf{x},  \mathbf{z}_\star, \mathbf{e}) = 1$, we can write,
\begin{equation*}
    \begin{aligned}
    \mu = \frac{\sum_{\mathbf{x}, \mathbf{z}_\star} P(\mathbf{x},  \mathbf{z}_\star, \mathbf{e}) f(\mathbf{x})}{\sum_{\mathbf{x}, \mathbf{z}_\star} P(\mathbf{x},  \mathbf{z}_\star, \mathbf{e})}
    \end{aligned}
\end{equation*}
Now let us denote the observed variables in $\mathbf{E}$ that are visited (requisite) by $\mathbf{E}_r$ and those that are not visited (not requisite) by $\mathbf{E}_n$. We can write, 
\begin{equation*}
    \begin{aligned}
    \mu = \frac{\sum_{\mathbf{x}, \mathbf{z}_\star} P(\mathbf{x},  \mathbf{z}_\star, \mathbf{e}_r) P(\mathbf{e}_n \mid \mathbf{x},  \mathbf{z}_\star, \mathbf{e}_r) f(\mathbf{x})}{\sum_{\mathbf{x}, \mathbf{z}_\star} P(\mathbf{x},  \mathbf{z}_\star, \mathbf{e}_r) P(\mathbf{e}_n \mid \mathbf{x},  \mathbf{z}_\star, \mathbf{e}_r)}
    \end{aligned}
\end{equation*}
The variables in $\mathbf{X} \cup \mathbf{Z}_\star$ pass the Bayes-balls to all their parents and all their children, but $\mathbf{E}_n$ is not visited by these balls. The correctness of the Bayes-ball algorithm ensures that there is no active path from $\mathbf{X} \cup \mathbf{Z}_\star$ to any $E_n$ in $\mathbf{E}_n$ given $\mathbf{E}_r$. Thus $\mathbf{X}, \mathbf{Z}_\star \perp \mathbf{E}_n \mid \mathbf{E}_r$ and $P(\mathbf{e}_n \mid \mathbf{x},  \mathbf{z}_\star, \mathbf{e}_r) = P(\mathbf{e}_n \mid \mathbf{e}_r)$. After cancelling out the common term $P(\mathbf{e}_n \mid \mathbf{e}_r)$, we get,
\begin{equation*}
    \begin{aligned}
    \mu = \frac{\sum_{\mathbf{x}, \mathbf{z}_\star} P(\mathbf{x},  \mathbf{z}_\star, \mathbf{e}_r) f(\mathbf{x})}{\sum_{\mathbf{x}, \mathbf{z}_\star} P(\mathbf{x},  \mathbf{z}_\star, \mathbf{e}_r)}
    \end{aligned}
\end{equation*}
Now let us denote observed variables in $\mathbf{E}_r$ that are only visited by $\mathbf{E}_\smwhitestar$ and that are visited as well as marked on top by $\mathbf{E}_\star$. After canceling out the common term, we get the desired result, 
\begin{equation*}
    \begin{aligned}
    \mu = \frac{\sum_{\mathbf{x}, \mathbf{z}_\star} P(\mathbf{x},  \mathbf{z}_\star, \mathbf{e}_\star \mid \mathbf{e}_\smwhitestar) P(\mathbf{e}_\smwhitestar) f(\mathbf{x})}{\sum_{\mathbf{x}, \mathbf{z}_\star} P(\mathbf{x},  \mathbf{z}_\star, \mathbf{e}_\star \mid \mathbf{e}_\smwhitestar) P(\mathbf{e}_\smwhitestar)}  = 
    \frac{\sum_{\mathbf{x}, \mathbf{z}_\star} P(\mathbf{x},  \mathbf{z}_\star, \mathbf{e}_\star \mid \mathbf{e}_\smwhitestar) f(\mathbf{x})}{\sum_{\mathbf{x}, \mathbf{z}_\star} P(\mathbf{x},  \mathbf{z}_\star, \mathbf{e}_\star \mid \mathbf{e}_\smwhitestar)}
    \end{aligned}
\end{equation*}
\end{proof}

\subsection{Proof of Proposition \ref{theorem: RV set}} \label{proof: RV set}
In this section, we present the detailed proof of Proposition \ref{theorem: RV set}.

\begin{proof}

First, we show that if $\mathtt{rv(A)}$ is in the least Herbrand model of $rv(\mathbb{P})$ (denoted by $M_{rv(\mathbb{P})}$) then $\mathbb{P}$ defines $\mathtt{A}$ as an RV: Since $\mathbb{P}$ is well-defined, there will be at least one clause for $\mathtt{A}$ in $ground(\mathbb{P})_\omega$ whose body is true with respect to $\omega$ due to the exhaustiveness condition. So, $\mathbb{P}$ defines $\mathtt{A}$ as an RV.

Now, we show the converse: 
Let $\omega$ be an assignment where each $\mathtt{B}$ is assigned a value if $\mathtt{rv(B)}$ is in $M_{rv(\mathbb{P})}$. Given $\omega$, we construct a ground program $ground(\mathbb{P})_\omega$ such that there is a clause $\mathtt{A \sim D \leftarrow Body}$ in $ground(\mathbb{P})_\omega$ with atoms $\{\mathtt{B_1 \cong V_1, \dots, B_m \cong V_m}\}$ in $\mathtt{Body}$ and these atoms are true with respect to $\omega$. Then, due to the way \dcplus programs have been defined in Definition \ref{definition: dc sharp}, there must be a definite clause $\mathtt{rv(A) \leftarrow rv(B_1), \dots, rv(B_m)}$ in $ground(rv(\mathbb{P}))$. Now, $\{\mathtt{rv(B_1), \dots, rv(B_m)}\}$ are in $M_{rv(\mathbb{P})}$ since those RVs are assigned in $\omega$. So, $\mathtt{rv(A)}$ will be in $M_{rv(\mathbb{P})}$ since the body of the definite clause will be true. This completes the proof. 
\end{proof}

\subsection{Proof of Proposition \ref{theorem: dep set}} \label{proof: dep set}

In this section, we present the detailed proof of Proposition \ref{theorem: dep set}.
\begin{proof}
First, we show that if $\mathtt{A}$ directly influences $\mathtt{B}$ then $\mathtt{pa(B,A)}$ is in the least Herbrand model of $dep(\mathbb{P})$: Since $\mathtt{A}$ directly influences $\mathtt{B}$ there must be a clause $\mathcal{C} \equiv \mathtt{B \sim D \leftarrow B_1, \dots, B_n} \in ground(\mathbb{P})_\omega$ for some possible world $\omega$ such that $\mathtt{n > 0}$, $\mathtt{A}$ is a RV term in the body of $\mathcal{C}$, and the body is true with respect to $\omega$. This implies that there must be a clause $\mathtt{rv(B) \leftarrow rv(A_1), \dots, rv(A_m)}$ in $ground(rv(\mathbb{P}))$ such that $\{\mathtt{A_1, \dots, A_m}\}$ is the set of RV terms in the body of $\mathcal{C}$ and $\mathtt{A \in \{\mathtt{A_1, \dots, A_m}\}}$. Consequently, there must be a definite clause $\mathcal{C}' \equiv \mathtt{pa(B,A)\leftarrow rv(B), rv(A_1), \dots, rv(A_m)}$ in $ground(dep(\mathbb{P}))$ by construction. $\mathtt{\{A_1, \dots, A_m\}}$ is assigned in $\omega$ so due to Proposition \ref{theorem: RV set}: $rv(\mathbb{P}) \models \mathtt{(rv(A_1), \dots, rv(A_m))}$, consequently, $rv(\mathbb{P}) \models \mathtt{rv(B)}$. This is also true in $dep(\mathbb{P})$ since all clauses in $rv(\mathbb{P})$ are in $dep(\mathbb{P})$. So, the body of clause $\mathcal{C}'$ will be true and $\mathtt{dep(\mathbb{P}) \models pa(B,A)}$.



Now, we show the converse: Since $dep(\mathbb{P}) \models \mathtt{pa(B,A)}$, there must be a definite clause $\mathtt{pa(B,A) \leftarrow rv(B), rv(A_1), \dots, rv(A_m)}$ in $ground(dep(\mathbb{P}))$ such that  $\mathtt{A \in \{A_1, \dots, A_m\}}$ and the clause's body is true in $ground(dep(\mathbb{P}))$. This implies that there must be a definite clause $\mathtt{rv(B) \leftarrow rv(A_1), \dots, rv(A_m)} \in ground(rv(\mathbb{P}))$ whose body is true in $ground(rv(\mathbb{P}))$. Due to the way \dcplus programs are defined (Definition \ref{definition: dc sharp}), there must be a distributional clause $\mathtt{B \sim D \leftarrow Body} \in ground(\mathbb{P})_\omega$ for some $\omega$ such that $\mathtt{\{A_1\cong V_1, \dots, A_m\cong V_m\}}$ belongs to $\mathtt{Body}$ that is true in $\omega$. So, $\mathtt{A}$ directly influences $\mathtt{B}$, which completes the proof.

\end{proof}

\subsection{Proof of Theorem \ref{Theorem: the expecation}}
In this section, we present the detailed proof of Theorem \ref{Theorem: the expecation}. 
\begin{proof}
The expectation $\mathbb{E}_{Q_\star}[W_{\mathbf{\dot{e}}_\star}]$ is given by
\begin{equation*}
\begin{aligned}
    \sum_{\mathbf{x}, \mathbf{z}_\star} \prod_{u_i \in \mathbf{x} \cup \mathbf{z}_\star} P(u_i \mid \mathbf{pa}(U_i)) \prod_{v_i \in \mathbf{\dot{e}}_\star} P(v_i \mid \mathbf{pa}(V_i)).
\end{aligned}
\end{equation*}
The basis $\mathbf{\dot{S}}_\star$ is a subset of $\mathbf{X} \cup \mathbf{Z}_\star$ by Definition \ref{definition: Basis} . Let us denote $(\mathbf{X} \cup \mathbf{Z}_\star) \setminus \mathbf{\dot{S}}_\star$ by $\mathbf{Z}_{\diamond}$. We can now rewrite the expectation as follows, 
\begin{equation*}
\begin{aligned}
    \sum_{\mathbf{\dot{s}}_\star, \mathbf{z}_{\diamond}} \prod_{u_i \in \mathbf{\dot{e}}_\star \cup \mathbf{\dot{s}}_\star} P(u_i \mid \mathbf{pa}(U_i)) \prod_{v_i \in \mathbf{z}_\diamond} P(v_i \mid \mathbf{pa}(V_i)).
\end{aligned}
\end{equation*}
We will show that $Pa \notin \mathbf{Z}_{\diamond}$ for any $Pa \in \mathbf{Pa}(U_i)$, which will then allow us to push the summation over $\mathbf{z}_{\diamond}$ inside. Let us consider two cases: 
\begin{itemize}
    \item For $U_i \in \mathbf{\dot{E}}_\star$, let $Pa \in \mathbf{Pa}(U_i)$ be an unobserved parent of $U_i$, then there will be a direct causal trail from $Pa$ to $U_i$, consequently $Pa$ will be in the set $\mathbf{\dot{S}}_{\star}$.
    \item For $U_i \in \mathbf{\dot{S}}_{\star}$, there will be a causal trail $U_i \rightarrow \cdots\ B_j\ \cdots \rightarrow E$ such that $E \in \mathbf{\dot{E}}_\star$ and such that either no $B_i$ is observed or there is no $B_i$. Let $Pa \in \mathbf{Pa}(U_i)$ be an unobserved parent of $U_i$ then there will be a direct causal trail from $Pa$ to $U_i$, consequently, there will be such causal trail from $Pa$ to $E$ and $Pa$ will be in the set $\mathbf{\dot{S}}_{\star}$.
\end{itemize}
Hence, we push the summation over  $\mathbf{z}_{\diamond}$ inside and use the fact that $\sum_{\mathbf{z}_{\diamond}} \prod_{v_i \in \mathbf{z}_{\diamond}} P(v_i \mid \mathbf{pa}(V_i)) = 1$, to get the desired result.  
\end{proof}

\subsection{Proof of Theorem \ref{theorem: cslw_main}}
In this section, we present the detailed proof of Theorem \ref{theorem: cslw_main}. 
\begin{proof}
Since $\mathbf{X}, \mathbf{Z}_\star, \mathbf{E}_\star, \mathbf{E}_\smwhitestar$ are variables of the Bayesian network $\mathcal{B}$ and they form a sub-network $\mathcal{B}_\star$ such that $\mathbf{E}_\smwhitestar$ do not have any parent, we can always write,
\begin{equation*}\label{equation: theorem4-1}
\begin{aligned}
    P(\mathbf{x}, \mathbf{z}_\star, \mathbf{e}_\star \mid \mathbf{e}_\smwhitestar) = \prod_{u_i \in \mathbf{x} \cup \mathbf{z}_\dagger \cup \mathbf{e}_\dagger} P(u_i \mid \mathbf{pa}(U_i)) \prod_{v_i \in \mathbf{z}_\ddagger \cup \mathbf{e}_\ddagger} P(v_i \mid \mathbf{pa}(V_i))
\end{aligned}
\end{equation*}
such that $p \in \mathbf{x} \cup \mathbf{z}_\star \cup \mathbf{e}_\star \cup \mathbf{e}_\smwhitestar$ for all $p \in \mathbf{pa}(U_i)$ or $p \in \mathbf{pa}(V_i)$. Now consider the summation over all possible assignments of variables in $\mathbf{X}, \mathbf{Z}_\star$, that is: $\sum_{\mathbf{x}, \mathbf{z}_\star} P(\mathbf{x}, \mathbf{z}_\star, \mathbf{e}_\star \mid \mathbf{e}_\smwhitestar)$. We will write this summation as a summation over contextual assignments $\psi \in \Psi$. The collection of the ranges of all $\psi \in \Psi$ will have all possible assignments of variables in $\mathbf{X}, \mathbf{Z}_\star$ without any duplicates as assignments $\psi \in \Psi$ are mutually exclusive. So, we can always write, 
\begin{equation}\label{equation: theorem4-2}
\begin{aligned}
    \sum_{\mathbf{x}, \mathbf{z}_\star} P(\mathbf{x}, \mathbf{z}_\star, \mathbf{e}_\star \mid \mathbf{e}_\smwhitestar) = \sum_{\psi \in \Psi}\sum_{\mathbf{z}_\ddagger[\psi]} P(\mathbf{x}[\psi], \mathbf{z}_\dagger[\psi], \mathbf{z}_\ddagger[\psi],  \mathbf{e}_\dagger[\psi], \mathbf{e}_\ddagger[\psi] \mid \mathbf{e}_\smwhitestar)
\end{aligned}
\end{equation}
In the above equation, notice that the inner summation is over the range of $\psi$. To simplify notation, from now we denote $\{ \mathbf{x}[\psi]$, $\mathbf{z}_\dagger[\psi]$, $\mathbf{Z}_\ddagger[\psi]$,  $\mathbf{e}_\dagger[\psi]$, $\mathbf{e}_\ddagger[\psi] \}$ by $\{ \mathbf{x}$, $\mathbf{z}_\dagger$, $\mathbf{Z}_\ddagger$,  $\mathbf{e}_\dagger$, $\mathbf{e}_\ddagger \}$. After using the definition of contextual assignments, we have that, 
\begin{equation*}
\begin{aligned}
    P(\mathbf{x}, \mathbf{z}_\dagger, \mathbf{z}_\ddagger,  \mathbf{e}_\dagger, \mathbf{e}_\ddagger \mid \mathbf{e}_\smwhitestar)  = \prod_{u_i \in \mathbf{x} \cup \mathbf{z}_\dagger \cup \mathbf{e}_\dagger} P(u_i \mid \mathbf{ppa}(U_i)) \prod_{v_i \in \mathbf{z}_\ddagger \cup \mathbf{e}_\ddagger} P(v_i \mid \mathbf{pa}(V_i))
\end{aligned}
\end{equation*}
Since $p \notin \mathbf{z}_\ddagger$ for any $p \in \mathbf{ppa}(U_i)$, we can push the summation over $\mathbf{z}_\ddagger$ inside to get, 
\begin{equation}\label{equation: theorem4-3}
\begin{aligned}
    \sum_{\psi \in \Psi}\sum_{\mathbf{z}_\ddagger} P(\mathbf{x}, \mathbf{z}_\dagger, \mathbf{z}_\ddagger,  \mathbf{e}_\dagger, \mathbf{e}_\ddagger \mid \mathbf{e}_\smwhitestar) = \sum_{\psi \in \Psi}\prod_{u_i \in \mathbf{x} \cup \mathbf{z}_\dagger \cup \mathbf{e}_\dagger} P(u_i \mid \mathbf{ppa}(U_i)) \sum_{\mathbf{z}_\ddagger}\prod_{v_i \in \mathbf{z}_\ddagger \cup \mathbf{e}_\ddagger} P(v_i \mid \mathbf{pa}(V_i)).
\end{aligned}
\end{equation}
However, we get a strange term $\sum_{\mathbf{z}_\ddagger}\prod_{v_i \in \mathbf{z}_\ddagger \cup \mathbf{e}_\ddagger} P(v_i \mid \mathbf{pa}(V_i))$. 
Let $\mathbf{S}_{\ddagger}$ denote the basis of residual $\mathbf{e}_\ddagger$. We have that $\mathbf{S}_{\ddagger} \subseteq \mathbf{Z}_\ddagger$ by Definition \ref{Definition: safe contexual assignments}. Let us denote $\mathbf{Z}_\ddagger \setminus \mathbf{S}_{\ddagger}$ with $\mathbf{Z}_{\diamond}$. Now the strange term can be rewritten as,
\begin{equation*}
\begin{aligned}
    \sum_{\mathbf{s}_{\ddagger}, \mathbf{z}_{\diamond}} \prod_{u_i \in \mathbf{e}_\ddagger \cup \mathbf{s}_{\ddagger}} P(u_i \mid \mathbf{pa}(U_i)) \prod_{v_i \in \mathbf{z}_{\diamond}} P(v_i \mid \mathbf{pa}(V_i)).
\end{aligned}
\end{equation*}
In the proof of Theorem \ref{Theorem: the expecation}, we showed that the summation over variables not in $\mathbf{S}_\ddagger$ can be pushed inside; hence, $\mathbf{Z}_{\diamond}$ can be pushed inside. After using the fact that $\sum_{\mathbf{z}_{\diamond}} \prod_{v_i \in \mathbf{z}_{\diamond}} P(v_i \mid \mathbf{pa}(V_i)) = 1$, we conclude that the strange term is actually the expectation $\mathbb{E}_{Q_\star}[W_{\mathbf{e}_\ddagger}]$. Using Equation (\ref{equation: bb}), (\ref{equation: theorem4-2}), (\ref{equation: theorem4-3}) and rearranging terms, the result follows. 
\end{proof}

\subsection{Proof of Lemma \ref{theorem: partial assignments}}
In this section, we present the detailed proof of Lemma  \ref{theorem: partial assignments}. 
\begin{proof}
It is clear that a subset of unobserved variables is assigned. Let $\mathbf{Z}_\ddagger$ be a set of unobserved variables left unassigned. Let $E \in \mathbf{E}_\star$ be an observed variable. Consider two cases: 
\begin{itemize}
    \item All ancestors of $E$ are in  $\mathbf{Z}_\ddagger \cup \mathbf{E}_\smwhitestar \cup \mathbf{E}_\star$.
    \item Some ancestors of $E$ are in $\mathbf{Z}_\ddagger \cup \mathbf{E}_\smwhitestar \cup \mathbf{E}_\star$ and some are in $\mathbf{X} \cup \mathbf{Z}_\dagger$. Let $A \in \mathbf{X} \cup \mathbf{Z}_\dagger$ and let $A \rightarrow \cdots\ B_i\ \cdots \rightarrow E$ be a causal trail.  Some $B_i$ are observed in all such trails. 
\end{itemize}
Clearly, $E$ will not be visited from any parent in the first case, and in the second case, the visit will be blocked by observed variables. Consequently, $E$ will not be weighted, which completes the proof. 
\end{proof}

\subsection{Proof of Theorem \ref{theorem: dc-partial-justification}}
In this section, we present the detailed proof of Theorem \ref{theorem: dc-partial-justification}. 
\begin{proof}
Variables in $\mathbf{Z}_\ddagger$ are not assigned in the simulation; hence, it follows immediately from Lemma \ref{theorem: DC CSI} that the assignment is contextual. Assume by contradiction that $A \in \mathbf{X} \cup \mathbf{Z}_\dagger$, $E \in \mathbf{E}_\ddagger$ and there is a causal trail $A \rightarrow \cdots\ B_i\ \cdots \rightarrow E$ such that no $B_i$ is observed or there is no $B_i$. Since $ A $ is assigned, all children of $ A $ will be visited, and following the trail, the variable $ E $ will also be visited from its parent since there is no observed variable in the trail to block the visit. Consequently, $E$ will be weighted, which contradicts our assumption that $E$ is not weighted. Hence, the assignment is also safe.
\end{proof}

\subsection{Proof of Lemma \ref{theorem: DC CSI}}
In this section, we present the detailed proof of Lemma \ref{theorem: DC CSI}.
\begin{proof}
Since $A$ is assigned/weighted and rules in $\mathbb{P}$ are exhaustive, a rule $\mathcal{R} \in \mathbb{P}$ with $A$ in its head must have fired. Let $\mathbf{d}$ be a body and $\mathcal{D}$ be a distribution in the head of $\mathcal{R}$. Since each $d_i \in \mathbf{d}$ must be true for $\mathcal{R}$ to fire, $\mathbf{d} \subseteq \mathbf{c}$. We assume that rules in $\mathbb{P}$ are mutually exclusive. Thus, among all rules for $A$, only $\mathcal{R}$ will fire even when an assignment of some variables in $\mathbf{B}$ is also given. Hence, by definition of the rule $\mathcal{R}$,
we have that,
\begin{equation*}
    \mathcal{D} = P(A \mid \mathbf{d}) = P(A \mid \mathbf{c}) = P(A \mid \mathbf{c}, \mathbf{B})
\end{equation*}
\end{proof}

\subsection{Proof of Theorem \ref{theorem: DC Plus CSI}} \label{proof: DC Plus CSI}

In this section, we present the detailed proof of Theorem \ref{theorem: DC Plus CSI}.
\begin{proof}
Since random variable $A$ is sampled/weighted, $k$ clauses out of $n$ ground distributional clauses for $A$ must have fired. Those $n-k$ clauses that do not fire must have at least one atom in their body that is false. Let $\mathbf{d}$ be an assignment of random variables in the body of the $k$ clauses and $\mathcal{D}$ be a distribution obtained after applying a combining rule on the multiset of distributions in the head of the $k$ clauses. Since each $d \in \mathbf{d}$ must be true for $k$ clauses to fire, $\mathbf{d} \subseteq \mathbf{c}$. Even if an assignment of some variables in $\mathbf{B}$ is also given, the $n-k$ clauses cannot fire since one atom in the body of these clauses is already false. Hence, by definition, we have that
\begin{equation*}
    \mathcal{D} = P(A \mid \mathbf{d}) = P(A \mid \mathbf{c}) = P(A \mid \mathbf{c}, \mathbf{B})
\end{equation*}
\end{proof}

\section{Advanced Constructs in \dcplus Framework}\label{Section: Advanced Constructs}
This section describes advanced constructs such as negation, aggregates, and statistical models in the \dcplus framework, which are useful while writing complex relational models \cite{kumar2020learning}.
\begin{example}\label{ex: advanced constructs}
The following program illustrates advanced constructs such as negation, aggregates, and statistical models.
{\begin{align*}
    & \mathtt{client(ann) \sim val(true).}\\
    & \mathtt{loan(l\_1) \sim bernoulli(0.9).}\\
    & \mathtt{loan(l\_2) \sim bernoulli(0.9).}\\
    & \mathtt{age(C) \sim gaussian(40, 10.5) \leftarrow client(C)\cong true.}\\
    & \mathtt{has\_loan(C,L) \sim bernoulli(0.2) \leftarrow client(C)\cong true, loan(L)\cong true.}\\
    & \mathtt{status(L) \sim discrete([0.3:appr, 0.7:decl]) \leftarrow loan(L)\cong true.}\\
    & \mathtt{credit\_score(C) \sim gaussian(700, 10.9) \leftarrow has\_loan(C,L)\cong true, status(L)\cong appr.}\\
    & \mathtt{credit\_score(C) \sim gaussian(600, 20.5) \leftarrow has\_loan(C,L)\cong true, status(L)\cong decl.}\\
    & \mathtt{credit\_score(C) \sim gaussian(M, 30.2) \leftarrow has\_loan(C,L)\cong true, \backslash \! + \! status(L)\cong \_.}\\
    & \mathtt{credit\_score(C) \sim gaussian(750, 15.9) \leftarrow age(C)\cong Y,}\\
    & \qquad \mathtt{mode(X,(has\_loan(C,L)\cong true,status(L)\cong X),appr), linear([Y],[20.1,30.9],M).}
\end{align*}}
\end{example}
Next, we explain the advanced constructs of the above program one by one in detail.
\paragraph{Negation} The exhaustiveness condition for well-defined programs is somewhat contrived. In practice, it is difficult to specify conditional probability distributions of a RV given all possible assignments of its parents because real-world data is often inadequate and contains missing values. In such cases, it is convenient to describe models using negative literals in the bodies of DCs.

The second last clause in the program of Example \ref{ex: advanced constructs} has a negation in the body. The negation is interpreted as {\em negation as failure} as usual in logic programming. We understand from the above program that the loan's status can take three values: $\mathtt{appr}$ (``approved''), $\mathtt{decl}$ (``declined''), and ``undefined''. That is, its value can be unknown (or undefined) in some possible worlds. The loan's status takes the undefined value with a probability of $1$ when the loan's identity is false, and the program specifies a probability distribution over such worlds.

To avoid {\em floundering} \cite{nilsson1990logic}, we allow writing only those clauses $\mathtt{A_0 \sim D \leftarrow L_1, \dots, L_n}$ in \dcplus programs, which satisfy this condition: if a logical variable $\mathtt{J}$ occurs in the RV term of a negative literal $\mathtt{L_i}$ then $\mathtt{J}$ occurs in the RV term of a positive literal in $\{\mathtt{A_0, L_1, \dots, L_{i-1}}\}$. This means writing {\em unsafe negations} are not allowed. For example, writing the clause
\begin{equation*}
\mathtt{credit\_score(C) \sim gaussian(500, 30.2) \leftarrow \backslash \! + \! status(L)\cong \_.}
\end{equation*}
is not allowed because the logical variable $\mathtt{L}$ occurs in the RV term of the negative literal but it does not occur in any positive literal.

\paragraph{Aggregates} An alternative approach to combining rules, extensively used for describing models for relational data, is the aggregation function or aggregate. 
It is a function that maps a multiset of values to a single value. Standard examples are average (if values are numerical), mode (most frequently occurring value), maximum, minimum, cardinality, etc. For example, the mode maps a multiset $\mathtt{\{appr, decl, appr\}}$ to a value $\mathtt{appr}$. To express these functions, built-in aggregation predicates are used in conjunction with the second-order predicate $\mathtt{findall/3}$ as follows: $\mathtt{findall(T, G, L), aggr(L, R)}$, where $\mathtt{T}$ is a target variable such that it occurs in a goal $\mathtt{G}$, and $\mathtt{L}$ unifies with the multiset of the instantiations that $\mathtt{T}$ gets successively on backtracking over $\mathtt{G}$. This is just like Prolog's $\mathtt{findall/3}$ predicate \cite{Wielemaker2011SWIProlog}. The atom $\mathtt{findall(T, G, L)}$ succeeds with an empty multiset if goal $\mathtt{G}$ has no solutions. A built-in predicate $\mathtt{aggr/2}$ implements an aggregation function (e.g., mode) that maps the multiset $\mathtt{L}$ to a single value $\mathtt{R}$. When the multiset $\mathtt{L}$ is empty, the aggregation atom $\mathtt{aggr(L, R)}$ fails. We will use a shorthand notation $\mathtt{aggr(T,G,R)}$ for $\mathtt{findall(T, G, L), aggr(L, R)}$ and call it an {\em aggregation atom}. 

The last clause in the program of Example \ref{ex: advanced constructs} has an aggregation atom in its body. Here, the idea is to consider the joint influence of statuses of all loans that a client has on the distribution of the client's credit score (instead of letting each loan have its own independent influence as in Example \ref{example: DC example with CR}). There are three situations in which the aggregation atom will fail, and the body of the last clause will be true: (i) the client has no loan, (ii) it is undefined that the client has loans or not, (iii) the client has loans, but statuses of those loans are undefined.





The support for aggregates, which in turn requires support for the second-order predicate $\mathtt{findall/3}$, takes \dcplus outside the domain of first-order probabilistic logic. However, the semantics of programs with aggregates can also be understood in terms of the ground programs. Just like Prolog, atom $\mathtt{aggr(T,G,R)}$ gets its truth value after checking for truths of all instantiations of goal $\mathtt{G}$ in the ground program.

\paragraph{Statistical Model Atom} 

It maps outcomes of RVs in the body of a DC to parameters of the distribution in the head. Formally, a DC with a statistical model is a clause of the form $\mathtt{A_0 \sim D_\phi \leftarrow A_1, \dots, A_n, M_\psi}$, where $\mathtt{M_{\psi}}$ is an atom implementing a mathematical function that relates the values of RVs in $\mathtt{\{A_1, \dots, A_n\}}$ with parameters $\phi$ in distribution $\mathtt{D_{\phi}}$ via parameters $\psi$.

An example is shown in the last clause of the program of Example \ref{ex: advanced constructs}. Here, $\mathtt{linear}$ is a built-in predicate that implements a linear model relating the value of client's age and mean $\mathtt{M}$ of the distribution in the head using parameters $\mathtt{\psi = [20.1, 30.9]}$ as follows: $\mathtt{M\ is\  20.1\times X + 30.9}$.

Many well-known statistical models, such as linear regression, logistic regression, softmax regression, etc., can be used to describe very complex relational models \cite{kumar2020learning}. Importantly, we have integrated the full expressiveness of logic programming with the strengths of statistical models in learning intricate patterns. 

\subsection{Inference in Programs with Advanced Constructs}\label{Section: Inference in first order DC}
For inference, first, we should identify RVs from \dcplus programs with advanced constructs. 
\begin{definition}
Let $\mathbb{P}$ be a well-defined \dcplus program with advanced constructs. The RV set $rv(\mathbb{P})$ of the program is the set of definite clauses obtained by transforming each clause $\mathtt{A \sim D \leftarrow L_1, \dots, L_m \in \mathbb{P}}$ like this: 
\begin{enumerate}
    \item Let $\mathtt{Body}$ be the empty set. For each positive or negative literal $\mathtt{L_i}$ of the form $\mathtt{T \cong V}$, an atom $\mathtt{rv(T)}$ is added to $\mathtt{Body}$.
    \item A clause $\mathtt{rv(A) \leftarrow Body}$ is added to $rv(\mathbb{P})$.
\end{enumerate}
\end{definition}
\begin{example}\label{example: rv set adv.}
The RV set for the program in Example \ref{ex: advanced constructs} is:
{\small \begin{align*}
    & \mathtt{rv(client(ann)).}\\
    & \mathtt{rv(loan(l\_1)).}\\
    & \mathtt{rv(loan(l\_2)).}\\
    & \mathtt{rv(age(C)) \leftarrow rv(client(C)).}\\
    & \mathtt{rv(has\_loan(C,L)) \leftarrow rv(client(C)), rv(loan(L)).}\\
    & \mathtt{rv(status(L)) \leftarrow rv(loan(L)).}\\
    & \mathtt{rv(credit\_score(C)) \leftarrow rv(has\_loan(C,L)), rv(status(L)).}\\
    & \mathtt{rv(credit\_score(C)) \leftarrow rv(age(C)).}
\end{align*}}
\end{example}
The above definition differs from Definition \ref{definition: RV set} in only one way. It additionally specifies the way negative literals of the form $\mathtt{\backslash \! + \! T\cong V}$ should be handled. However, nothing special is done for them because just like positive literals that compare outcomes of ground RV terms with values, negative literals compare the outcomes with values or with the ``undefined'' value. Additionally, comparison and statistical model atoms are ignored while constructing the RV set. This is because they do not contain RV terms. The aggregation atoms do contain RV terms, but they 
do not constrain instantiations of logical variables appearing in other atoms. So, they are also ignored.

However, we must ensure that Proposition \ref{theorem: RV set} is valid even after using negations that relax the exhaustiveness condition. Now, it may happen that $rv(\mathbb{P}) \models \mathtt{rv(A)}$ but no distribution is specified for $\mathtt{A}$ in some ground programs. In this case, however, the ``undefined'' value is assigned to $\mathtt{A}$, so the exhaustiveness condition is still implicitly present and Proposition \ref{theorem: RV set} is valid.


Next, we should be able to identify direct influence relationships among RVs. However, the second-order aggregation atoms
complicate the construction of dependency sets. We add an additional rule in Definition \ref{definition: dependency set} to deal with it.
\begin{definition}
Let $rv(\mathbb{P})$ be the RV set of a well-defined \dcplus program $\mathbb{P}$ with advanced constructs. The dependency set $dep(\mathbb{P})$ is the union of $rv(\mathbb{P})$ and the set of definite clauses $pa(\mathbb{P})$ obtained by transforming each clause $\mathcal{C}\equiv \mathtt{A \sim D \leftarrow L_1, \dots, L_m \in \mathbb{P}}$ with non empty body as follows:
\begin{enumerate}
    \item Let $\mathtt{rv(A) \leftarrow rv(T_1), \dots, rv(T_n)}$ be the clause in $rv(\mathbb{P})$ corresponding to $\mathcal{C}$. 
    \item For each $\mathtt{rv(T_i)}$, a clause $\mathtt{pa(A,T_i) \leftarrow rv(A), rv(T_1), \dots, rv(T_n)}$ is added to $pa(\mathbb{P})$.
    \item For each literal $\mathtt{L_i}$ of the form $\mathtt{aggr(T,Q,R)}$,
    \begin{enumerate}
        \item Let $\mathtt{Q}$ be of the form $\mathtt{(Q_1, \dots, Q_n)}$ and let $\mathtt{Body}$ be the empty set. 
        \item For each positive or negative literal $\mathtt{Q_i}$ of the form $\mathtt{T \cong V}$, an atom $\mathtt{rv(T)}$ is added to $\mathtt{Body}$. 
        \item For each positive or negative literal $\mathtt{Q_i}$ of the form $\mathtt{T \cong V}$, a clause
        \begin{align*}
            \mathtt{pa(A,T) \leftarrow rv(A), rv(T_1), \dots, rv(T_n), Body}
        \end{align*}
        is added to $pa(\mathbb{P})$.
    \end{enumerate}
\end{enumerate}
\end{definition}

However, note that in Prolog, all free variables appearing in $\mathtt{findall/3}$ are bound with the existential operator. So, all free variables in $\mathtt{aggr/3}$ should be renamed before constructing the dependency set. 
\begin{example}
In addition to the RV set of Example \ref{example: rv set adv.}, the dependency set of Example \ref{ex: advanced constructs} contains the following clauses. Notice the renamings in the last two clauses.
{\small \begin{align*}
    & \mathtt{pa(age(C), client(C)) \leftarrow rv(age(C)), rv(client(C)).}\\
    & \mathtt{pa(has\_loan(C,L), client(C)) \leftarrow rv(has\_loan(C,L)), rv(client(C)), rv(loan(L)).}\\
    & \mathtt{pa(has\_loan(C,L), loan(L)) \leftarrow rv(has\_loan(C,L)), rv(client(C)), rv(loan(L)).}\\
    & \mathtt{pa(status(L), loan(L)) \leftarrow rv(status(L)), rv(loan(L)).}\\
    & \mathtt{pa(credit\_score(C), has\_loan(C,L)) \leftarrow rv(credit\_score(C)), rv(has\_loan(C,L)), rv(status(L)).}\\
    & \mathtt{pa(credit\_score(C), status(L)) \leftarrow rv(credit\_score(C)), rv(has\_loan(C,L)), rv(status(L)).}\\
    & \mathtt{pa(credit\_score(C), age(C)) \leftarrow rv(credit\_score(C)), rv(age(C)).}\\
    & \mathtt{pa(credit\_score(C), has\_loan(C,L_1)) \leftarrow rv(credit\_score(C)), rv(age(C)),}\\
    & \qquad \qquad \mathtt{rv(has\_loan(C,L_1)), rv(status(L_1)).}\\
    & \mathtt{pa(credit\_score(C), status(L_2)) \leftarrow rv(credit\_score(C)), rv(age(C)),}\\
    & \qquad \qquad \mathtt{rv(has\_loan(C,L_2)), rv(status(L_2)).}
\end{align*}}
\end{example}
After identifying RVs and direct influences among them, the simulation of programs defined in Algorithm \ref{algorithm: forward-backward first-order} is straightforward. One can easily add rules to the proof procedure (Algorithm \ref{algorithm: proof first-order}) for resolving these advanced constructs. In the implementation, predicates $\mathtt{rv/1, pa/2}$ should be {\em tabled} \cite{Warren1992MemoingFL} for efficiency.

\end{document}